\PassOptionsToPackage{table}{xcolor}
\documentclass[10pt,twocolumn,letterpaper]{article}

\usepackage[cvpr]{config/cvpr}  %

\definecolor{cvprblue}{rgb}{0.21,0.49,0.74}
\definecolor{feat2gsred}{RGB}{178,76,51}
\usepackage[pagebackref,breaklinks,colorlinks,allcolors=cvprblue,urlcolor=feat2gsred]{hyperref}

\usepackage{graphicx}
\usepackage{booktabs}
\usepackage{wrapfig}
\usepackage{array}
\usepackage{tikz}
\usepackage{pgfplots}
\usepackage{colortbl}
\usepackage[accsupp]{axessibility}  %
\usepackage{orcidlink}
\usepackage{tabularx}
\usepackage{tcolorbox}
\usepackage{tikz}
\usepackage{microtype}
\usepackage{fontawesome}
\usepackage[switch]{lineno} 
\usepackage[toc,page]{appendix}
\newcommand{\qheading}[1]{\noindent\mbox{\textbf{#1}}}
\newcommand{\pheading}[1]{\medskip\noindent\textbf{#1}}

\definecolor{bestgreen}{RGB}{153,200,76}
\definecolor{worstred}{RGB}{192,0,0}

\newcommand{\bgcolor}[2]{\setlength{\fboxsep}{0pt}\colorbox{#1}{\strut #2}}

\definecolor{cbad}{HTML}{FFD0D0} 
\definecolor{cmedium}{HTML}{FFF0D0}
\definecolor{cgood}{HTML}{90C060}  

\pgfdeclarehorizontalshading{rankgrad}{100bp}{
    color(0bp)=(cbad);
    color(50bp)=(cmedium);
    color(100bp)=(cgood)
}

\newcommand{\BGcolor}[3][HTML]{\definecolor{mycolor}{HTML}{#2}\bgcolor{mycolor}{#3}}

\newlength\savewidth\newcommand\shline{\noalign{\global\savewidth\arrayrulewidth
  \global\arrayrulewidth 1pt}\hline\noalign{\global\arrayrulewidth\savewidth}}

\definecolor{DeltaColor}{rgb}{0.039,0.73,0.71}
\definecolor{SigmaColor}{rgb}{0.98,0.45,0.0}
\definecolor{AlphaColor}{rgb}{0,0,0.8}
\definecolor{BetaColor}{rgb}{0.8,0,0.8}
\definecolor{GammaColor}{rgb}{0.514,0.34,0.224}
\definecolor{EpsilonColor}{rgb}{0.353,0.725,0.906}
\definecolor{PurpleColor}{HTML}{9839ff}
\definecolor{RedColor}{rgb}{0.949,0.275, 0.224}
\definecolor{citecolor}{HTML}{0071bc}

\newcommand{\dtu}{\href{https://fanegg.github.io/Feat2GS/\#dtu}{{\texttt{webpage}}}\xspace}
\newcommand{\vid}{\href{https://fanegg.github.io/Feat2GS/}{{\texttt{webpage}}}\xspace}

\newcommand{\suppl}{\textcolor{magenta}{\emph{Sup.Mat.}}\xspace}

\definecolor{customcolor3D}{HTML}{FED273}
\definecolor{customcolor2D}{HTML}{BBC990}
\newcommand{\highlightboxyellow}[1]{%
  \tikz[baseline=(n.base)] 
    \node[rounded corners=2pt, fill=customcolor3D, inner sep=1pt, draw=none] (n) {#1};%
}
\newcommand{\highlightboxgreen}[1]{%
  \tikz[baseline=(n.base)] 
    \node[rounded corners=2pt, fill=customcolor2D, inner sep=1pt, draw=none] (n) {#1};%
}

\newcommand{\gt}{{ground-truth}\xspace}

\newcommand{\itw}{\mbox{in-the-wild}\xspace}

\newcommand{\sota}{state-of-the-art\xspace}

\newcommand{\dustr}{\mbox{DUSt3R}\xspace}
\newcommand{\mastr}{\mbox{MASt3R}\xspace}
\newcommand{\midas}{\mbox{MiDaS}\xspace}
\newcommand{\dino}{\mbox{DINO}\xspace}
\newcommand{\dinotwo}{\mbox{DINOv2}\xspace}
\newcommand{\sam}{\mbox{SAM}\xspace}
\newcommand{\clip}{\mbox{CLIP}\xspace}
\newcommand{\radio}{\mbox{RADIO}\xspace}
\newcommand{\mae}{\mbox{MAE}\xspace}

\newcommand{\instantsplat}{\mbox{InstantSplat}\xspace}
\newcommand{\probe}{\mbox{Probe3D}\xspace}

\newcommand{\gmode}{\mbox{\textbf{G}eometry mode}\xspace}
\newcommand{\tmode}{\mbox{\textbf{T}exture mode}\xspace}
\newcommand{\amode}{\mbox{\textbf{A}ll mode}\xspace}

\newcommand{\geo}{\mbox{\underline{\textbf{G}eometry}}\xspace}
\newcommand{\tex}{\mbox{\underline{\textbf{T}exture}}\xspace}
\newcommand{\all}{\mbox{\underline{\textbf{A}ll}}\xspace}

\newcommand{\featgs}{\mbox{{Feat2GS}}\xspace}
\newcommand{\longtitle}{Probing Visual Foundation Models with Gaussian Splatting}
\newcommand{\ourtitle}{\featgs: \longtitle}

\definecolor{PurpleColor}{HTML}{8B008B}
\definecolor{OrangeColor}{rgb}{0.914,0.541,0.0.141}
\definecolor{GreenColor}{rgb}{0.137,0.573,0.565}

\DeclareMathOperator*{\argmin}{arg\,min}

\definecolor{cvprblue}{rgb}{0.21,0.49,0.74}

\title{
\ourtitle
}

\author{
Yue Chen$^{1,2}$ \quad Xingyu Chen$^{1,2}$ \quad Anpei Chen$^{2,3}$ \quad Gerard Pons-Moll$^{3,4}$ \quad Yuliang Xiu$^{2,5\dagger}$ \vspace{5pt}\\
$^1$Zhejiang University \quad $^2$Westlake University 
\quad $^3$University of Tübingen, Tübingen AI Center \\
$^4$Max Planck Institute for Informatics, Saarland Informatics Campus \\
$^5$Max Planck Institute for Intelligent Systems \\
\href{https://fanegg.github.io/Feat2GS/}{\texttt{\small fanegg.github.io/Feat2GS}}
}

\begin{document}

\twocolumn[{%
\renewcommand\twocolumn[1][]{#1}%
\maketitle
\begin{center}
\vspace{-15pt}
    \centering
    \captionsetup{type=figure}
        \begin{subfigure}[b]{0.5944\linewidth}
          \centering
          \includegraphics[width=\linewidth,page=1]{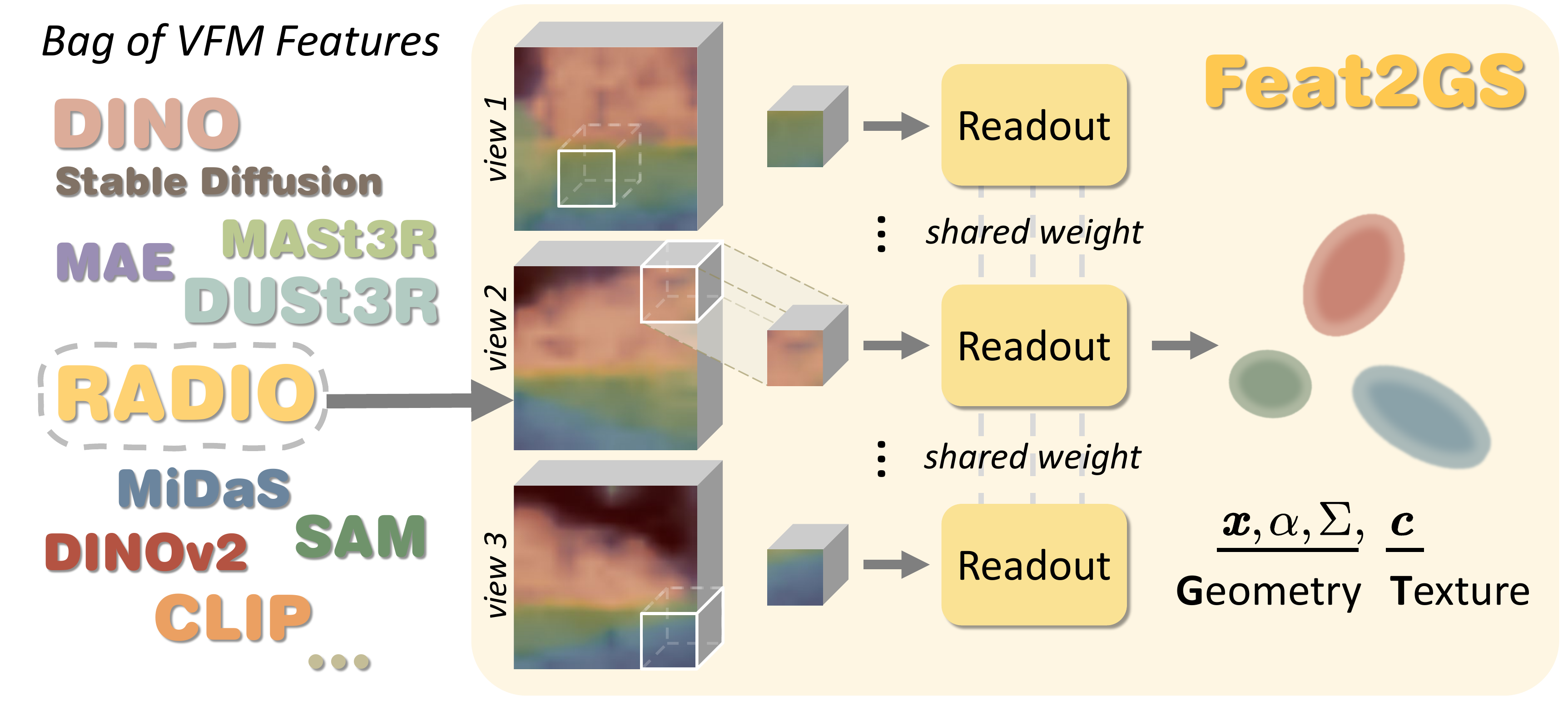}
          \caption{Feat2GS as VFM Probe}
          
          \label{fig:probe}
        \end{subfigure}
        \hspace{0.5pt}
        \begin{subfigure}[b]{0.2569\linewidth}
          \centering
          \includegraphics[width=\linewidth,page=1]{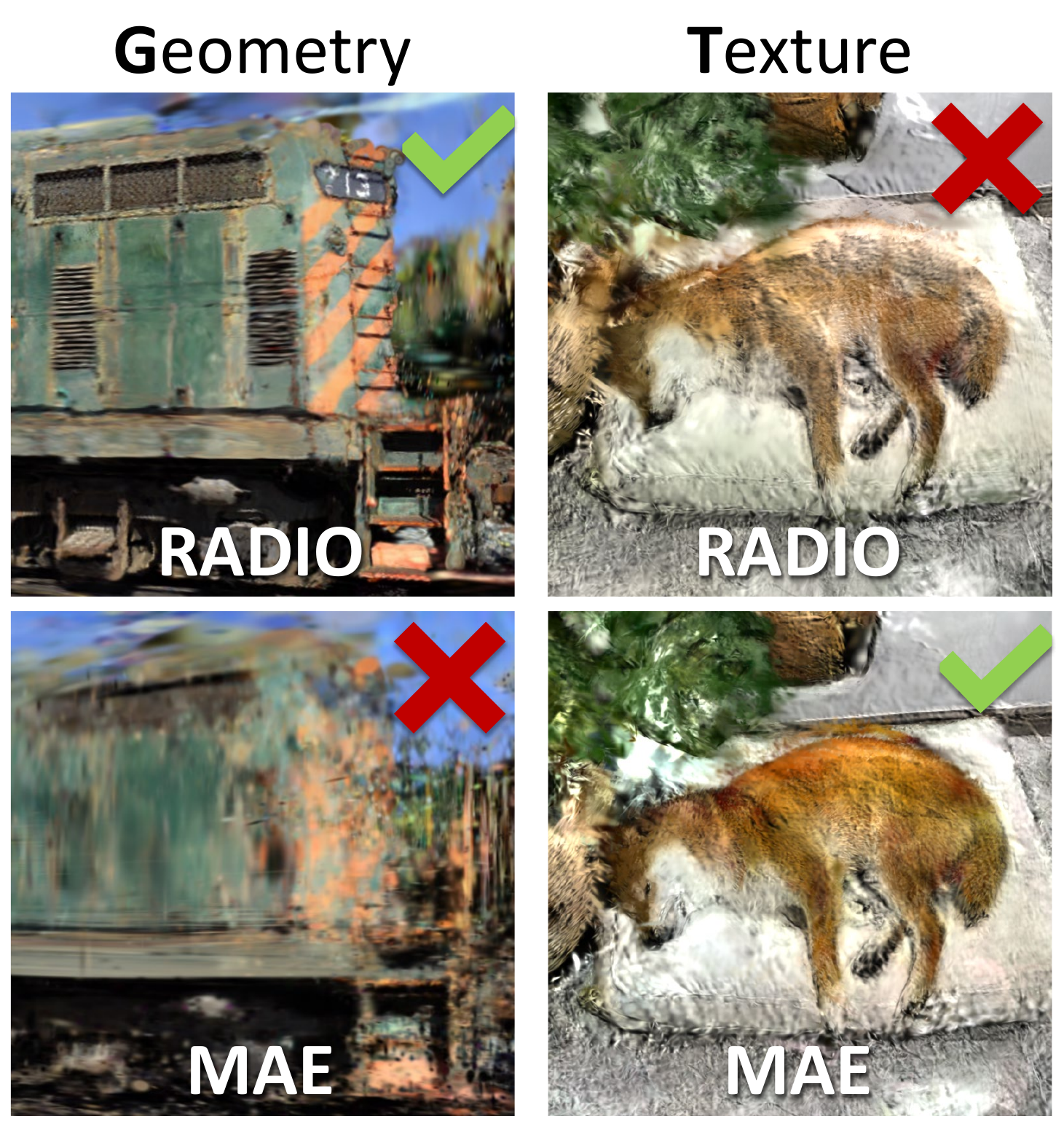}
          \caption{Extensive Analysis of VFM}
          
          \label{fig:analysis}
        \end{subfigure}
        \hspace{0.5pt}
        \begin{subfigure}[b]{0.1282\linewidth}
            \centering
            \includegraphics[width=\linewidth,page=1]{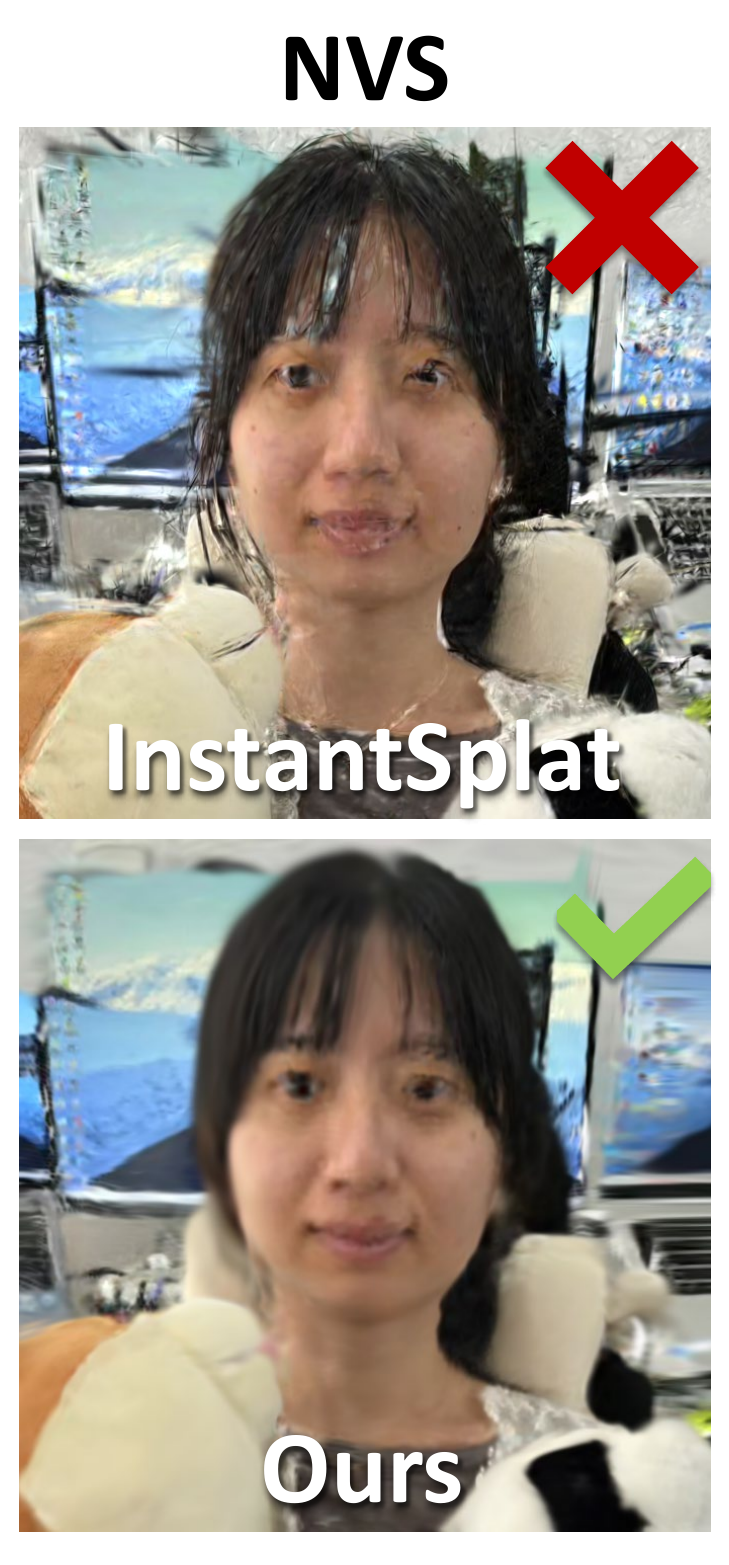}
            \caption{Strong Baseline}
            \label{fig:baseline}
          \end{subfigure}
          
    \caption{We present \textbf{\featgs}, 
    a unified framework to probe ``texture and geometry awareness'' of visual foundation models (VFMs).
    Novel-view synthesis (NVS) serves as an effective proxy for 3D evaluation.
    (a) Casually captured photos are input into VFMs to extract features and into a stereo reconstructor to obtain relative poses.
    Pixel-wise features are transformed into 3D Gaussians (3DGS) using a lightweight readout layer trained with photometric loss.
    (b) 3DGS parameters, grouped into \textbf{G}eometry and \textbf{T}exture, enable separate analysis of geometry/texture awareness in VFMs, evaluated by the NVS quality on diverse, unposed open-world images.
    (c) Our baseline derived from extensive empirical analysis, achieves superior performance for NVS by simply concatenating features from diverse VFMs.
    }
    
    \label{fig:pipe}
\end{center}
}]

\def\thefootnote{$\dagger$}\footnotetext{Corresponding Author}

\begin{abstract}
  Given that visual foundation models (VFMs) are trained on extensive datasets but often limited to 2D images, a natural question arises: how well do they understand the 3D world?
  With the differences in architecture and training protocols (\ie, objectives, proxy tasks),
  a unified framework to fairly and comprehensively probe their 3D awareness is urgently needed.
  Existing works on 3D probing suggest single-view 2.5D estimation (\eg, depth and normal) or two-view sparse 2D correspondence (\eg, matching and tracking). Unfortunately, these tasks ignore texture awareness, and require 3D data as \gt, which limits the scale and diversity of their evaluation set.
  To address these issues, we introduce \featgs, which readout 3D Gaussians attributes from VFM features extracted from unposed images.
  This allows us to probe 3D awareness for geometry and texture via novel view synthesis, without requiring 3D data.
  Additionally, the disentanglement of 3DGS parameters -- geometry ($\boldsymbol{x}, \alpha, \Sigma$) and texture ($\boldsymbol{c}$) -- enables separate analysis of texture and geometry awareness.
  Under \featgs, we conduct extensive experiments to probe the 3D awareness of several VFMs, and investigate the ingredients that lead to a 3D aware VFM.
  Building on these findings,
  we develop several variants that achieve \sota across diverse datasets. This makes \featgs useful for probing VFMs, and as a simple-yet-effective baseline for novel-view synthesis. Code and data are available at \href{https://fanegg.github.io/Feat2GS/}{\texttt{\small fanegg.github.io/Feat2GS}}.
\end{abstract}

\begin{figure}[t!]
  \centering
  \vspace{-5mm}
  \includegraphics[width=1\linewidth,page=1]{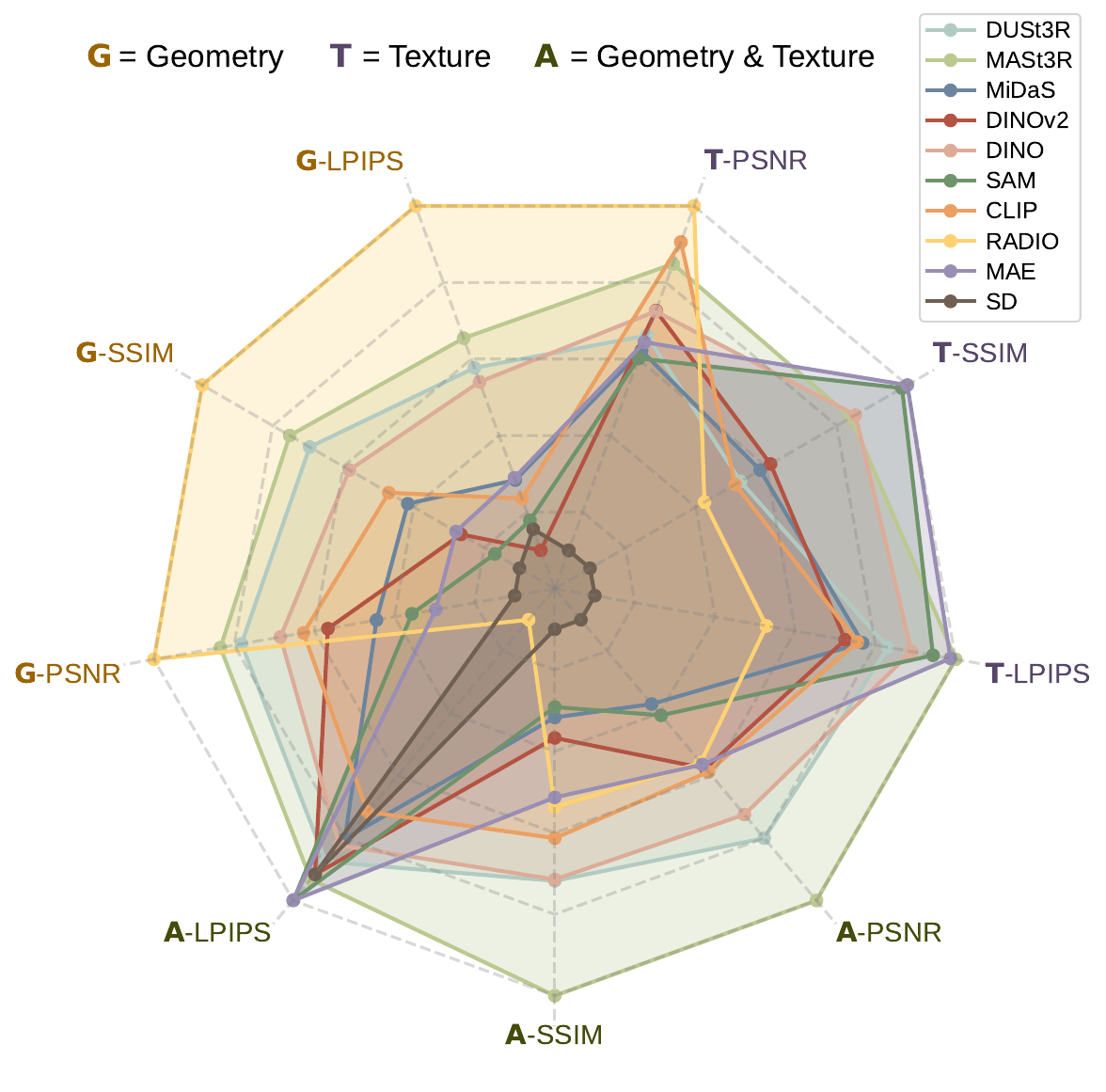}
  \vspace{-10pt}
  \caption{
      {\bf Texture+Geometry probing of mainstream VFMs.} Normalized average metrics for novel view synthesis (NVS) across six datasets are plotted on axes, with higher values away from the center indicating better performance. Try the interactive visualization demo on~\href{https://fanegg.github.io/Feat2GS/\#chart}{\texttt{\small fanegg.github.io/Feat2GS}}.
  }
  \label{fig:spider-chart}
  \vspace{-12pt}
\end{figure}

\section{Introduction}
\label{sec:intro}

Visual foundation models (VFMs)~\cite{FMs} have emerged as the 
basis for various 2D reasoning tasks~\cite{CLIP,SAM} and as a critical component for 3D fine-tuning~\cite{MiDaS,Marigold,DSINE,Spatialvlm,DUSt3R,pixelsplat,Lrm,stablenormal,fit3d}.
Their strong few-shot or zero-shot generalization ability stems mainly from
the expressive features~\cite{DINO,SD,RADIO}.
But what is the key for the 3D expressiveness? \textit{Does 3D awareness have to come from 3D data?} Some VFMs, such as \dinotwo and \mae, are trained using only 2D images. \textit{How important is the training approach?} VFMs differ in many aspects, such as learning strategies (\eg, self-supervised~\cite{MAE,DINO}, supervised learning~\cite{DUSt3R,MASt3R,SAM}), and proxy tasks (\eg, depthmap regression~\cite{DUSt3R}, cross-view completion~\cite{Croco}, generation~\cite{SD}). These differences make fair and comprehensive benchmarking difficult.

To answer these questions, recent works~\cite{Probe3D,rayr2d3} evaluate the geometry awareness of VFMs 
using two proxy tasks: 1) 2.5D depth/normal/token estimation from a single image~\cite{Probe3D,rayr2d3}, and 2) 2D matching/tracking between two views~\cite{Probe3D,aydemir2024can}. 
Though it does analyze the 3D awareness of current VFMs, it does not probe the texture awareness and multi-view dense consistency of VFM features, which are critical for 3D-related tasks, such as reconstruction and generation. 

For \textbf{``texture awareness''}, texture-invariant training improves geometry estimation but can harm texture preservation (see RADIO in \cref{fig:spider-chart}). However, recovering original textures from VFM features is key for training on large-scale 2D data with photometric loss~\cite{NoPoSplat,LSM,Lrm,Splatt3R} and texture synthesis~\cite{efros1999texture}.
\textbf{``Multi-view dense probing''}, like novel view synthesis (NVS)~\cite{FlowMap,acezero}, allows every input pixel to contribute to the evaluation, rather than just sparse matching points. Unlike 2D sparse matching, NVS only requires images, eliminating the need for costly labeling of visual correspondences.
With the numerous public multi-view datasets~\cite{Tanks,llff,Mip_nerf_360,Mvimgnet,ling2024dl3dv} available, covering diverse scenes and viewpoints, a new 3D probing approach using these datasets to evaluate texture and geometry awareness in dense mode could be invaluable.

Thus, we introduce \featgs, short for Feature2Gaussian, which evaluates both texture and geometry awareness of VFMs, in the NVS task, using only 2D multi-view data.
As shown in \cref{fig:pipe}, during training, \featgs extracts image features from the input views using pre-trained VFMs. A shallow MLP readout layer then regresses the parameters of 3D Gaussians~\cite{3DGS} from these features. Multi-view photometric loss minimizes the visual difference between renderings and inputs. During testing, visual similarity metrics (\ie, PSNR, SSIM, LPIPS) are measured for unseen views, across diverse datasets, with \cref{tab:DTU} demonstrating 
that these 2D metrics align well with 3D metrics.
To handle sparse and uncalibrated casual images, we initialize camera parameters using \dustr~\cite{DUSt3R} and refine them with photometric loss.

\setlength{\tabcolsep}{4pt}
\begin{table}[t]
    \centering
    \vspace{-5mm}
    \resizebox{1.0\linewidth}{!}{
        \begin{tabular}{l|ccc|c}
            \multicolumn{1}{l|}{\featgs} & -\textbf{G}eometry & -\textbf{T}exture & -\textbf{A}ll & \instantsplat~\cite{InstantSplat} \\
            \shline
            Feature-Readout & $\boldsymbol{x}, \alpha, \Sigma$ & $\boldsymbol{c}$ & $\boldsymbol{x}, \boldsymbol{c}, \alpha, \Sigma$ & -\\
            Free-Optimize & $\boldsymbol{c}$ & $\boldsymbol{x}, \alpha, \Sigma$ & - & $\boldsymbol{x}, \boldsymbol{c}, \alpha, \Sigma$ \\
        \end{tabular}}
\caption{\textbf{GTA probing schemes}. Unlike \instantsplat, \featgs uses shallow readout layer to parse VFM features into 3DGS. GTA modes include: \geo -- VFM features to Gaussian geometric parameters (\ie, positions $\boldsymbol{x}$, opacity $\alpha$, covariance $\Sigma$), \tex -- VFM features to Gaussian textural parameters, \ie, SH coefficients $\boldsymbol{c}$, and \all -- all parameters are regressed from features.}
\label{tab:feat2gs}
\vspace{-10pt}
\end{table}

The parameters of 3DGS, grouped into geometry ($\boldsymbol{x}, \alpha, \Sigma$) and texture ($\boldsymbol{c}$), enable separate analysis of VFM's texture and geometry awareness. Each group could switch between the ``Feature-Readout" and ``Free-Optimize" modes to use VFM features as input or free-optimized.
This leads to three probing schemes (short as GTA), as shown in \cref{tab:feat2gs}.

\medskip
In summary, our key contributions are as follows:\\
\qheading{1) \featgs as VFM probe.} \featgs offers a unified framework (\cref{fig:pipe}) to probe the 3D awareness (texture and geometry) of pre-trained VFMs, without using 3D labels.\\
\qheading{2) Extensive analysis of VFM.} 
We evaluate a wide range of mainstream VFMs (\cref{fig:spider-chart}) across diverse multi-view datasets (\cref{tab:GTA}), spanning from simple scenes to causal captures. These experiments reveal common drawbacks of VFMs and shed light on how to 
improve them (\cref{sec:results}).\\
\qheading{3) Strong baseline for NVS.} 
Motivated by these findings,
we design three variants of \featgs that outperform the current SOTA \instantsplat~\cite{InstantSplat} in all metrics (\cref{tab:app}).

\section{Related Work}
\label{sec:related}

\pheading{Measuring 3D Awareness of VFMs.}
There is no doubt that, visual foundation models~\cite{FMs}, short as VFMs, have significantly advanced various 3D vision tasks, such as geometric cue estimation~\cite{MiDaS,Marigold,DepthCrafter,DSINE,geowizard,stablenormal,khirodkar2024sapiens}, 6D pose estimation~\cite{foundpose}, visual tracking~\cite{dino_tracker}, and spatial reasoning~\cite{Spatialvlm,Chatpose}, \etc.
However, behind these advances and everyday SOTA records, \textit{are these VFMs truly 3D-aware, even when trained without any 3D data? If so, to what extent? And what enables such awareness?}
There is a line of works that try to answer these questions through multi-view object consistency~\cite{mochi}, spatial visual question answering~\cite{zuo2024towards,BLINK}, visual perspective taking~\cite{3D_PC} and robot learning~\cite{CortexBench,Theia,SPA}. Although in the same spirit of 3D probing, these existing works mainly focus on coarse-grained semantic reasoning, such as determining ``which marker is closer'' instead of fine-grained, or even pixel-wise spatial reasoning, like depth estimation.
Regarding the fine-grained 3D probes, they either use 2.5D proxy tasks, such as geometric cues estimation (\ie, depth, normal)~\cite{Probe3D} and view token estimation~\cite{rayr2d3}, or use two-view sparse point matching~\cite{Probe3D} and tracking~\cite{aydemir2024can} to assess the 3D awareness of VFMs. 
The main constraint of these fine-grained 3D probes is their reliance on labeled 3D data, which significantly limits fair and comprehensive evaluation on large-scale visual data.
\featgs addresses this by first regressing 3DGS from VFM features and then benchmarking 3D awareness via novel view synthesis. This comes with two advantages: ALL raw pixels can contribute to the final evaluation, and ANY multi-view captures can be leveraged. \featgs enables ``dense'' and ``diverse'' 3D probing.

\pheading{NVS from Casual Images.}
Novel view synthesis has made significant progress in recent years~\cite{NeRF,instant_ngp,3DGS,TensoRF,nerfstudio}. When it comes to sparse and causal captures, which is a quite challenging scenerio, various regularizers~\cite{Regnerf,Freenerf} or visual priors, such as depth~\cite{Depth_supervised_nerf,FSGS}, pre-trained visual features~\cite{DietNeRF,SinNeRF}, diffusion priors~\cite{Nerfbusters,Reconfusion} and feed-forward modeling~\cite{pixelNeRF,Zero-1-to-3,ZeroNVS,MegaScenes,pixelsplat,MVSplat,Lrm,LaRa,Long-LRM,LVSM}, have been introduced. 
However, these methods assume known camera poses from Structure-from-Motion~\cite{colmap}, which are not available for sparse captures with minimal overlap. Although some works attemp to optimize camera poses alongside NVS optimization~\cite{nerfmm}, using techniques like coarse-to-fine encoding~\cite{Barf}, local-to-global registration ~\cite{l2g}, geometric constrains~\cite{SCNeRF}, adversarial objectives~\cite{gnerf}, dense correspondence~\cite{Sparf}, and external priors~\cite{Nope-nerf,cf3dgs,LocalRF}, they can only handle dense-view or video-like sequences --- not sparse-view images.
Groundbreaking methods like \dustr~\cite{DUSt3R}, \mastr~\cite{MASt3R}, and subsequent works~\cite{spann3r,MonST3R} address these limitations by training models on large-scale datasets. They approach the pairwise reconstruction problem as a regression of point maps, easing the strict constraints of traditional projective camera models. This enables ``Unconstrained Stereo 3D Reconstruction'' of arbitrary image collections, without needing prior information about camera calibration or viewpoint poses. 
The predicted pointmap can directly initialize 3DGS~\cite{3DGS}, which can then be regressed in a two-view feedforward~\cite{Splatt3R,LSM,NoPoSplat} or optimized with multi-view photometric losses~\cite{InstantSplat}.
\instantsplat~\cite{InstantSplat} closely mirrors our target of optimizing 3DGS from sparse captures using \dustr estimated cameras. What sets our \featgs apart is that we readout 3DGS using visual features, instead of optimizing it in free form, see \cref{tab:feat2gs}. This can be done with a shallow readout MLP, helping to prevent overfitting.

\section{Method}
\label{sec:method}

\subsection{\featgs}
\label{sec:feat2gs}
We illustrate our pipeline in \cref{fig:pipe}. 
After extracting frozen feature maps from various visual foundation models (VFMs), we take the following steps to ensure fair probing: unifying the feature channel dimensions using Principal Component Analysis (PCA)~\cite{pca}, standardizing the spatial dimensions via bilinear upsampling, and maintaining a consistent network architecture for different VFM features.
Specifically, \featgs takes the compact features $\mathbf{f}_i$ of each pixel $i \in \left\{1, 2, \dots, n\right\}$ as input and 
output per-pixel Gaussian primitive via a readout layer $g_{\Theta}$:
\begin{align}
    \label{eqn:readout}
    \boldsymbol{G}_i = g_{\Theta}(\mathbf{f}_i)
\end{align}
where each Gaussian $\boldsymbol{G}_i$ is parameterized by: position $\boldsymbol x\in \mathbb{R}^3$, opacity $\alpha\in \mathbb{R}$, covariance matrix $\Sigma\in \mathbb{R}^{3\times 3}$, and three order of spherical harmonic (SH) coefficients $\left\{\boldsymbol c_i\in \mathbb{R}^{48}|i=1,2,...,n\right\}$.

To ensure the readout layer acts purely as an information conduit rather than a memory storage, 
we minimize its parameters 
that forcing the 3D Gaussians are decoded from the features. 
Specifically, the readout layer is constructed using 
a 2-layer MLP with 256 units per-layer and ReLU activation instead alternatives like dense prediction transformer~\cite{Probe3D}. 
With ReLU activation, it forms the minimal setup for nonlinear mapping.
Then we splat 3D Gaussians onto images via differentiable rasterization.
Note that, to enable our method to evaluate casually captured, sparse, and uncalibrated images, we use an unconstrained stereo reconstructor~\cite{MASt3R,spann3r,MonST3R}, DUSt3R~\cite{DUSt3R} in our experiments, to initialize camera poses $\boldsymbol{T}$, which are then jointly updated with the readout layer $g_{\Theta}^{(mode)}$ or freely-optimized Gaussian parameters $\boldsymbol{O}^{(mode)}$ in a specific mode,
using a simple photometric loss between renderings $\mathcal{R}_{v}(\cdot)$ and images $\{\mathcal{I}_v\}_{v=1}^{N}$:
\begin{align}
    \label{eqn:photometric}
    \Theta^{*}, \boldsymbol{O}^{*},\boldsymbol{T}^{*} = \underset{\Theta,\boldsymbol{O}, 
    \boldsymbol{T}}{\argmin} \sum\limits_{v \in N} \left\Vert \mathcal{R}_{v}(g_{\Theta}(\mathbf{f}),\boldsymbol{O},\boldsymbol{T}) - \mathcal{I}_{v} \right\Vert
\end{align}

To decouple the geometry and texture awareness, we propose three probing modes:
\geo reads out geometric parameters from the 2D image features, and freely optimizes textural parameters $\boldsymbol{c}_i$:
\begin{align}
\label{eqn:feat2gs-g}
    \{\boldsymbol{x}_i, \alpha_i, \Sigma_i\} = g_{\Theta}^{(G)}(\mathbf{f}_i), 
    \{\boldsymbol{c}_i\} = \boldsymbol{O}^{(G)}
\end{align}
\tex reads out textural parameters, and directly optimizes geometric parameters $\{\boldsymbol{x}_i, \alpha_i, \Sigma_i\}$: 
\begin{align}
    \label{eqn:feat2gs-t}
    \{\boldsymbol{c}_i\} = g_{\Theta}^{(T)}(\mathbf{f}_i), 
    \{\boldsymbol{x}_i, \alpha_i, \Sigma_i\} = \boldsymbol{O}^{(T)}
\end{align}
\all reads out all Gaussian parameters:
\begin{align}
    \label{eqn:feat2gs-a}
    \{\boldsymbol{x}_i, \alpha_i, \Sigma_i, \boldsymbol{c}_i\} = g_{\Theta}^{(A)}(\mathbf{f}_i), 
    \{\} = \boldsymbol{O}^{(A)}
\end{align}

\subsection{Warm Start}
\label{sec:warm_start}
We find that directly decoding 3D structures from 2D image features can easily stuck in local minimal due to the sparse nature of casual images.
To ensure robust evaluation of features from diverse foundation models, we warm up our optimization using a point cloud regression:
\begin{align}
    \label{eqn:warm_start}
    \min_{\Theta} & \left\Vert g_{\Theta}(\mathbf{f}) - \boldsymbol{G}_{init} \right\Vert
\end{align}
where $\boldsymbol{G}_{init}$ refers to the initialization point cloud comes from DUSt3R~\cite{DUSt3R}.

\subsection{Evaluation}
\label{sec:evaluation}
We choose to evaluate on NVS from casual (sparse and uncalibrated) images~\cite{InstantSplat} for two main reasons: 
(1) Diversity. The capability of handling casual images helps diversify the evaluation data by lowering the requirements for acquisition techniques and view setups.
(2) Discrepancy. This task poses more of a challenge compared to dense-view NVS, making it better to differentiate the performance of various VFM features.
To enable our evaluation to cover arbitrary casual capturing from 3 to N views, we uniformly estimate the camera parameters of both training and test views across all datasets via unconstrained stereo reconstructor.
Subsequently, we perform test-time pose optimization~\cite{Nope-nerf, cf3dgs, Barf, nerfmm, InstantSplat} via photometric loss to further refine the test poses before evaluating view synthesis quality.

\begin{table}[t!]
    \centering
    \setlength\tabcolsep{2pt}
    \scriptsize
    \begin{tabularx}{\columnwidth}{lcccl}
  \toprule
  \textbf{VFM} & \textbf{Arch.} & \textbf{Channel} & \textbf{Supervision} & \textbf{Dataset} \\
  \midrule
  DUSt3R~\cite{DUSt3R}  & ViT-L/16      & 1024      & Point Regression              & \highlightboxyellow{3D} DUSt3R-Mix    \\
  MASt3R~\cite{MASt3R}  & ViT-L/16      & 1024      & Point Regression              & \highlightboxyellow{3D} MASt3R-Mix    \\
  MiDaS~\cite{MiDaS}    & ViT-L/16      & 1024      & Depth Regression              & \highlightboxyellow{3D} MiDaS-Mix     \\
  DINOv2~\cite{DINOv2}  & ViT-B/14      & 768       & Self Distillation             & \highlightboxgreen{2D} LVD-142M      \\
  DINO~\cite{DINO}      & ViT-B/16      & 768       & Self Distillation             & \highlightboxgreen{2D} ImageNet-1k   \\
  SAM~\cite{SAM}        & ViT-B/16      & 768       & Segmentation                  & \highlightboxgreen{2D} SA-1B             \\
  CLIP~\cite{CLIP}      & ViT-B/16      & 512       & Contrastive VLM               & \highlightboxgreen{2D} WIT-400M          \\
  RADIO~\cite{RADIO}    & ViT-H/16      & 1280      & Multi-teacher Distillation    & \highlightboxgreen{2D} DataComp-1B       \\
  MAE~\cite{MAE}        & ViT-B/16      & 768       & Image Reconstruction          & \highlightboxgreen{2D} ImageNet-1k       \\
  SD~\cite{SD}          & UNet          & 1280      & Denoising VLM                 & \highlightboxgreen{2D} LAION             \\
  \bottomrule
  \end{tabularx}
  \vspace{-6pt}
  \caption{
    \textbf{VFMs for Evaluation.} For fair comparison, we use checkpoints with comparable architectures and training scales, unify the feature channel dimensions via PCA, and maintain a consistent probing network architecture for all VFMs.}
    \label{tab:VFM}
    \vspace{-10pt}
  \end{table}

\section{Experiments}
\label{sec:exp}

\subsection{Experimental Setup}
\noindent\textbf{Features.} 
We focus our experiments on 10 VFMs that show strong potential for generalizable 3D awareness, comparing models trained on different data types (2D \textit{vs.} 3D) and supervision strategies (\eg, supervised \textit{vs.} self-supervised, point \textit{vs.} depth). An overview is provided in \cref{tab:VFM}, with more details in \suppl To make the comparison as fair as possible, we use publicly available checkpoints and select those with comparable architectures and training scales. We also incorporate IUVRGB, comprising image index (I), pixel coordinates (UV), and colors (RGB), as a baseline.

\begin{table}[t!]
\centering
\setlength\tabcolsep{4pt}
\scriptsize
\begin{tabularx}{\columnwidth}{lcccc}
\toprule
\textbf{Dataset} & \textbf{Scene Type} & \textbf{Complexity} & \textbf{View Range} & \textbf{Views} \\
\midrule
LLFF~\cite{llff}                     & Indoor                  & Simple           & Small                    & 2             \\
DTU~\cite{DTU}                       & Indoor Object           & Simple           & Small                    & 3             \\
DL3DV~\cite{ling2024dl3dv}           & Indoor / Outdoor        & Moderate         & Medium                   & 5-6           \\
Casual                               & Daily Scenario          & Moderate         & Medium                   & 4-7           \\
MipNeRF360~\cite{Mip_nerf_360}       & Unbounded               & Moderate         & 360                      & 6             \\
MVimgNet~\cite{Mvimgnet}             & Outdoor Object          & Moderate         & 180-360                  & 2-4           \\
T\&T~\cite{Tanks}                    & Indoor / Outdoor        & High             & Large                    & 6             \\
\bottomrule
\end{tabularx}
\vspace{-6pt}
\caption{{\bf Datasets for Evaluation.} Classified by scene type, complexity, viewpoint variation, and sampled views.}
\label{tab:dataset_comparison}
\vspace{-14pt}
\end{table}

\pheading{Datasets.} To reliably evaluate different features, our experiments utilize seven multi-view datasets, with sparse views sampled spanning from 2 to 7, and test viewpoints far from the training viewpoints. These datasets, as shown in \cref{tab:dataset_comparison}, rich in diversity, provide us with a more comprehensive perspective compared to datasets with 3D \gt .

\pheading{Metrics.} We evaluate novel view synthesis across seven datasets using standard metrics: PSNR, SSIM, and LPIPS. For metric calculation, we follow Splatt3R~\cite{Splatt3R} by applying masks to both the rendered and test images. These masks define valid pixels as those inside the frustum of at least one view and with reprojected depths aligned with DUST3R predicted depth. All metrics are computed over the entire image. On the DTU dataset, we measure the distance between reconstructed 3DGS and point cloud ground truth (\cref{tab:DTU}), reporting average accuracy, completeness, and distance, as in prior works~\cite{DTU,spann3r}. Accuracy is the smallest Euclidean distance from a reconstructed point to ground truth, and completeness is the smallest Euclidean distance from a ground-truth point to the reconstruction. Distance is the Euclidean distance based on ground-truth point matching.

\begin{figure*}[t!]
    \centering
    \includegraphics[trim={0 0 0 0}, clip, width=1\linewidth,page=1]{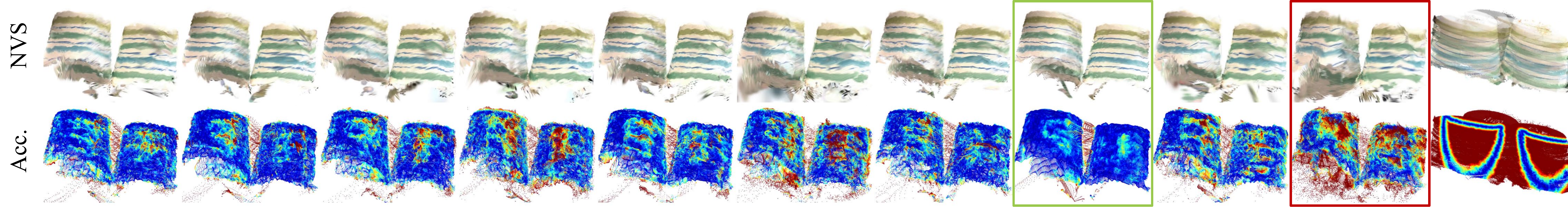}
    \scriptsize
    \setlength{\tabcolsep}{0.00pt}     %
    \begin{tabularx}{\linewidth}{
      >{\centering\arraybackslash}m{0.025\linewidth}
      >{\centering\arraybackslash}m{0.08864\linewidth}
      >{\centering\arraybackslash}m{0.08864\linewidth}
      >{\centering\arraybackslash}m{0.08864\linewidth}
      >{\centering\arraybackslash}m{0.08864\linewidth}
      >{\centering\arraybackslash}m{0.08864\linewidth}
      >{\centering\arraybackslash}m{0.08864\linewidth}
      >{\centering\arraybackslash}m{0.08864\linewidth}
      >{\centering\arraybackslash}m{0.08864\linewidth}
      >{\centering\arraybackslash}m{0.08864\linewidth}
      >{\centering\arraybackslash}m{0.08864\linewidth}
      >{\centering\arraybackslash}m{0.08864\linewidth}
      }
      \quad & \dustr~\cite{DUSt3R} & \mastr~\cite{MASt3R} & \midas~\cite{MiDaS} & \dinotwo~\cite{DINOv2} & \dino~\cite{DINO} & \sam~\cite{SAM} & CLIP~\cite{CLIP} & \radio~\cite{RADIO} & \mae~\cite{MAE} & SD~\cite{SD} & IUVRGB
   \end{tabularx}
    \vspace{-12pt}
    \caption{{\bf Novel View Synthesis as Proxy Task to Assess 3D.} We present qualitative examples from the DTU dataset, including NVS and Accuracy.
    Results show that NVS quality aligns with 3D metrics, proving its reliability as an indicator for 3D assessment.
    RADIO performs \textcolor{bestgreen}{best}, SD \textcolor{worstred}{worst}, with IUVRGB as a reference.
    \faSearch~\textbf{Zoom in} or check our \dtu to see more details.
    }
    \label{fig:metric3d}
    \vspace{-16pt}
  \end{figure*}

\subsection{Motivation Validation}
\label{sec:validation}
\vspace{-5pt}
\pheading{Novel View Synthesis Correlates with 3D Metrics.}
Using 2D metrics instead of 3D ones allows us to bypass the need for 3D \gt. Motivated by this, we propose \featgs to assess the 3D awareness of VFMs through the 2D metric of novel view synthesis (NVS).
The key question is: \textit{Can novel view synthesis effectively serve as a proxy for 3D metrics?} We posit that high-quality NVS strongly correlates with an accurate 3D representation. To validate this hypothesis, we conduct experiments on the DTU dataset~\cite{DTU} with dense pointclouds as 3D \gt, evaluating both the 2D NVS and the 3D point cloud regression tasks. We then calculated the correlation between these results, as shown in \cref{tab:DTU}. The results reveal a strong correlation between 2D and 3D metrics, supporting NVS as an indicator for 3D assessment.
We further qualitatively demonstrate this correlation in \cref{fig:metric3d} (See more details in \suppl's \cref{fig:supp_metric3d}). The results indicate a strong relationship between NVS and 3D metrics, confirming that high-quality NVS aligns closely with accurate 3D representations.

\begin{table}[t]
    \centering
    \begin{minipage}{0.58\columnwidth}
      \setlength{\tabcolsep}{2pt}
    \resizebox{\columnwidth}{!}{
    \begin{tabular}{l|ccc|ccc}
    \toprule
    \multicolumn{1}{c|}{} & \multicolumn{3}{c|}{2D Metrics} & \multicolumn{3}{c}{3D Metrics} \\
    \midrule
    Feature & \fontsize{8.5pt}{9pt}\selectfont{PSNR$\uparrow$} & \fontsize{8.5pt}{9pt}\selectfont{SSIM$\uparrow$} & \fontsize{8.5pt}{9pt}\selectfont{LPIPS$\downarrow$} & \fontsize{8.5pt}{9pt}\selectfont{Acc.$\downarrow$} & \fontsize{8.5pt}{9pt}\selectfont{Comp.$\downarrow$} & \fontsize{8.5pt}{9pt}\selectfont{Dist.$\downarrow$} \\
    \midrule
    DUSt3R  &             \cellcolor[rgb]{0.83,0.87,0.64}21.36 &             \cellcolor[rgb]{0.83,0.87,0.64}.7772 &                \cellcolor[rgb]{0.83,0.87,0.64}.2195 &               \cellcolor[rgb]{0.83,0.87,0.64}2.439 &                \cellcolor[rgb]{0.74,0.83,0.55}1.316 &                \cellcolor[rgb]{0.91,0.90,0.73}6.955 \\
    MASt3R  &             \cellcolor[rgb]{0.65,0.79,0.46}21.44 &             \cellcolor[rgb]{0.65,0.79,0.46}.7792 &                \cellcolor[rgb]{0.65,0.79,0.46}.2177 &               \cellcolor[rgb]{0.65,0.79,0.46}2.321 &                \cellcolor[rgb]{0.65,0.79,0.46}1.286 &                \cellcolor[rgb]{0.65,0.79,0.46}6.557 \\
    MiDaS   &             \cellcolor[rgb]{1.00,0.94,0.82}21.09 &             \cellcolor[rgb]{1.00,0.94,0.82}.7712 &                \cellcolor[rgb]{1.00,0.94,0.82}.2254 &               \cellcolor[rgb]{1.00,0.94,0.82}2.934 &                \cellcolor[rgb]{1.00,0.87,0.82}1.412 &                \cellcolor[rgb]{1.00,0.94,0.82}8.230 \\
    DINOv2  &             \cellcolor[rgb]{1.00,0.92,0.82}21.01 &             \cellcolor[rgb]{1.00,0.92,0.82}.7695 &                \cellcolor[rgb]{1.00,0.92,0.82}.2277 &               \cellcolor[rgb]{1.00,0.89,0.82}3.101 &                \cellcolor[rgb]{1.00,0.94,0.82}1.337 &                \cellcolor[rgb]{1.00,0.89,0.82}8.588 \\
    DINO    &             \cellcolor[rgb]{0.74,0.83,0.55}21.40 &             \cellcolor[rgb]{0.74,0.83,0.55}.7783 &                \cellcolor[rgb]{0.74,0.83,0.55}.2187 &               \cellcolor[rgb]{0.91,0.90,0.73}2.440 &                \cellcolor[rgb]{0.74,0.83,0.55}1.316 &                \cellcolor[rgb]{0.83,0.87,0.64}6.885 \\
    SAM     &             \cellcolor[rgb]{1.00,0.87,0.82}20.93 &             \cellcolor[rgb]{1.00,0.87,0.82}.7660 &                \cellcolor[rgb]{1.00,0.87,0.82}.2304 &               \cellcolor[rgb]{1.00,0.87,0.82}3.176 &                \cellcolor[rgb]{1.00,0.89,0.82}1.339 &                \cellcolor[rgb]{1.00,0.87,0.82}8.785 \\
    CLIP    &             \cellcolor[rgb]{0.91,0.90,0.73}21.26 &             \cellcolor[rgb]{0.91,0.90,0.73}.7752 &                \cellcolor[rgb]{0.91,0.90,0.73}.2215 &               \cellcolor[rgb]{0.74,0.83,0.55}2.357 &                \cellcolor[rgb]{0.56,0.75,0.38}1.209 &                \cellcolor[rgb]{0.74,0.83,0.55}6.739 \\
    RADIO   &             \cellcolor[rgb]{0.56,0.75,0.38}21.78 &             \cellcolor[rgb]{0.56,0.75,0.38}.7871 &                \cellcolor[rgb]{0.56,0.75,0.38}.2042 &               \cellcolor[rgb]{0.56,0.75,0.38}1.886 &                \cellcolor[rgb]{0.91,0.90,0.73}1.326 &                \cellcolor[rgb]{0.56,0.75,0.38}5.431 \\
    MAE     &             \cellcolor[rgb]{1.00,0.89,0.82}20.96 &             \cellcolor[rgb]{1.00,0.89,0.82}.7666 &                \cellcolor[rgb]{1.00,0.89,0.82}.2289 &               \cellcolor[rgb]{1.00,0.92,0.82}2.963 &                \cellcolor[rgb]{1.00,0.94,0.82}1.337 &                \cellcolor[rgb]{1.00,0.92,0.82}8.374 \\
    SD      &             \cellcolor[rgb]{1.00,0.84,0.82}20.76 &             \cellcolor[rgb]{1.00,0.84,0.82}.7638 &                \cellcolor[rgb]{1.00,0.84,0.82}.2343 &               \cellcolor[rgb]{1.00,0.84,0.82}4.334 &                \cellcolor[rgb]{1.00,0.84,0.82}1.603 &               \cellcolor[rgb]{1.00,0.84,0.82}11.594 \\
    IUVRGB  &             \cellcolor[rgb]{1.00,0.82,0.82}16.09 &             \cellcolor[rgb]{1.00,0.82,0.82}.6825 &                \cellcolor[rgb]{1.00,0.82,0.82}.3134 &              \cellcolor[rgb]{1.00,0.82,0.82}13.015 &               \cellcolor[rgb]{1.00,0.82,0.82}16.957 &               \cellcolor[rgb]{1.00,0.82,0.82}46.671 \\
    \bottomrule
    \end{tabular}
    }
    \vspace{-8pt}
    \caption*{\footnotesize{(a) 2D Metrics \textit{vs.} 3D Metrics}}
    \vspace{-8pt}
    \label{tab:corr_DTU}
  \end{minipage}
  \begin{minipage}{0.41\columnwidth}
    \centering
    \includegraphics[width=\columnwidth]{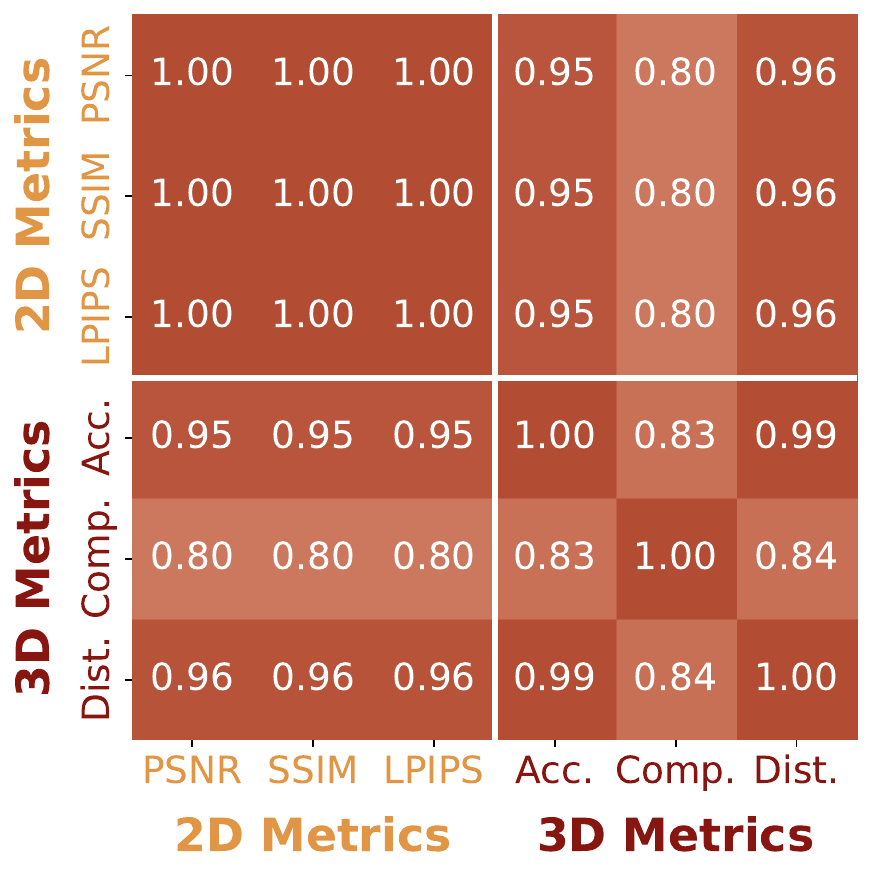}
    \vspace{-16pt}
    \caption*{\footnotesize{(b) Correlation Matrix}}
    \vspace{-8pt}
    \label{fig:corr_DTU}
  \end{minipage}
  \caption{{\bf Novel View Synthesis Aligns Well with 3D Metrics.} (a) We report NVS quality and the Euclidean distance between reconstructed 3DGS positions and pointcloud ground truth on the DTU dataset. (b) Strong 2D-3D metric correlation supports NVS as a benchmark for 3D assessment.}
  \label{tab:DTU}
  \vspace{-16pt}
\end{table}

\pheading{Data Diversity Matters for Comprehensive Probing.} 
Testing on small-scale data can lead to biased conclusions. 
As shown in \cref{tab:GTA}, the evaluation results vary across probing GTA modes and different datasets.
For instance, LLFF is relatively simple for novel view synthesis due to its dense view capturing and small scale. \mastr, \dustr, and \dino show superior geometry results on LLFF. However, none of them ranks higher than \radio on T\&T dataset, which features more challenging scenes.
We sort the datasets from easy to hard in ~\cref{tab:GTA} to derive insights from the variations.
Additionally, we show performance correlations in \suppl's \cref{fig:correlation_dataset}. 
We observe that no method consistently performs well across all datasets when the evaluation set is large. 
Dataset evaluation bias is inevitable. By removing the need for 3D ground truth, we can evaluate on large-scale diverse captures, thereby ensuring that the results are 
much less biased.
Therefore, we base our findings on overall performance and discuss outliers separately.

\subsection{Findings}
\label{sec:results}

\begin{figure*}[t!]
  \centering
  \includegraphics[trim={0 2cm 0 0}, clip, width=1\linewidth,page=1]{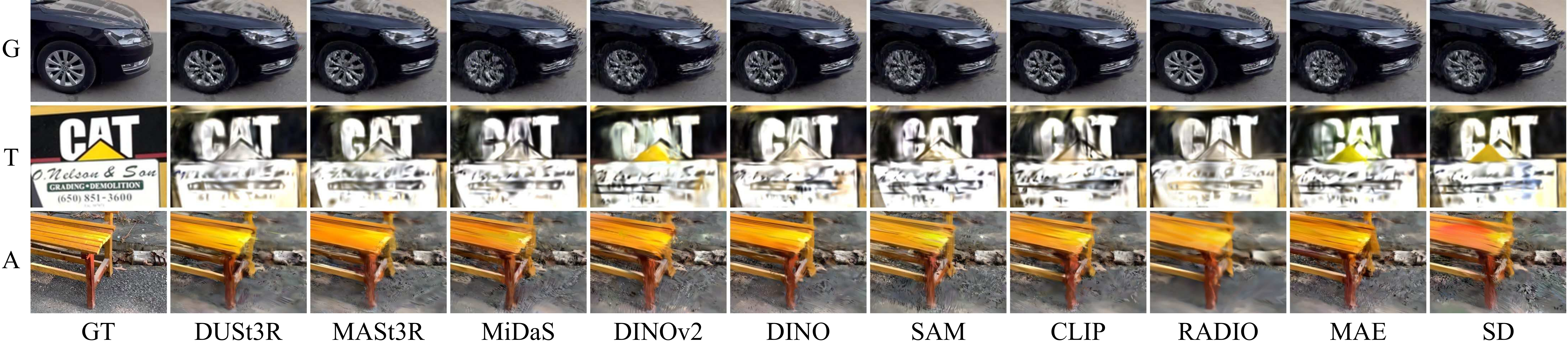}
  \scriptsize
  \begin{tabularx}{\linewidth}{
 >{\centering\arraybackslash}X
 @{\hskip 20pt}
 >{\centering\arraybackslash}X
 >{\centering\arraybackslash}X
  @{\hskip 20pt}
 >{\centering\arraybackslash}X
 >{\centering\arraybackslash}X
   @{\hskip 20pt}
 >{\centering\arraybackslash}X
 >{\centering\arraybackslash}X
 >{\centering\arraybackslash}X
 >{\centering\arraybackslash}X
 >{\centering\arraybackslash}X
 >{\centering\arraybackslash}X
}
\qquad GT & \dustr~\cite{DUSt3R} & \mastr~\cite{MASt3R} & \midas~\cite{MiDaS} & \dinotwo~\cite{DINOv2} & \dino~\cite{DINO} & \sam~\cite{SAM} & CLIP~\cite{CLIP} & \radio~\cite{RADIO} & \mae~\cite{MAE} & SD~\cite{SD} 
 \end{tabularx}
 \vspace{-15pt}
  \caption{{\bf Qualitative Examples.} We compare novel view renderings across VFM features. In \gmode, 
  the multi-teacher-distillation method (RADIO) and point-regression-based methods (MASt3R, DUSt3R) 
  produce more plausible geometry, \eg, vehicle front and the wheel, indicating better multi-view consistency. 
  All VFM features struggle in \tmode, and renderings in the \amode are notably blurred, both reflecting the limited texture awareness of current VFMs.
  }
  \label{fig:GTA}
  \vspace{-16pt}
\end{figure*}

\begin{figure}[t!]
  \centering
  \begin{subfigure}[b]{1\linewidth}
    \centering
    \includegraphics[width=\linewidth,page=1]{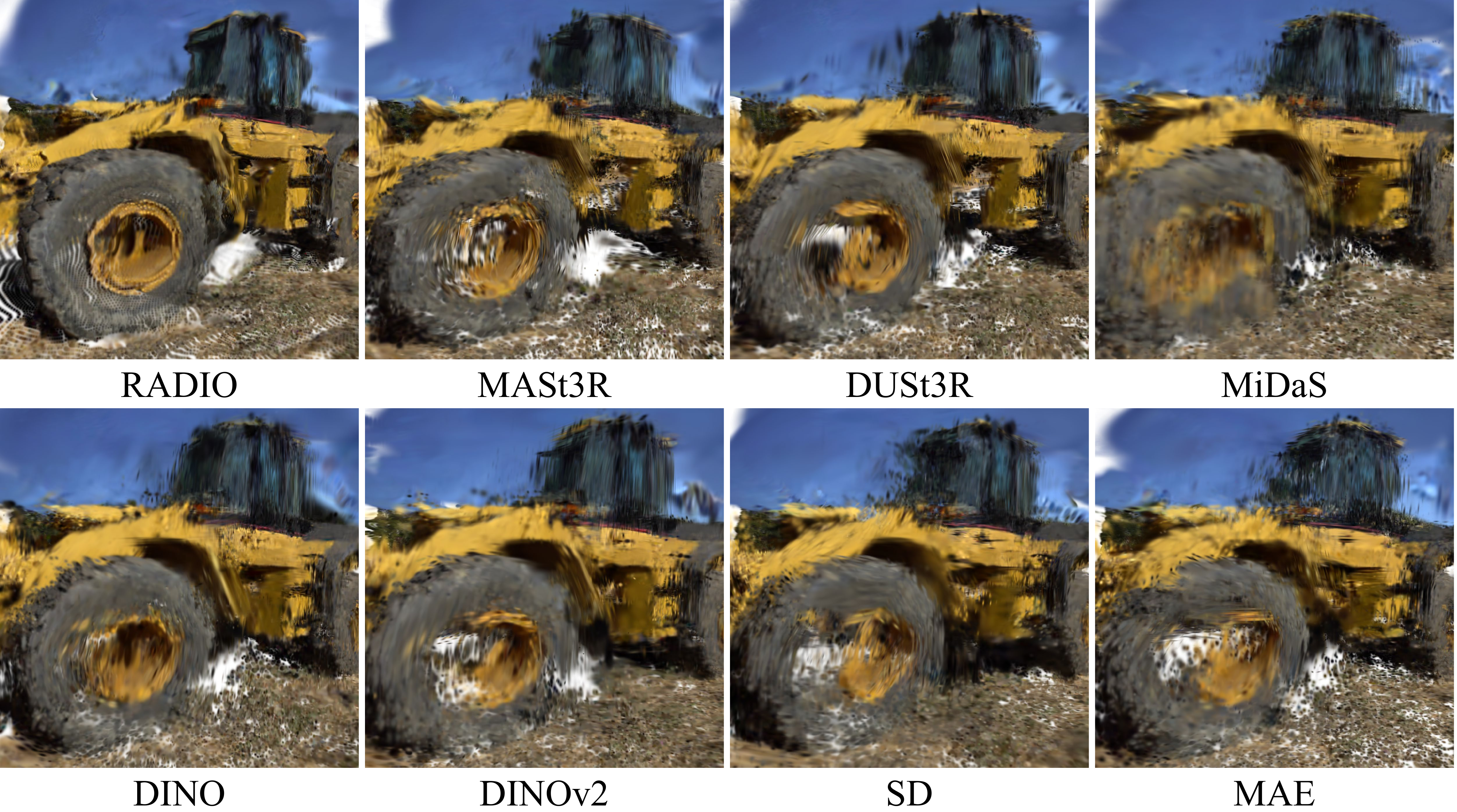}
    \caption{NVS Comparison on Geometry Awareness}
    \label{fig:Geometry}
  \end{subfigure}
  \begin{subfigure}[b]{1\linewidth}
    \centering
    \includegraphics[width=\linewidth,page=1]{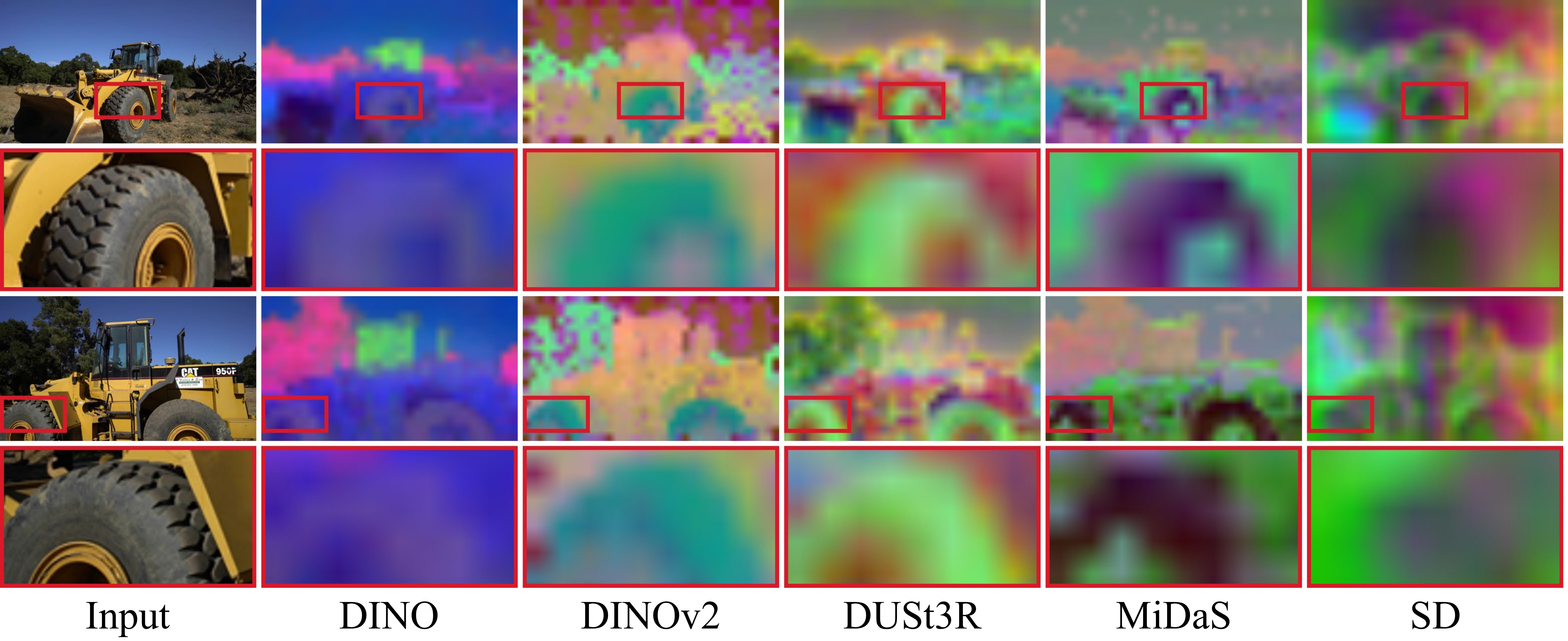}
    \caption{VFM Features from Training Views}
    \label{fig:features}
  \end{subfigure}
  \vspace{-20pt}
  \caption{{\bf Feature Consistency Reflected in NVS.}
  (a) RADIO, MASt3R, DUSt3R, and DINO effectively capture geometry. 
  (b) DINO is consistent across training views, but PE artifacts appear in DINOv2. 3D data proves beneficial, as DUSt3R is consistent; however, MiDaS shows inconsistency, suggesting that pointmap representation is more reasonable than depth. SD also exhibits inconsistency. These inconsistencies lead to poor NVS.
  }
  \label{fig:gmode}
  \vspace{-16pt}
\end{figure}

\noindent\textbf{Overall Performance.}
\Cref{tab:GTA} benchmarks VFM features with three probing modes: \geo, \tex and \all. The mean scores across diverse datasets are plotted in~\cref{fig:spider-chart}.
The top three performers in \gmode are \radio $>$ \mastr $>$ \dustr. However, they show significantly different rating in \tmode, with \mae\ $>$ \sam $>$ \mastr. In the \amode, \mastr and \dustr achieve the best score, followed by \dino.
In~\cref{tab:GTA}, Stable Diffusion (SD) performs the worst in most metrics, \cref{fig:GTA}-A shows its significant color drift and broken geometry, check \suppl for more qualitative results of geometry. 
This aligns with the conclusion about SD in Probe3D~\cite{Probe3D}. Large viewpoint changes cause inconsistency in the feature space (see \cref{fig:features}).
In the following sections, we provide a comprehensive analysis of the insights behind the above ratings.

\begin{figure}[t]
  \centering
  \includegraphics[width=1\linewidth,page=1]{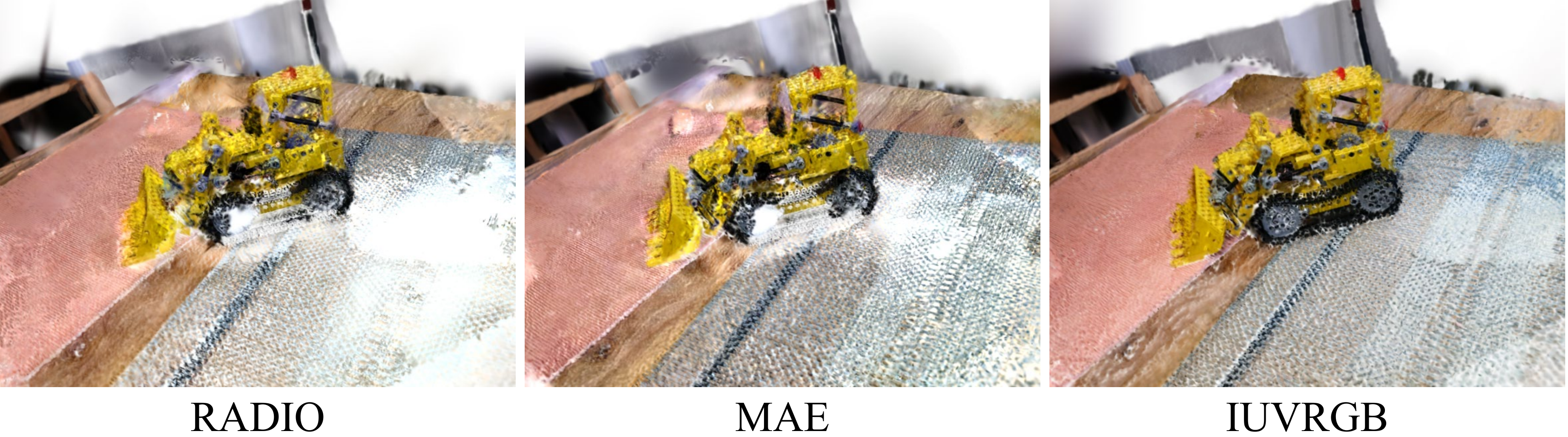}
  \vspace{-18pt}
  \caption{{\bf Texture Blurriness Comparison.} MAE preserves sharper texture over RADIO. IUVRGB is shown for reference.
  }
  \label{fig:texture}
  \vspace{-12pt}
\end{figure}

\begin{figure}[t]
  \centering
  \includegraphics[width=1\linewidth,page=1]{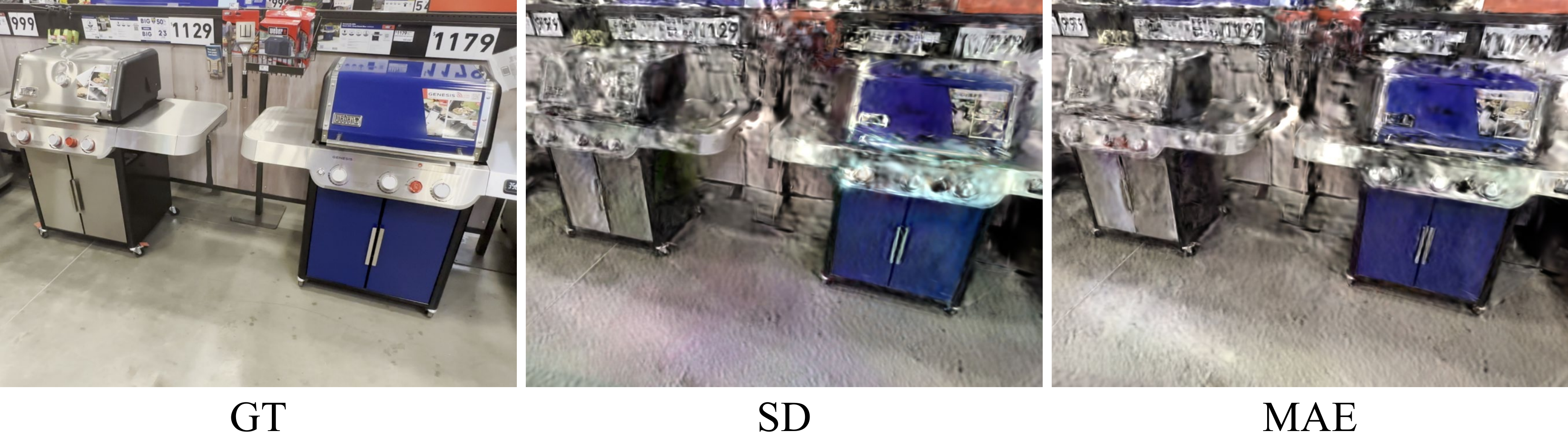}
  \vspace{-18pt}
  \caption{
  {\bf MAE \textit{vs.} SD on Texture Awareness.} 
  While both MAE and SD are trained to reconstruct images (MAE in pixel space with an MSE loss and SD in feature space with a denoising loss), SD tends to result in a significant color shift.
  }
  \label{fig:sd_shift}
  \vspace{-18pt}
\end{figure}

\setlength{\tabcolsep}{8pt}
\begin{table*}[t!]
\setlength{\tabcolsep}{4pt}
\centering
\vspace{8mm}
\resizebox{\textwidth}{!}{
\begin{tabular}{l|>{\raggedleft\arraybackslash}p{0.85cm}>{\raggedleft\arraybackslash}p{0.85cm}>{\raggedleft\arraybackslash}p{0.85cm}|>{\raggedleft\arraybackslash}p{0.85cm}>{\raggedleft\arraybackslash}p{0.85cm}>{\raggedleft\arraybackslash}p{0.85cm}|>{\raggedleft\arraybackslash}p{0.85cm}>{\raggedleft\arraybackslash}p{0.85cm}>{\raggedleft\arraybackslash}p{0.85cm}|>{\raggedleft\arraybackslash}p{0.85cm}>{\raggedleft\arraybackslash}p{0.85cm}>{\raggedleft\arraybackslash}p{0.85cm}|>{\raggedleft\arraybackslash}p{0.85cm}>{\raggedleft\arraybackslash}p{0.85cm}>{\raggedleft\arraybackslash}p{0.85cm}|>{\raggedleft\arraybackslash}p{0.85cm}>{\raggedleft\arraybackslash}p{0.85cm}>{\raggedleft\arraybackslash}p{0.85cm}|>{\raggedleft\arraybackslash}p{0.85cm}>{\raggedleft\arraybackslash}p{0.85cm}>{\raggedleft\arraybackslash}p{0.85cm}|>{\raggedleft\arraybackslash}p{0.85cm}>{\raggedleft\arraybackslash}p{0.85cm}>{\raggedleft\arraybackslash}p{0.85cm}|>{\raggedleft\arraybackslash}p{0.85cm}>{\raggedleft\arraybackslash}p{0.85cm}>{\raggedleft\arraybackslash}p{0.85cm}}
\toprule
\multicolumn{1}{c|}{} & \multicolumn{9}{c|}{LLFF} & \multicolumn{9}{c|}{DL3DV} & \multicolumn{9}{c}{Casual} \\
\midrule
\multicolumn{1}{c|}{} & \multicolumn{3}{c|}{\textbf{G}eometry} & \multicolumn{3}{c|}{\textbf{T}exture} & \multicolumn{3}{c|}{\textbf{A}ll} & \multicolumn{3}{c|}{\textbf{G}eometry} & \multicolumn{3}{c|}{\textbf{T}exture} & \multicolumn{3}{c|}{\textbf{A}ll} & \multicolumn{3}{c|}{\textbf{G}eometry} & \multicolumn{3}{c|}{\textbf{T}exture} & \multicolumn{3}{c}{\textbf{A}ll} \\
\bottomrule
\end{tabular}
}
\vspace{-10pt}
\caption{{\bf Quantitative Results.} We evaluate geometry and texture awareness of VFMs on NVS using \geo, \tex, and \all probing modes. Results indicate that VFM performance varies across datasets, highlighting the importance of dataset diversity. The lack of texture awareness in VFMs limits both \tmode and \amode, especially in LPIPS. Performance is ranked by color, from \BGcolor{ffd0d0}{w}\BGcolor{ffd5d0}{o}\BGcolor{ffdad0}{r}\BGcolor{ffdfd0}{s}\BGcolor{ffe5d0}{t}\BGcolor{ffead0}{ }\BGcolor{ffefd0}{t}\BGcolor{ece8bd}{o}\BGcolor{dadfaa}{ }\BGcolor{c7d797}{b}\BGcolor{b5d085}{e}\BGcolor{a2c772}{s}\BGcolor{90c05f}{t}.}
\label{tab:GTA}
\vspace{-10pt}
\end{table*}

\pheading{Texture-unfriendly Training Strategies.}
As shown in \cref{tab:GTA} and \cref{fig:GTA}, VFM features perform poorly in \tmode, even worse than the simple IUVRGB encoding shown in \cref{fig:texture}.
It suggests that current VFM features lack texture awareness,
as noted in~\cite{NoPoSplat,Splatt3R}.
One likely explanation for this is that VFMs are often trained for semantic understanding or 3D estimation, which require texture-invariant features to avoid shortcuts~\cite{biased,Shortcut}.
For example, \dustr is trained to be texture-invariant for better 3D robustness on diverse \itw captures. Heavy data augmentations in SSL (\ie, DINO~\cite{DINO}, BYOL~\cite{grill2020bootstrap}, SimCLR~\cite{pmlr-v119-chen20j}), such as color jittering, Gaussian blur, and solarization, encourage the model to produce consistent outputs despite changes in appearance or lightning.
Since CLIP is trained on weakly aligned image-text pairs, it often includes ambiguous and coarse semantics that are not discriminative enough to model low-level visual patterns, like colors, materials, and textures~\cite{wang2024instantid}. 
\radio distills \dino and \clip, achieving excellent geometry awareness, but also inherits their poor texture awareness (see \cref{fig:spider-chart} and \cref{fig:texture}).

\pheading{Texture Benefits from Masked Image Reconstruction.}
\Cref{tab:GTA} shows that the \amode is impeded by \tmode, leading to worse performance in LPIPS (by an average of $+0.05$) than \gmode, which does not use VFM features for 3DGS color regression. 
Terrible texture awareness prevents \radio from being versatile, as \cref{fig:spider-chart} shows. Visually, as displayed in~\cref{fig:GTA}, novel view renderings in the \amode tends to appear blurred. \Cref{fig:GTA_same} also shows that \tmode, which excludes VFM features for 3DGS geometry regression, exhibits broken structures, while \amode appears more blurred than the same regions in \gmode because \amode relies on VFM features for 3DGS color regression, unlike \gmode, which freely optimizes colors.
To further analyze the mutual correlation of GTA modes, we compute their correlation matrix using average metrics across all datasets, as shown in \suppl's ~\cref{fig:correlation}. 
Results indicate that the \amode is more strongly correlated with \gmode in PSNR and SSIM, which primarily reflect structural consistency, but is more closely related to \tmode in LPIPS, a metric used to evaluate image sharpness. This further supports the notion that the blurriness observed in the \amode stems from a lack of texture awareness in VFMs.
Texture is obviously crucial for photorealism.
How can it be retained in VFMs?
As illustrated in \cref{fig:spider-chart}, VFMs with masked-image-reconstruction pre-training (\ie, \mae, \mastr, \dustr) rank top in T-LPIPS, and \cref{fig:texture} backs this. MAE's ability to recover sharp textures might be attributed to using only \textit{cropping-only} augmentation. Color jittering degrades results, so it’s not included~\cite{MAE}. 
Additionally, denoise-based image reconstruction leads to  color shift, as shown in \cref{fig:sd_shift}.

\pheading{Geometry Benefits from 3D Data.} 
In \cref{fig:spider-chart}, \radio, \mastr, \dustr and \dino rank among the top-4 in geometry awareness metrics.
In \cref{fig:Geometry}, these four features
help reconstruct a more complete structure of the excavator, whereas others result in floating artifacts and distortion. 
Better geometry awareness implies stronger cross-view consistency, which is also supported by \cref{fig:features}. 
\textit{What is the key ingredient to achieve geometry awareness? One crucial factor is 3D data.} Both of MASt3R and DUSt3R are trained with pointmap. What about 2.5D data, like depth or normal maps? It is much worse, see \dustr \textit{vs.} \midas at~\cref{fig:Geometry}. Please note that, \midas and \dustr shares the same ViT-L/16 encoder architecture (see \cref{tab:VFM}) and comparable training scales (3M \textit{vs.} 2M).
The depth map estimation may cause inconsistent features for the same object when viewed from different distances.
In contrast, pointmap regression~\cite{DUSt3R} encourages the network to generate consistent features across views, as the scene coordinates remain unchanged when the view changes~\cite{acezero,shotton2013scene}.

\begin{figure}[t!]
  \centering
  \begin{subfigure}[b]{1\linewidth}
    \centering
    \vspace{-5pt}
    \includegraphics[width=\linewidth,page=1]{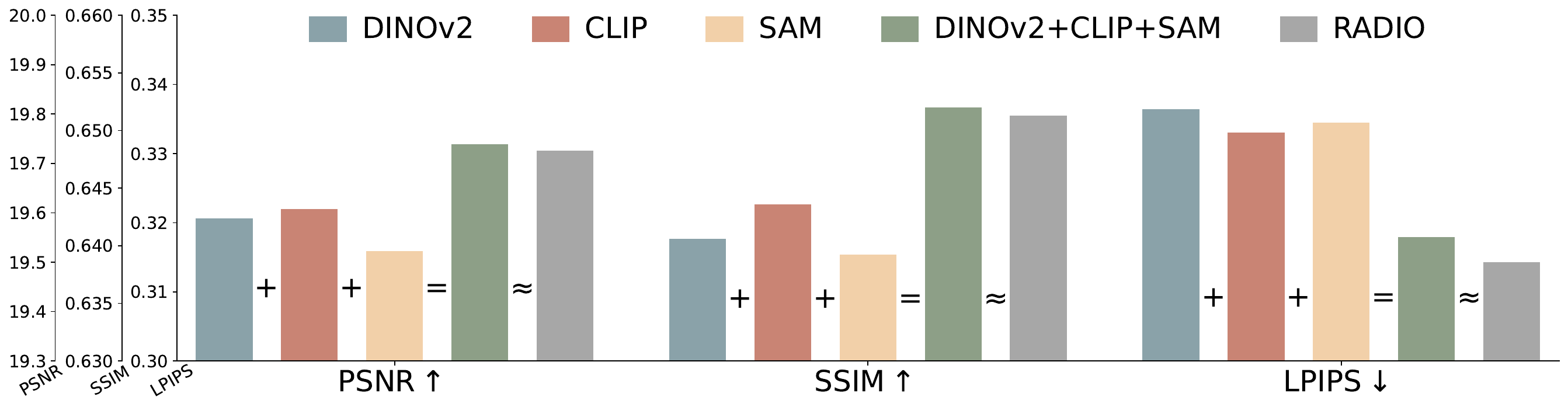}
    \caption{Concatenated Features \textit{vs.} RADIO 
    }
    \label{fig:cat_vs_radio}
  \end{subfigure}
  \begin{subfigure}[b]{1\linewidth}
    \centering
    \includegraphics[width=\linewidth,page=1]{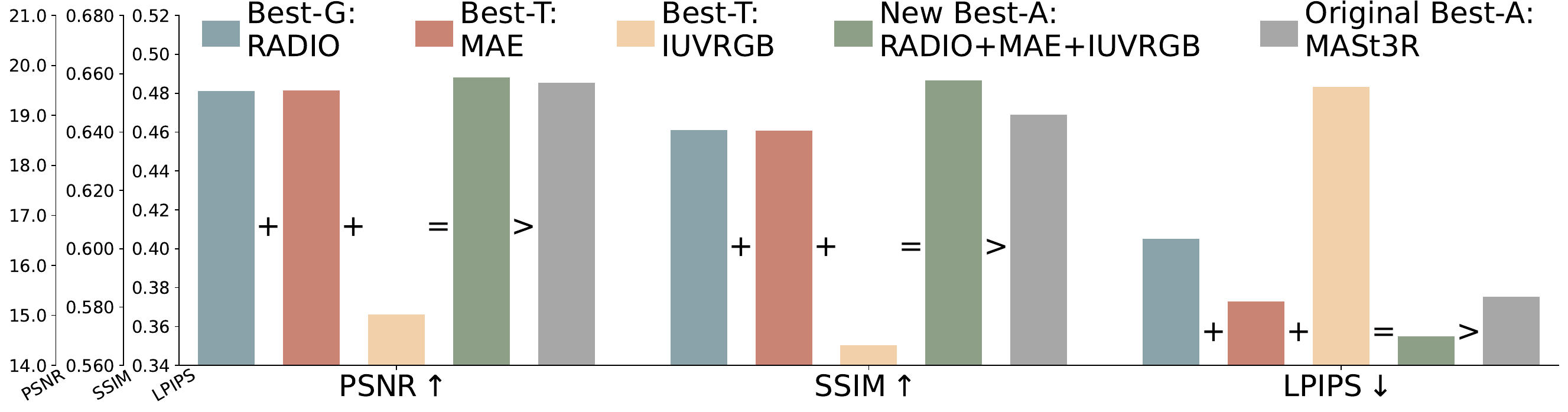}
    \caption{Concatenated Features \textit{vs.} MASt3R}
    \label{fig:cat_vs_mast3r}
  \end{subfigure}
  \vspace{-20pt}
  \caption{{\bf Feature Concatenation.} (a) RADIO, distilling DINOv2, CLIP, and SAM, achieves superior geometry awareness. \gmode with concatenated features from the three yields results comparable to RADIO. (b) The \amode using concatenated VFM features from best \gmode and \tmode, outperforms the original best \amode. \(\pmb{+}\): Feature concatenation; \(\pmb{\approx}\): Comparable performance; \(\pmb{>}\): Superior performance.}
  \label{fig:cat_feats}
  \vspace{-16pt}
\end{figure}

\pheading{Model Ensembling Help.}
RADIO, distilling DINOv2, CLIP and SAM into a single model, achieves the best geometry awareness, as shown in \cref{fig:spider-chart,fig:GTA,fig:Geometry}.
A natural question arises: \textit{Could simply concatenating these features yield comparable results? Yes!} 
Specifically, we concatenate features of DINOv2, CLIP and SAM, and then apply PCA to reduce feature channels to 256, keeping 
the size of the network unchanged for a fair comparison.
\Cref{fig:cat_vs_radio} shows that, in \gmode, feature concatenation (DINOv2+CLIP+SAM) outperforms model distillation (\radio).
This inspires us to further explore: \textit{What if we combine the best \gmode feature and the best \tmode feature?} 
\suppl's \cref{fig:correlation} has indicated that the optimal \amode should have no weakness in either texture or geometry.
As shown in~\cref{fig:cat_vs_mast3r}, the \amode using concatenated features from RADIO (best in \gmode) with MAE and IUVRGB (best in \tmode), outperforms the original best \amode with \mastr features.
This exploration shows the potential of our probing method.

\section{Application}
\label{sec:application}

\begin{figure*}[t!]
  \centering
  \includegraphics[width=1\linewidth,page=1]{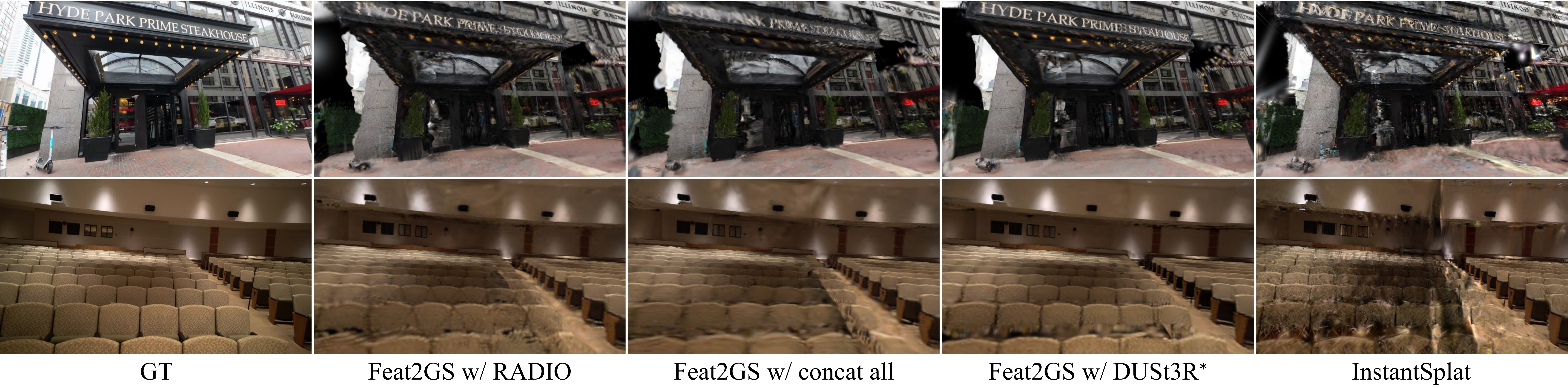}
  \vspace{-20pt}
  \caption{{\bf NVS from Casual (Sparse and Uncalibrated) Images.} We compare 
  our \gmode baselines
  against InstantSplat, which overfits to training views and results in broken structures. \featgs with RADIO produces more consistent results due to alignment from features and compact readout. Concatenating all VFM features leads to more complete geometry, \eg, auditorium seat backs (X-shaped gaps \textit{vs.} straight-line gaps), while fine-tuned DUSt3R$^*$ features further refine details, \eg, signage.}
  \label{fig:SOTA}
  \vspace{-16pt}
\end{figure*}

\qheading{Feature Pickup.} Inspired by \cref{sec:results}, we make three \featgs variants to compare with \instantsplat~\cite{InstantSplat}, in the NVS task using casual (sparse and uncalibrated) images.
Specifically, we pick up the TOP1 of \gmode, RADIO, as the first baseline.
As shown in ~\cref{tab:app}, \featgs with RADIO features achieves better PSNR and SSIM over \instantsplat.
The qualitative results in \cref{fig:SOTA} show that \instantsplat often produces broken structures and discontinuity artifacts. This occurs because optimizing millions of 3DGS for sparse viewpoints leads to overfitting high-frequency details, resulting in suboptimal parameters.
In contrast, \gmode of \featgs with RADIO features can produce higher-quality synthesized novel views, thanks to RADIO's strong geometry awareness. This allows us to read out 3DGS from the deep features using a very lightweight (2-layer) MLP, which is crucial for avoiding overfitting.

\begin{figure}[t!]
  \centering
  \includegraphics[width=1\linewidth,page=1]{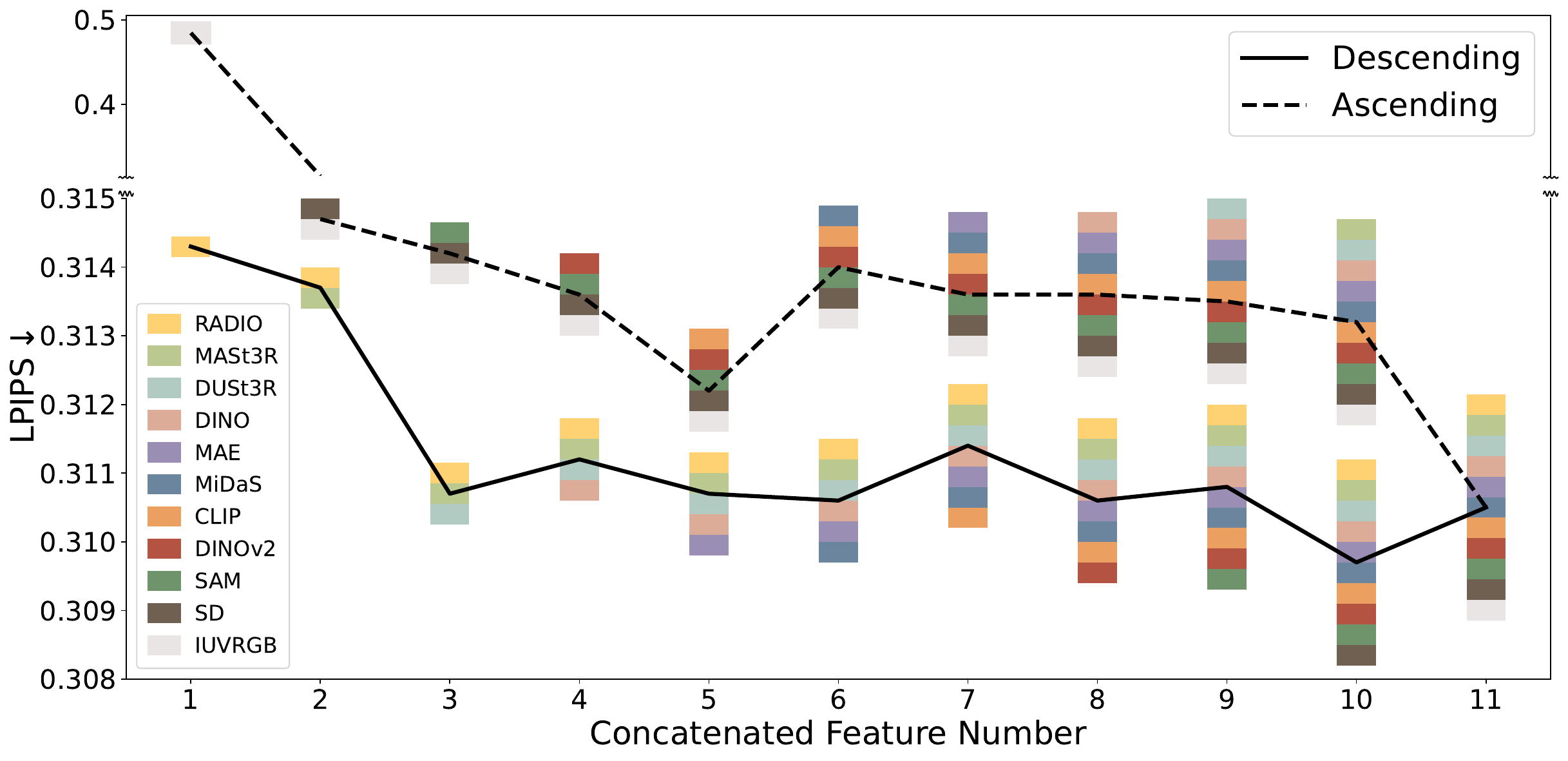}
  \vspace{-20pt}
  \caption{{\bf Ranking-Ordered Feature Concatenation.} Based on performance on \gmode,
  VFM features are ranked and concatenated in two orders: descending (best to worst) and ascending (worst to best). Performance improves with more concatenated features and with higher-ranking features. 
  }
  \label{fig:line_ranking}
  \vspace{-16pt}
\end{figure}

\pheading{Feature Ensembling.} Since simple feature concatenation is effective (see \cref{fig:cat_feats}), it makes sense to consider a straightforward approach: concatenating different VFM features. However, exploring all possible feature combinations is impractical. 
Thus, we rank the features based on \gmode performance, 
and concatenate them in descending (from best- to worst-performing) and ascending (from worst- to best-performing) orders, followed by PCA to reduce the feature dimensions to 256.
The results are in~\cref{fig:line_ranking}. The curve indicates that performance improves as more VFM features are concatenated, with additional gains when higher-ranking VFM features are merged.
Based on this observation, we simply concatenate all VFM features as our second baseline.
Compared to using only RADIO features, as shown in \cref{fig:SOTA}, ``\featgs w/ concat all'' yields better results.
For example, with RADIO features, the auditorium seat backs show X-shaped gaps, whereas concat-all model correctly recover the straight-line gaps between them.
Quantitative \cref{tab:app} also show improvements compared to \featgs based on RADIO features, yet it still falls short of \instantsplat in terms of LPIPS. 
This limitation mainly arises from the low-resolution features extracted from VFM encoders~\cite{LiFT,featup}. Though the feature upsampler~\cite{featup} is leveraged to improve the spatial resolution of features, it does not bring much benefits (detailed in \suppl).

\setlength{\tabcolsep}{4pt}
\begin{table}[t]
\centering
\begin{tabular}{l|ccc}
\multicolumn{1}{c|}{} & \multicolumn{3}{c}{All Datasets} \\
\shline
Method                & PSNR $\uparrow$ & SSIM $\uparrow$ & LPIPS $\downarrow$ \\
\shline
InstantSplat~\cite{InstantSplat}          &           18.87 &          0.6044 &             0.3039 \\
\featgs w/ RADIO      &           19.73 &          0.6513 &             0.3143 \\
\featgs w/ concat all &  \textbf{19.80} &          0.6545 &             0.3105 \\
\featgs w/ DUSt3R     &           19.66 &          0.6469 &             0.3247 \\
\featgs w/ DUSt3R$^*$ &           19.75 & \textbf{0.6561} &    \textbf{0.2928} \\
\end{tabular}
\vspace{-8pt}
\caption{{\bf Baselines of \featgs} in NVS from casual (sparse
and uncalibrated) images. We compare \gmode with RADIO features, concatenation of all VFM features, DUSt3R, and fine-tuned DUSt3R$^*$ features against the current SOTA InstantSplat. 
}
\label{tab:app}
\vspace{-16pt}
\end{table}

\pheading{Feature Finetuning.} Lastly, we explore whether feature fine-tuning during the warm-start stage is beneficial.
The results after fine-tuning different VFM features showed minimal differences, indicating that fine-tuning is effective with any well-initialized features (detailed in \suppl).
Since we use DUSt3R to warm-start the readout layer, for simplicity, we compare vanilla \dustr with fine-tuned DUSt3R* at~\cref{tab:app}. Feature fine-tuning could improve NVS. \Cref{fig:SOTA} demonstrates a clear improvement over SOTA \instantsplat.

\section{Conclusion}

We now return to our original question: Are visual foundation models (VFMs) aware of geometry and texture? 
To give an answer based on diverse datasets, we proposed \featgs, a method that maps features of VFMs to 3DGS, allowing us to explore their geometry and texture awareness through 2D images without requiring 3D \gt. 

Our work reveals new insights: correlations between novel view synthesis and 3D metrics (\cref{tab:DTU}, \cref{fig:metric3d}), color shift in SD (\cref{fig:sd_shift}), and better view consistency from models trained on pointmaps over depth (\cref{fig:gmode}). 
In addition, \featgs effectively harnesses VFMs for novel view synthesis task on sparse casual captures.
These findings suggest that predicting 3D Gaussians from various views in a canonical space and training the model with photometric loss is a promising strategy for developing 3D VFMs, also noted at~\cite{NoPoSplat,LSM}.
Moreover, VFM feature ensembling is an interesting topic worth exploring~\cite{RADIO,Theia}, and we demonstrate that this can be effectively achieved in \featgs through simple concatenation.
We hope these insights, along with \featgs --- a versatile tool for future model exploration --- will advance VFM research and drive progress in 3D vision.

\bigskip

\noindent\textbf{Acknowledgments.} 
We thank \textit{Yuxuan Xue}, \textit{Vladimir Guzov}, \textit{Garvita Tiwari} for their valuable feedback, and the members of \textit{Endless AI Lab} and \textit{Real Virtual Humans} for their help and discussions. This work is funded by the Research Center for Industries of the Future (RCIF) at Westlake University, the Westlake Education Foundation, the Deutsche Forschungsgemeinschaft (DFG, German Research Foundation) - 409792180 (EmmyNoether Programme, project: Real Virtual Humans), and the German Federal Ministry of Education and Research (BMBF): Tübingen AI Center, FKZ: 01IS18039A.
\textit{Yuliang Xiu} also received funding from the European Union's Horizon $2020$ research and innovation programme under the Marie Skłodowska-Curie grant agreement No.$860768$ (\href{https://www.clipe-itn.eu}{CLIPE} project), and Max Planck Institute for Intelligent Systems.
\textit{Gerard Pons-Moll} is a Professor at the University of Tübingen endowed by the Carl Zeiss Foundation, at the Department of Computer Science and a member of the Machine Learning Cluster of Excellence, EXC number 2064/1 – Project number 390727645.

{
    \small
    \bibliographystyle{config/ieeenat_fullname}
    \bibliography{main}

@String(CVPR= {IEEE Conf. Comput. Vis. Pattern Recog.})

@String(ICCV= {Int. Conf. Comput. Vis.})

@String(ECCV= {Eur. Conf. Comput. Vis.})

@String(TOG= {ACM Trans. Graph.})

@String(ICLR = {Int. Conf. Learn. Represent.})

@String(CVPR  = {CVPR})

@String(ICCV  = {ICCV})

@String(ECCV  = {ECCV})

@String(TOG   = {ACM TOG})

@String(ICLR  = {ICLR})

@article{InstantSplat,
  title={Instantsplat: Unbounded sparse-view pose-free gaussian splatting in 40 seconds},
  author={Fan, Zhiwen and Cong, Wenyan and Wen, Kairun and Wang, Kevin and Zhang, Jian and Ding, Xinghao and Xu, Danfei and Ivanovic, Boris and Pavone, Marco and Pavlakos, Georgios and others},
  journal={arXiv preprint arXiv:2403.20309},
  year={2024}
}

@article{Splatt3R,
  title={Splatt3R: Zero-shot Gaussian Splatting from Uncalibrated Image Pairs}, 
  author={Brandon Smart and Chuanxia Zheng and Iro Laina and Victor Adrian Prisacariu},
  journal={arXiv preprint arXiv:2408.13912},
  year={2024}
}

@inproceedings{Probe3D,
  title={Probing the 3d awareness of visual foundation models},
  author={El Banani, Mohamed and Raj, Amit and Maninis, Kevis-Kokitsi and Kar, Abhishek and Li, Yuanzhen and Rubinstein, Michael and Sun, Deqing and Guibas, Leonidas and Johnson, Justin and Jampani, Varun},
  booktitle={Proceedings of the IEEE/CVF Conference on Computer Vision and Pattern Recognition},
  pages={21795--21806},
  year={2024}
}

@InProceedings{pmlr-v119-chen20j,
  title = 	 {A Simple Framework for Contrastive Learning of Visual Representations},
  author =       {Chen, Ting and Kornblith, Simon and Norouzi, Mohammad and Hinton, Geoffrey},
  booktitle = 	 {Proceedings of the 37th International Conference on Machine Learning},
  pages = 	 {1597--1607},
  year = 	 {2020},
  editor = 	 {III, Hal Daumé and Singh, Aarti},
  volume = 	 {119},
  series = 	 {Proceedings of Machine Learning Research},
  month = 	 {13--18 Jul},
  publisher =    {PMLR},
  url = 	 {https://proceedings.mlr.press/v119/chen20j.html}
}

@article{grill2020bootstrap,
  title={Bootstrap your own latent-a new approach to self-supervised learning},
  author={Grill, Jean-Bastien and Strub, Florian and Altch{\'e}, Florent and Tallec, Corentin and Richemond, Pierre and Buchatskaya, Elena and Doersch, Carl and Avila Pires, Bernardo and Guo, Zhaohan and Gheshlaghi Azar, Mohammad and others},
  journal={Advances in neural information processing systems},
  volume={33},
  pages={21271--21284},
  year={2020}
}

@article{wang2024instantid,
  title={InstantID: Zero-shot Identity-Preserving Generation in Seconds},
  author={Wang, Qixun and Bai, Xu and Wang, Haofan and Qin, Zekui and Chen, Anthony},
  journal={arXiv preprint arXiv:2401.07519},
  year={2024}
}

@inproceedings{khirodkar2024sapiens,
  title={Sapiens: Foundation for Human Vision Models},
  author={Khirodkar, Rawal and Bagautdinov, Timur and Martinez, Julieta and Zhaoen, Su and James, Austin and Selednik, Peter and Anderson, Stuart and Saito, Shunsuke},
  booktitle=ECCV,
  year={2024}
}

@article{paszke2019pytorch,
  title={Pytorch: An imperative style, high-performance deep learning library},
  author={Paszke, Adam and Gross, Sam and Massa, Francisco and Lerer, Adam and Bradbury, James and Chanan, Gregory and Killeen, Trevor and Lin, Zeming and Gimelshein, Natalia and Antiga, Luca and others},
  journal={Advances in neural information processing systems},
  volume={32},
  year={2019}
}

@article{aydemir2024can,
            title={Can Visual Foundation Models Achieve Long-term Point Tracking?},
            author={Aydemir, G{\"o}rkay and Xie, Weidi and G{\"u}ney, Fatma},
            journal={arXiv preprint arXiv:2408.13575},
            year={2024}
          }

@article{rayr2d3,
  title={R2D3: Imparting Spatial Reasoning by Reconstructing 3D Scenes from 2D Images},
  author={Ray, Arijit and Bashkirova, Dina and Tan, Reuben and Zeng, Kuo-Hao and Plummer, Bryan A and Krishna, Ranjay and Saenko, Kate}
}

@inproceedings{Marigold,
  title={Repurposing diffusion-based image generators for monocular depth estimation},
  author={Ke, Bingxin and Obukhov, Anton and Huang, Shengyu and Metzger, Nando and Daudt, Rodrigo Caye and Schindler, Konrad},
  booktitle={Proceedings of the IEEE/CVF Conference on Computer Vision and Pattern Recognition},
  pages={9492--9502},
  year={2024}
}

@article{MiDaS,
  title={Towards robust monocular depth estimation: Mixing datasets for zero-shot cross-dataset transfer},
  author={Ranftl, Ren{\'e} and Lasinger, Katrin and Hafner, David and Schindler, Konrad and Koltun, Vladlen},
  journal={IEEE transactions on pattern analysis and machine intelligence},
  volume={44},
  number={3},
  pages={1623--1637},
  year={2020},
  publisher={IEEE}
}

@article{DepthCrafter,
  title={Depthcrafter: Generating consistent long depth sequences for open-world videos},
  author={Hu, Wenbo and Gao, Xiangjun and Li, Xiaoyu and Zhao, Sijie and Cun, Xiaodong and Zhang, Yong and Quan, Long and Shan, Ying},
  journal={arXiv preprint arXiv:2409.02095},
  year={2024}
}

@article{FMs,
  title={On the opportunities and risks of foundation models},
  author={Bommasani, Rishi and Hudson, Drew A and Adeli, Ehsan and Altman, Russ and Arora, Simran and von Arx, Sydney and Bernstein, Michael S and Bohg, Jeannette and Bosselut, Antoine and Brunskill, Emma and others},
  journal={arXiv preprint arXiv:2108.07258},
  year={2021}
}

@inproceedings{dino_tracker,
    author        = {Tumanyan, Narek and Singer, Assaf and Bagon, Shai and Dekel, Tali},
    title         = {DINO-Tracker: Taming DINO for Self-Supervised Point Tracking in a Single Video},
    year          = {2024},
    booktitle = {European Conference on Computer Vision (ECCV)},
}

@inproceedings{DSINE,
    title     = {Rethinking Inductive Biases for Surface Normal Estimation},
    author    = {Gwangbin Bae and Andrew J. Davison},
    booktitle = {IEEE/CVF Conference on Computer Vision and Pattern Recognition (CVPR)},
    year      = {2024}
}

@inproceedings{geowizard,
  title={GeoWizard: Unleashing the Diffusion Priors for 3D Geometry Estimation from a Single Image},
  author={Fu, Xiao and Yin, Wei and Hu, Mu and Wang, Kaixuan and Ma, Yuexin and Tan, Ping and Shen, Shaojie and Lin, Dahua and Long, Xiaoxiao},
  booktitle={ECCV},
  year={2024}
}

@article{stablenormal,
  title={StableNormal: Reducing Diffusion Variance for Stable and Sharp Normal},
  author={Ye, Chongjie and Qiu, Lingteng and Gu, Xiaodong and Zuo, Qi and Wu, Yushuang and Dong, Zilong and Bo, Liefeng and Xiu, Yuliang and Han, Xiaoguang},
  journal={ACM Transactions on Graphics (TOG)},
  year={2024},
  publisher={ACM New York, NY, USA}
}

@inproceedings{Spatialvlm,
  title={Spatialvlm: Endowing vision-language models with spatial reasoning capabilities},
  author={Chen, Boyuan and Xu, Zhuo and Kirmani, Sean and Ichter, Brain and Sadigh, Dorsa and Guibas, Leonidas and Xia, Fei},
  booktitle={Proceedings of the IEEE/CVF Conference on Computer Vision and Pattern Recognition},
  pages={14455--14465},
  year={2024}
}

@inproceedings{Chatpose,
  title={Chatpose: Chatting about 3d human pose},
  author={Feng, Yao and Lin, Jing and Dwivedi, Sai Kumar and Sun, Yu and Patel, Priyanka and Black, Michael J},
  booktitle={Proceedings of the IEEE/CVF Conference on Computer Vision and Pattern Recognition},
  pages={2093--2103},
  year={2024}
}

@article{foundpose,
  author    = {{\"O}rnek, Evin P{\i}nar and Labb\'e, Yann and Tekin, Bugra and Ma, Lingni and Keskin, Cem and Forster, Christian and Hoda{\v{n}}, Tom{\'a}{\v{s}}},
  title     = {FoundPose: Unseen Object Pose Estimation with Foundation Features}, 
  journal   = {ECCV},
  year      = {2024},
}

@misc{mochi,
      title={Evaluating Multiview Object Consistency in Humans and Image Models}, 
      author={Tyler Bonnen and Stephanie Fu and Yutong Bai and Thomas O'Connell and Yoni Friedman and Nancy Kanwisher and Joshua B. Tenenbaum and Alexei A. Efros},
      year={2024},
      eprint={2409.05862},
      archivePrefix={arXiv},
      primaryClass={cs.CV},
      url={https://arxiv.org/abs/2409.05862}, 
      }

@article{zuo2024towards,
  title={Towards Foundation Models for 3D Vision: How Close Are We?},
  author={Zuo, Yiming and Kayan, Karhan and Wang, Maggie and Jeon, Kevin and Deng, Jia and Griffiths, Thomas L},
  journal={arXiv preprint arXiv:2410.10799},
  year={2024}
}

@article{BLINK,
  title={BLINK: Multimodal Large Language Models Can See but Not Perceive},
  author={Fu, Xingyu and Hu, Yushi and Li, Bangzheng and Feng, Yu and Wang, Haoyu and Lin, Xudong and Roth, Dan and Smith, Noah A and Ma, Wei-Chiu and Krishna, Ranjay},
  booktitle={ECCV},
  year={2024}
}

@article{3D_PC,
  title={The 3D-PC: a benchmark for visual perspective taking in humans and machines},
  author={Linsley, Drew and Zhou, Peisen and Ashok, Alekh Karkada and Nagaraj, Akash and Gaonkar, Gaurav and Lewis, Francis E and Pizlo, Zygmunt and Serre, Thomas},
  journal={arXiv preprint arXiv:2406.04138},
  year={2024}
}

@article{CortexBench,
  title={Where are we in the search for an artificial visual cortex for embodied intelligence?},
  author={Majumdar, Arjun and Yadav, Karmesh and Arnaud, Sergio and Ma, Jason and Chen, Claire and Silwal, Sneha and Jain, Aryan and Berges, Vincent-Pierre and Wu, Tingfan and Vakil, Jay and others},
  journal={Advances in Neural Information Processing Systems},
  volume={36},
  pages={655--677},
  year={2023}
}

@inproceedings{Theia,
    title={Theia: Distilling Diverse Vision Foundation Models for Robot Learning},
    author={Jinghuan Shang and Karl Schmeckpeper and Brandon B. May and Maria Vittoria Minniti and Tarik Kelestemur and David Watkins and Laura Herlant},
    booktitle={8th Annual Conference on Robot Learning},
    year={2024},
    url={https://openreview.net/forum?id=ylZHvlwUcI}
}

@article{SPA,
  title={SPA: 3D Spatial-Awareness Enables Effective Embodied Representation},
  author={Zhu, Haoyi and Yang, Honghui and Wang, Yating and Yang, Jiange and Wang, Limin and He, Tong},
  journal={arXiv preprint arXiv:2410.08208},
  year={2024}
}

@inproceedings{NeRF,
 title={NeRF: Representing Scenes as Neural Radiance Fields for View Synthesis},
 author={Ben Mildenhall and Pratul P. Srinivasan and Matthew Tancik and Jonathan T. Barron and Ravi Ramamoorthi and Ren Ng},
 year={2020},
 booktitle={ECCV},
}

@article{instant_ngp,
    author = {Thomas M\"uller and Alex Evans and Christoph Schied and Alexander Keller},
    title = {Instant Neural Graphics Primitives with a Multiresolution Hash Encoding},
    journal = {ACM Trans. Graph.},
    issue_date = {July 2022},
    volume = {41},
    number = {4},
    month = jul,
    year = {2022},
    pages = {102:1--102:15},
    articleno = {102},
    numpages = {15},
    url = {https://doi.org/10.1145/3528223.3530127},
    doi = {10.1145/3528223.3530127},
    publisher = {ACM},
    address = {New York, NY, USA}
}

@Article{3DGS,
      author       = {Kerbl, Bernhard and Kopanas, Georgios and Leimk{\"u}hler, Thomas and Drettakis, George},
      title        = {3D Gaussian Splatting for Real-Time Radiance Field Rendering},
      journal      = {ACM Transactions on Graphics},
      number       = {4},
      volume       = {42},
      month        = {July},
      year         = {2023},
      url          = {https://repo-sam.inria.fr/fungraph/3d-gaussian-splatting/}
}

@INPROCEEDINGS{TensoRF,
  author = {Anpei Chen and Zexiang Xu and Andreas Geiger and Jingyi Yu and Hao Su},
  title = {TensoRF: Tensorial Radiance Fields},
  booktitle = {European Conference on Computer Vision (ECCV)},
  year = {2022}
}

@inproceedings{nerfstudio,
	title        = {Nerfstudio: A Modular Framework for Neural Radiance Field Development},
	author       = {
		Tancik, Matthew and Weber, Ethan and Ng, Evonne and Li, Ruilong and Yi, Brent
		and Kerr, Justin and Wang, Terrance and Kristoffersen, Alexander and Austin,
		Jake and Salahi, Kamyar and Ahuja, Abhik and McAllister, David and Kanazawa,
		Angjoo
	},
	year         = 2023,
	booktitle    = {ACM SIGGRAPH 2023 Conference Proceedings},
	series       = {SIGGRAPH '23}
}

@article{gsplat,
    title={gsplat: An Open-Source Library for {Gaussian} Splatting}, 
    author={Vickie Ye and Ruilong Li and Justin Kerr and Matias Turkulainen and Brent Yi and Zhuoyang Pan and Otto Seiskari and Jianbo Ye and Jeffrey Hu and Matthew Tancik and Angjoo Kanazawa},
    year={2024},
    eprint={2409.06765},
    journal={arXiv preprint arXiv:2409.06765},
    archivePrefix={arXiv},
    primaryClass={cs.CV},
    url={https://arxiv.org/abs/2409.06765}, 
}

@inproceedings{Regnerf,
  title={Regnerf: Regularizing neural radiance fields for view synthesis from sparse inputs},
  author={Niemeyer, Michael and Barron, Jonathan T and Mildenhall, Ben and Sajjadi, Mehdi SM and Geiger, Andreas and Radwan, Noha},
  booktitle={Proceedings of the IEEE/CVF Conference on Computer Vision and Pattern Recognition},
  pages={5480--5490},
  year={2022}
}

@inproceedings{Depth_supervised_nerf,
  title={Depth-supervised nerf: Fewer views and faster training for free},
  author={Deng, Kangle and Liu, Andrew and Zhu, Jun-Yan and Ramanan, Deva},
  booktitle={Proceedings of the IEEE/CVF Conference on Computer Vision and Pattern Recognition},
  pages={12882--12891},
  year={2022}
}

@inproceedings{Nerfbusters,
      title = {Nerfbusters: Removing Ghostly Artifacts from Casually Captured NeRFs},
      author = {Frederik Warburg* and Ethan Weber* and Matthew Tancik and Aleksander Hołyński and Angjoo Kanazawa},
      booktitle = {International Conference on Computer Vision (ICCV)},
      year = {2023},
}

@InProceedings{DietNeRF,
  author = {Jain, Ajay and Tancik, Matthew and Abbeel, Pieter},
  title = {Putting NeRF on a Diet: Semantically Consistent Few-Shot View Synthesis},
  booktitle = {Proceedings of the IEEE/CVF International Conference on Computer Vision (ICCV)},
  month = {October},
  year = {2021},
  pages = {5885-5894}
}

@InProceedings{SinNeRF,
    author = {Xu, Dejia and Jiang, Yifan and Wang, Peihao and Fan, Zhiwen and Shi, Humphrey and Wang, Zhangyang},
    title = {SinNeRF: Training Neural Radiance Fields on Complex Scenes from a Single Image},
    journal={arXiv preprint arXiv:2204.00928},
    year={2022}
}

@inproceedings{Freenerf,
  title={Freenerf: Improving few-shot neural rendering with free frequency regularization},
  author={Yang, Jiawei and Pavone, Marco and Wang, Yue},
  booktitle={Proceedings of the IEEE/CVF conference on computer vision and pattern recognition},
  pages={8254--8263},
  year={2023}
}

@misc{FSGS,
    title={FSGS: Real-Time Few-Shot View Synthesis using Gaussian Splatting},
    author={Zehao Zhu and Zhiwen Fan and Yifan Jiang and Zhangyang Wang},
    year={2023},
    eprint={2312.00451},
    archivePrefix={arXiv},
    primaryClass={cs.CV}
}

@inproceedings{Reconfusion,
  title={Reconfusion: 3d reconstruction with diffusion priors},
  author={Wu, Rundi and Mildenhall, Ben and Henzler, Philipp and Park, Keunhong and Gao, Ruiqi and Watson, Daniel and Srinivasan, Pratul P and Verbin, Dor and Barron, Jonathan T and Poole, Ben and others},
  booktitle={Proceedings of the IEEE/CVF Conference on Computer Vision and Pattern Recognition},
  pages={21551--21561},
  year={2024}
}

@inproceedings{colmap,
    author={Sch\"{o}nberger, Johannes Lutz and Frahm, Jan-Michael},
    title={Structure-from-Motion Revisited},
    booktitle={Conference on Computer Vision and Pattern Recognition (CVPR)},
    year={2016},
}

@article{nerfmm,
  title={Ne{RF}$--$: Neural Radiance Fields Without Known Camera Parameters},
  author={Zirui Wang and Shangzhe Wu and Weidi Xie and Min Chen and Victor Adrian Prisacariu},
  journal={arXiv preprint arXiv:2102.07064},
  year={2021}
}

@inproceedings{Barf,
  title={Barf: Bundle-adjusting neural radiance fields},
  author={Lin, Chen-Hsuan and Ma, Wei-Chiu and Torralba, Antonio and Lucey, Simon},
  booktitle={Proceedings of the IEEE/CVF international conference on computer vision},
  pages={5741--5751},
  year={2021}
}

@inproceedings{l2g,
  title={Local-to-global registration for bundle-adjusting neural radiance fields},
  author={Chen, Yue and Chen, Xingyu and Wang, Xuan and Zhang, Qi and Guo, Yu and Shan, Ying and Wang, Fei},
  booktitle={Proceedings of the IEEE/CVF Conference on Computer Vision and Pattern Recognition},
  pages={8264--8273},
  year={2023}
}

@inproceedings{SCNeRF,
  title={Self-calibrating neural radiance fields},
  author={Jeong, Yoonwoo and Ahn, Seokjun and Choy, Christopher and Anandkumar, Anima and Cho, Minsu and Park, Jaesik},
  booktitle={Proceedings of the IEEE/CVF International Conference on Computer Vision},
  pages={5846--5854},
  year={2021}
}

@InProceedings{gnerf,
    author = {Meng, Quan and Chen, Anpei and Luo, Haimin and Wu, Minye and Su, Hao and Xu, Lan and He, Xuming and Yu, Jingyi},
    title = {{G}{N}e{R}{F}: {G}{A}{N}-based {N}eural {R}adiance {F}ield without {P}osed {C}amera},
    booktitle = {Proceedings of the IEEE/CVF International Conference on Computer Vision (ICCV)},
    year = {2021}
}

@inproceedings{Sparf,
  title={Sparf: Neural radiance fields from sparse and noisy poses},
  author={Truong, Prune and Rakotosaona, Marie-Julie and Manhardt, Fabian and Tombari, Federico},
  booktitle={Proceedings of the IEEE/CVF Conference on Computer Vision and Pattern Recognition},
  pages={4190--4200},
  year={2023}
}

@inproceedings{Nope-nerf,
  title={Nope-nerf: Optimising neural radiance field with no pose prior},
  author={Bian, Wenjing and Wang, Zirui and Li, Kejie and Bian, Jia-Wang and Prisacariu, Victor Adrian},
  booktitle={Proceedings of the IEEE/CVF Conference on Computer Vision and Pattern Recognition},
  pages={4160--4169},
  year={2023}
}

@InProceedings{cf3dgs,
    author    = {Fu, Yang and Liu, Sifei and Kulkarni, Amey and Kautz, Jan and Efros, Alexei A. and Wang, Xiaolong},
    title     = {COLMAP-Free 3D Gaussian Splatting},
    booktitle = {Proceedings of the IEEE/CVF Conference on Computer Vision and Pattern Recognition (CVPR)},
    month     = {June},
    year      = {2024},
    pages     = {20796-20805}
}

@inproceedings{LocalRF,
  title={Progressively optimized local radiance fields for robust view synthesis},
  author={Meuleman, Andreas and Liu, Yu-Lun and Gao, Chen and Huang, Jia-Bin and Kim, Changil and Kim, Min H and Kopf, Johannes},
  booktitle={Proceedings of the IEEE/CVF Conference on Computer Vision and Pattern Recognition},
  pages={16539--16548},
  year={2023}
}

@inproceedings{pixelNeRF,
      title={{pixelNeRF}: Neural Radiance Fields from One or Few Images},
      author={Alex Yu and Vickie Ye and Matthew Tancik and Angjoo Kanazawa},
      year={2021},
      booktitle={CVPR},
}

@inproceedings{Zero-1-to-3,
  title={Zero-1-to-3: Zero-shot one image to 3d object},
  author={Liu, Ruoshi and Wu, Rundi and Van Hoorick, Basile and Tokmakov, Pavel and Zakharov, Sergey and Vondrick, Carl},
  booktitle={Proceedings of the IEEE/CVF international conference on computer vision},
  pages={9298--9309},
  year={2023}
}

@inproceedings{ZeroNVS,
  title={ZeroNVS: Zero-Shot 360-Degree View Synthesis from a Single Image},
  author={Sargent, Kyle and Li, Zizhang and Shah, Tanmay and Herrmann, Charles and Yu, Hong-Xing and Zhang, Yunzhi and Chan, Eric Ryan and Lagun, Dmitry and Fei-Fei, Li and Sun, Deqing and others},
  booktitle={Proceedings of the IEEE/CVF Conference on Computer Vision and Pattern Recognition},
  pages={9420--9429},
  year={2024}
}

@inproceedings{MegaScenes,
  title={MegaScenes: Scene-Level View Synthesis at Scale}, 
  author={Tung, Joseph and Chou, Gene and Cai, Ruojin and Yang, Guandao and Zhang, Kai and Wetzstein, Gordon and Hariharan, Bharath and Snavely, Noah},
  booktitle={ECCV},
  year={2024}
}

@inproceedings{pixelsplat,
  title={pixelsplat: 3d gaussian splats from image pairs for scalable generalizable 3d reconstruction},
  author={Charatan, David and Li, Sizhe Lester and Tagliasacchi, Andrea and Sitzmann, Vincent},
  booktitle={Proceedings of the IEEE/CVF Conference on Computer Vision and Pattern Recognition},
  pages={19457--19467},
  year={2024}
}

@article{MVSplat,
    title   = {MVSplat: Efficient 3D Gaussian Splatting from Sparse Multi-View Images},
    author  = {Chen, Yuedong and Xu, Haofei and Zheng, Chuanxia and Zhuang, Bohan and Pollefeys, Marc and Geiger, Andreas and Cham, Tat-Jen and Cai, Jianfei},
    journal = {ECCV},
    year    = {2024},
}

@article{Lrm,
  title={Lrm: Large reconstruction model for single image to 3d},
  author={Hong, Yicong and Zhang, Kai and Gu, Jiuxiang and Bi, Sai and Zhou, Yang and Liu, Difan and Liu, Feng and Sunkavalli, Kalyan and Bui, Trung and Tan, Hao},
  journal={ICLR},
  year={2024}
}

@inproceedings{LaRa,
         author = {Anpei Chen and Haofei Xu and Stefano Esposito and Siyu Tang and Andreas Geiger},
         title = {LaRa: Efficient Large-Baseline Radiance Fields},
         booktitle = {European Conference on Computer Vision (ECCV)},
         year = {2024}
        }

@misc{LVSM,
      title={LVSM: A Large View Synthesis Model with Minimal 3D Inductive Bias}, 
      author={Haian Jin and Hanwen Jiang and Hao Tan and Kai Zhang and Sai Bi and Tianyuan Zhang and Fujun Luan and Noah Snavely and Zexiang Xu},
      year={2024},
      eprint={2410.17242},
      archivePrefix={arXiv},
      primaryClass={cs.CV},
      url={https://arxiv.org/abs/2410.17242}, 
}

@article{Long-LRM,
  title={Long-LRM: Long-sequence Large Reconstruction Model for Wide-coverage Gaussian Splats},
  author={Ziwen, Chen and Tan, Hao and Zhang, Kai and Bi, Sai and Luan, Fujun and Hong, Yicong and Fuxin, Li and Xu, Zexiang},
  journal={arXiv preprint 2410.12781},
  year={2024}
}

@InProceedings{DUSt3R,
    author    = {Wang, Shuzhe and Leroy, Vincent and Cabon, Yohann and Chidlovskii, Boris and Revaud, Jerome},
    title     = {DUSt3R: Geometric 3D Vision Made Easy},
    booktitle = {Proceedings of the IEEE/CVF Conference on Computer Vision and Pattern Recognition (CVPR)},
    month     = {June},
    year      = {2024},
    pages     = {20697-20709}
}

@misc{MASt3R,
      title={Grounding Image Matching in 3D with MASt3R}, 
      author={Vincent Leroy and Yohann Cabon and Jerome Revaud},
      year={2024},
      eprint={2406.09756},
      archivePrefix={arXiv},
      primaryClass={cs.CV}
}

@article{spann3r,
  title={3D Reconstruction with Spatial Memory},
  author={Wang, Hengyi and Agapito, Lourdes},
  journal={arXiv preprint arXiv:2408.16061},
  year={2024}
}

@article{MonST3R,
  title={MonST3R: A Simple Approach for Estimating Geometry in the Presence of Motion},
  author={Zhang, Junyi and Herrmann, Charles and Hur, Junhwa and Jampani, Varun and Darrell, Trevor and Cole, Forrester and Sun, Deqing and Yang, Ming-Hsuan},
  journal={arXiv preprint arxiv:2410.03825},
  year={2024}
}

@article{LSM,
  title={Large Spatial Model: End-to-end Unposed Images to Semantic 3D},
  author={Fan, Zhiwen and Zhang, Jian and Cong, Wenyan and Wang, Peihao and Li, Renjie and Wen, Kairun and Zhou, Shijie and Kadambi, Achuta and Wang, Zhangyang and Xu, Danfei and others},
  journal={NeurIPS},
  year={2024}
}

@misc{NoPoSplat,
      title={No Pose, No Problem: Surprisingly Simple 3D Gaussian Splats from Sparse Unposed Images}, 
      author={Botao Ye and Sifei Liu and Haofei Xu and Xueting Li and Marc Pollefeys and Ming-Hsuan Yang and Songyou Peng},
      year={2024},
      eprint={2410.24207},
      archivePrefix={arXiv},
      primaryClass={cs.CV},
      url={https://arxiv.org/abs/2410.24207}, 
}

@article{Tanks,
  title={Tanks and temples: Benchmarking large-scale scene reconstruction},
  author={Knapitsch, Arno and Park, Jaesik and Zhou, Qian-Yi and Koltun, Vladlen},
  journal={ACM Transactions on Graphics (ToG)},
  volume={36},
  number={4},
  pages={1--13},
  year={2017},
  publisher={ACM New York, NY, USA}
}

@article{llff,
  title={Local Light Field Fusion: Practical View Synthesis with Prescriptive Sampling Guidelines},
  author={Ben Mildenhall and Pratul P. Srinivasan and Rodrigo Ortiz-Cayon and Nima Khademi Kalantari and Ravi Ramamoorthi and Ren Ng and Abhishek Kar},
  journal={ACM Transactions on Graphics (TOG)},
  year={2019},
}

@inproceedings{Mip_nerf_360,
  title={Mip-nerf 360: Unbounded anti-aliased neural radiance fields},
  author={Barron, Jonathan T and Mildenhall, Ben and Verbin, Dor and Srinivasan, Pratul P and Hedman, Peter},
  booktitle={Proceedings of the IEEE/CVF conference on computer vision and pattern recognition},
  pages={5470--5479},
  year={2022}
}

@inproceedings{Mvimgnet,
  title={Mvimgnet: A large-scale dataset of multi-view images},
  author={Yu, Xianggang and Xu, Mutian and Zhang, Yidan and Liu, Haolin and Ye, Chongjie and Wu, Yushuang and Yan, Zizheng and Zhu, Chenming and Xiong, Zhangyang and Liang, Tianyou and others},
  booktitle={Proceedings of the IEEE/CVF conference on computer vision and pattern recognition},
  pages={9150--9161},
  year={2023}
}

@inproceedings{ling2024dl3dv,
  title={Dl3dv-10k: A large-scale scene dataset for deep learning-based 3d vision},
  author={Ling, Lu and Sheng, Yichen and Tu, Zhi and Zhao, Wentian and Xin, Cheng and Wan, Kun and Yu, Lantao and Guo, Qianyu and Yu, Zixun and Lu, Yawen and others},
  booktitle={Proceedings of the IEEE/CVF Conference on Computer Vision and Pattern Recognition},
  pages={22160--22169},
  year={2024}
}

@article{FlowMap,
  title={FlowMap: High-Quality Camera Poses, Intrinsics, and Depth via Gradient Descent},
  author={Smith, Cameron and Charatan, David and Tewari, Ayush and Sitzmann, Vincent},
  journal={arXiv preprint arXiv:2404.15259},
  year={2024}
}

@inproceedings{acezero,
    title={Scene Coordinate Reconstruction: Posing of Image Collections via Incremental Learning of a Relocalizer},
    author={Brachmann, Eric and Wynn, Jamie and Chen, Shuai and Cavallari, Tommaso and Monszpart, {\'{A}}ron and Turmukhambetov, Daniyar and Prisacariu, Victor Adrian},
    booktitle={ECCV},
    year={2024},
}

@inproceedings{ye2023featurenerf,
  title={FeatureNeRF: Learning Generalizable NeRFs by Distilling Foundation Models},
  author={Ye, Jianglong and Wang, Naiyan and Wang, Xiaolong},
  booktitle={Proceedings of the IEEE/CVF International Conference on Computer Vision},
  pages={8962--8973},
  year={2023}
}

@inproceedings{lerf2023,
 author = {Kerr, Justin* and Kim, Chung Min* and Goldberg, Ken and Kanazawa, Angjoo and Tancik, Matthew},
 title = {LERF: Language Embedded Radiance Fields},
 booktitle = {International Conference on Computer Vision (ICCV)},
 year = {2023},
}

@article{ravi2024sam2,
  title={Sam 2: Segment anything in images and videos},
  author={Ravi, Nikhila and Gabeur, Valentin and Hu, Yuan-Ting and Hu, Ronghang and Ryali, Chaitanya and Ma, Tengyu and Khedr, Haitham and R{\"a}dle, Roman and Rolland, Chloe and Gustafson, Laura and others},
  journal={arXiv preprint arXiv:2408.00714},
  year={2024}
}

@inproceedings{tschernezki22neural,
  author     = {Vadim Tschernezki and Iro Laina and 
                Diane Larlus and Andrea Vedaldi},
  booktitle  = {Proceedings of the International Conference
                on {3D} Vision (3DV)},
  title      = {{Neural Feature Fusion Fields}: {3D} Distillation
                of Self-Supervised {2D} Image Representations},
  year       = {2022}
}

@inproceedings{
li2022languagedriven,
title={Language-driven Semantic Segmentation},
author={Boyi Li and Kilian Q Weinberger and Serge Belongie and Vladlen Koltun and Rene Ranftl},
booktitle={International Conference on Learning Representations},
year={2022},
url={https://openreview.net/forum?id=RriDjddCLN}
}

@inproceedings{Decomposing_NeRF,
  title={Decomposing NeRF for Editing via Feature Field Distillation},
  author={Sosuke Kobayashi and Eiichi Matsumoto and Vincent Sitzmann},
  booktitle={Advances in Neural Information Processing Systems},
  volume = {35},
  url = {https://arxiv.org/pdf/2205.15585.pdf},
  year={2022}
}

@inproceedings{Peng2023OpenScene,
      title     = {OpenScene: 3D Scene Understanding with Open Vocabularies},
      author    = {Peng, Songyou and Genova, Kyle and Jiang, Chiyu "Max" and Tagliasacchi, Andrea and Pollefeys, Marc and Funkhouser, Thomas},
      booktitle = CVPR,
      year      = {2023}
  }

@inproceedings{featup,
    title={FeatUp: A Model-Agnostic Framework for Features at Any Resolution},
    author={Stephanie Fu and Mark Hamilton and Laura E. Brandt and Axel Feldmann and Zhoutong Zhang and William T. Freeman},
    booktitle={The Twelfth International Conference on Learning Representations},
    year={2024},
    url={https://openreview.net/forum?id=GkJiNn2QDF}
}

@InProceedings{RADIO,
    author    = {Ranzinger, Mike and Heinrich, Greg and Kautz, Jan and Molchanov, Pavlo},
    title     = {AM-RADIO: Agglomerative Vision Foundation Model Reduce All Domains Into One},
    booktitle = {Proceedings of the IEEE/CVF Conference on Computer Vision and Pattern Recognition (CVPR)},
    month     = {June},
    year      = {2024},
    pages     = {12490-12500}
}

@inproceedings{DINO,
  title={Emerging properties in self-supervised vision transformers},
  author={Caron, Mathilde and Touvron, Hugo and Misra, Ishan and J{\'e}gou, Herv{\'e} and Mairal, Julien and Bojanowski, Piotr and Joulin, Armand},
  booktitle={Proceedings of the IEEE/CVF international conference on computer vision},
  pages={9650--9660},
  year={2021}
}

@article{DINOv2,
  title={Dinov2: Learning robust visual features without supervision},
  author={Oquab, Maxime and Darcet, Timoth{\'e}e and Moutakanni, Th{\'e}o and Vo, Huy and Szafraniec, Marc and Khalidov, Vasil and Fernandez, Pierre and Haziza, Daniel and Massa, Francisco and El-Nouby, Alaaeldin and others},
  journal={arXiv preprint arXiv:2304.07193},
  year={2023}
}

@article{wang2021clip,
  title={CLIP-NeRF: Text-and-Image Driven Manipulation of Neural Radiance Fields},
  author={Wang, Can and Chai, Menglei and He, Mingming and Chen, Dongdong and Liao, Jing},
  journal={arXiv preprint arXiv:2112.05139},
  year={2021}
}

@inproceedings{CLIP,
  title={Learning transferable visual models from natural language supervision},
  author={Radford, Alec and Kim, Jong Wook and Hallacy, Chris and Ramesh, Aditya and Goh, Gabriel and Agarwal, Sandhini and Sastry, Girish and Askell, Amanda and Mishkin, Pamela and Clark, Jack and others},
  booktitle={International conference on machine learning},
  pages={8748--8763},
  year={2021},
  organization={PMLR}
}

@inproceedings{SAM,
  title={Segment anything},
  author={Kirillov, Alexander and Mintun, Eric and Ravi, Nikhila and Mao, Hanzi and Rolland, Chloe and Gustafson, Laura and Xiao, Tete and Whitehead, Spencer and Berg, Alexander C and Lo, Wan-Yen and others},
  booktitle={Proceedings of the IEEE/CVF International Conference on Computer Vision},
  pages={4015--4026},
  year={2023}
}

@inproceedings{MAE,
  title={Masked autoencoders are scalable vision learners},
  author={He, Kaiming and Chen, Xinlei and Xie, Saining and Li, Yanghao and Doll{\'a}r, Piotr and Girshick, Ross},
  booktitle={Proceedings of the IEEE/CVF conference on computer vision and pattern recognition},
  pages={16000--16009},
  year={2022}
}

@inproceedings{SD,
  title={High-resolution image synthesis with latent diffusion models},
  author={Rombach, Robin and Blattmann, Andreas and Lorenz, Dominik and Esser, Patrick and Ommer, Bj{\"o}rn},
  booktitle={Proceedings of the IEEE/CVF conference on computer vision and pattern recognition},
  pages={10684--10695},
  year={2022}
}

@inproceedings{shotton2013scene,
  title={Scene coordinate regression forests for camera relocalization in RGB-D images},
  author={Shotton, Jamie and Glocker, Ben and Zach, Christopher and Izadi, Shahram and Criminisi, Antonio and Fitzgibbon, Andrew},
  booktitle={Proceedings of the IEEE conference on computer vision and pattern recognition},
  pages={2930--2937},
  year={2013}
}

@article{Shortcut,
  title={Shortcut learning in deep neural networks},
  author={Geirhos, Robert and Jacobsen, J{\"o}rn-Henrik and Michaelis, Claudio and Zemel, Richard and Brendel, Wieland and Bethge, Matthias and Wichmann, Felix A},
  journal={Nature Machine Intelligence},
  volume={2},
  number={11},
  pages={665--673},
  year={2020},
  publisher={Nature Publishing Group UK London}
}

@inproceedings{biased,
title={ImageNet-trained {CNN}s are biased towards texture; increasing shape bias improves accuracy and robustness.},
author={Robert Geirhos and Patricia Rubisch and Claudio Michaelis and Matthias Bethge and Felix A Wichmann and Wieland Brendel},
booktitle={International Conference on Learning Representations},
year={2019},
url={https://openreview.net/forum?id=Bygh9j09KX},
}

@article{Croco,
  title={Croco: Self-supervised pre-training for 3d vision tasks by cross-view completion},
  author={Weinzaepfel, Philippe and Leroy, Vincent and Lucas, Thomas and Br{\'e}gier, Romain and Cabon, Yohann and Arora, Vaibhav and Antsfeld, Leonid and Chidlovskii, Boris and Csurka, Gabriela and Revaud, J{\'e}r{\^o}me},
  journal={Advances in Neural Information Processing Systems},
  volume={35},
  pages={3502--3516},
  year={2022}
}

@inproceedings{DVT,
  author = {Yang, Jiawei and Luo, Katie Z and Li, Jiefeng and Deng, Congyue and Guibas, Leonidas J. and Krishnan, Dilip and Weinberger, Kilian Q and Tian, Yonglong and Wang, Yue},
  title = {DVT: Denoising Vision Transformers},
  journal = {ECCV},
  year = {2024},
}

@inproceedings{EmerNeRF,
  title={EmerNeRF: Emergent Spatial-Temporal Scene Decomposition via Self-Supervision},
  author={Yang, Jiawei and Ivanovic, Boris and Litany, Or and Weng, Xinshuo and Kim, Seung Wook and Li, Boyi and Che, Tong and Xu, Danfei and Fidler, Sanja and Pavone, Marco and others},
  booktitle={International Conference on Learning Representations},
  year={2024},
}

@inproceedings{Registers,
title={Vision Transformers Need Registers},
author={Timoth{\'e}e Darcet and Maxime Oquab and Julien Mairal and Piotr Bojanowski},
booktitle={The Twelfth International Conference on Learning Representations},
year={2024},
url={https://openreview.net/forum?id=2dnO3LLiJ1}
}

@inproceedings{fit3d,
  title     = {{Improving 2D Feature Representations by 3D-Aware Fine-Tuning}},
  author    = {Yue, Yuanwen and Das, Anurag and Engelmann, Francis and Tang, Siyu and Lenssen, Jan Eric},
  booktitle = {European Conference on Computer Vision (ECCV)},
  year      = {2024}
}

@inproceedings{ConDense,
  title={ConDense: Consistent 2D/3D Pre-training for Dense and Sparse Features from Multi-View Images},
  author={Zhang, Xiaoshuai and Wang, Zhicheng and Zhou, Howard and Ghosh, Soham and Gnanapragasam, Danushen and Jampani, Varun and Su, Hao and Guibas, Leonidas},
  booktitle={European Conference on Computer Vision},
  pages={},
  year={2024},
  organization={Springer}
}

@inproceedings{DPT,
  title={Vision transformers for dense prediction},
  author={Ranftl, Ren{\'e} and Bochkovskiy, Alexey and Koltun, Vladlen},
  booktitle={Proceedings of the IEEE/CVF international conference on computer vision},
  pages={12179--12188},
  year={2021}
}

@inproceedings{LiFT,
  title={LiFT: A Surprisingly Simple Lightweight Feature Transform for Dense ViT Descriptors},
  author={Suri, Saksham and Walmer, Matthew and Gupta, Kamal and Shrivastava, Abhinav},
  booktitle={European Conference on Computer Vision},
  pages={110--128},
  year={2025},
  organization={Springer}
}

@article{pca,
  title={Finding structure with randomness: Probabilistic algorithms for constructing approximate matrix decompositions},
  author={Halko, Nathan and Martinsson, Per-Gunnar and Tropp, Joel A},
  journal={SIAM review},
  volume={53},
  number={2},
  pages={217--288},
  year={2011},
  publisher={SIAM}
}

@article{DTU,
  title={Large-scale data for multiple-view stereopsis},
  author={Aan{\ae}s, Henrik and Jensen, Rasmus Ramsb{\o}l and Vogiatzis, George and Tola, Engin and Dahl, Anders Bjorholm},
  journal={International Journal of Computer Vision},
  volume={120},
  pages={153--168},
  year={2016},
  publisher={Springer}
}

@inproceedings{Objaverse,
  title={Objaverse: A universe of annotated 3d objects},
  author={Deitke, Matt and Schwenk, Dustin and Salvador, Jordi and Weihs, Luca and Michel, Oscar and VanderBilt, Eli and Schmidt, Ludwig and Ehsani, Kiana and Kembhavi, Aniruddha and Farhadi, Ali},
  booktitle={Proceedings of the IEEE/CVF Conference on Computer Vision and Pattern Recognition},
  pages={13142--13153},
  year={2023}
}

@article{forgetting,
  title={Overcoming catastrophic forgetting in neural networks},
  author={Kirkpatrick, James and Pascanu, Razvan and Rabinowitz, Neil and Veness, Joel and Desjardins, Guillaume and Rusu, Andrei A and Milan, Kieran and Quan, John and Ramalho, Tiago and Grabska-Barwinska, Agnieszka and others},
  journal={Proceedings of the national academy of sciences},
  volume={114},
  number={13},
  pages={3521--3526},
  year={2017},
  publisher={National Acad Sciences}
}

@article{som,
  title={Shape of motion: 4d reconstruction from a single video},
  author={Wang, Qianqian and Ye, Vickie and Gao, Hang and Austin, Jake and Li, Zhengqi and Kanazawa, Angjoo},
  journal={arXiv e-prints},
  pages={arXiv--2407},
  year={2024}
}

@inproceedings{nerfwild,
  title={Nerf in the wild: Neural radiance fields for unconstrained photo collections},
  author={Martin-Brualla, Ricardo and Radwan, Noha and Sajjadi, Mehdi SM and Barron, Jonathan T and Dosovitskiy, Alexey and Duckworth, Daniel},
  booktitle={Proceedings of the IEEE/CVF conference on computer vision and pattern recognition},
  pages={7210--7219},
  year={2021}
}

@inproceedings{Hallucinated,
  title={Hallucinated neural radiance fields in the wild},
  author={Chen, Xingyu and Zhang, Qi and Li, Xiaoyu and Chen, Yue and Feng, Ying and Wang, Xuan and Wang, Jue},
  booktitle={Proceedings of the IEEE/CVF Conference on Computer Vision and Pattern Recognition},
  pages={12943--12952},
  year={2022}
}

@article{Splatfacto-W,
  title={Splatfacto-W: A Nerfstudio Implementation of Gaussian Splatting for Unconstrained Photo Collections},
  author={Xu, Congrong and Kerr, Justin and Kanazawa, Angjoo},
  journal={arXiv preprint arXiv:2407.12306},
  year={2024}
}

@inproceedings{Gaussianwild,
  title={Gaussian in the wild: 3d gaussian splatting for unconstrained image collections},
  author={Zhang, Dongbin and Wang, Chuming and Wang, Weitao and Li, Peihao and Qin, Minghan and Wang, Haoqian},
  booktitle={European Conference on Computer Vision},
  pages={341--359},
  year={2025},
  organization={Springer}
}

@article{Wildgaussians,
  title={Wildgaussians: 3d gaussian splatting in the wild},
  author={Kulhanek, Jonas and Peng, Songyou and Kukelova, Zuzana and Pollefeys, Marc and Sattler, Torsten},
  journal={arXiv preprint arXiv:2407.08447},
  year={2024}
}

@incollection{Phototourism,
  title={Photo tourism: exploring photo collections in 3D},
  author={Snavely, Noah and Seitz, Steven M and Szeliski, Richard},
  booktitle={ACM siggraph 2006 papers},
  pages={835--846},
  year={2006}
}

@inproceedings{Semantic_NeRF,
  title={In-place scene labelling and understanding with implicit scene representation},
  author={Zhi, Shuaifeng and Laidlow, Tristan and Leutenegger, Stefan and Davison, Andrew J},
  booktitle={Proceedings of the IEEE/CVF International Conference on Computer Vision},
  pages={15838--15847},
  year={2021}
}

@inproceedings{Panoptic,
  title={Panoptic lifting for 3d scene understanding with neural fields},
  author={Siddiqui, Yawar and Porzi, Lorenzo and Bul{\'o}, Samuel Rota and M{\"u}ller, Norman and Nie{\ss}ner, Matthias and Dai, Angela and Kontschieder, Peter},
  booktitle={Proceedings of the IEEE/CVF Conference on Computer Vision and Pattern Recognition},
  pages={9043--9052},
  year={2023}
}

@inproceedings{Gaussian_grouping,
  title={Gaussian grouping: Segment and edit anything in 3d scenes},
  author={Ye, Mingqiao and Danelljan, Martin and Yu, Fisher and Ke, Lei},
  booktitle={European Conference on Computer Vision},
  pages={162--179},
  year={2025},
  organization={Springer}
}

@inproceedings{Garfield,
  title={Garfield: Group anything with radiance fields},
  author={Kim, Chung Min and Wu, Mingxuan and Kerr, Justin and Goldberg, Ken and Tancik, Matthew and Kanazawa, Angjoo},
  booktitle={Proceedings of the IEEE/CVF Conference on Computer Vision and Pattern Recognition},
  pages={21530--21539},
  year={2024}
}

@inproceedings{Neural_fields,
  title={Neural fields in visual computing and beyond},
  author={Xie, Yiheng and Takikawa, Towaki and Saito, Shunsuke and Litany, Or and Yan, Shiqin and Khan, Numair and Tombari, Federico and Tompkin, James and Sitzmann, Vincent and Sridhar, Srinath},
  booktitle={Computer Graphics Forum},
  volume={41},
  number={2},
  pages={641--676},
  year={2022},
  organization={Wiley Online Library}
}

@inproceedings{Feature_3dgs,
  title={Feature 3dgs: Supercharging 3d gaussian splatting to enable distilled feature fields},
  author={Zhou, Shijie and Chang, Haoran and Jiang, Sicheng and Fan, Zhiwen and Zhu, Zehao and Xu, Dejia and Chari, Pradyumna and You, Suya and Wang, Zhangyang and Kadambi, Achuta},
  booktitle={Proceedings of the IEEE/CVF Conference on Computer Vision and Pattern Recognition},
  pages={21676--21685},
  year={2024}
}

@inproceedings{Langsplat,
  title={Langsplat: 3d language gaussian splatting},
  author={Qin, Minghan and Li, Wanhua and Zhou, Jiawei and Wang, Haoqian and Pfister, Hanspeter},
  booktitle={Proceedings of the IEEE/CVF Conference on Computer Vision and Pattern Recognition},
  pages={20051--20060},
  year={2024}
}

@inproceedings{Dynamic_3D_Gaussians,
  title={Dynamic 3D Gaussians: Tracking by Persistent Dynamic View Synthesis},
  author={Luiten, Jonathon and Kopanas, Georgios and Leibe, Bastian and Ramanan, Deva},
  booktitle={3DV},
  year={2024}
}

@inproceedings{efros1999texture,
  title={Texture synthesis by non-parametric sampling},
  author={Efros, Alexei A and Leung, Thomas K},
  booktitle={Proceedings of the IEEE/CVF International Conference on Computer Vision},
  volume={2},
  pages={1033--1038},
  year={1999},
  organization={IEEE}
}
}

\clearpage

\renewcommand{\thefigure}{R.\arabic{figure}}
\renewcommand{\thetable}{R.\arabic{table}}
\renewcommand{\thesection}{R.\arabic{section}}

\makeatletter
\newcommand*{\addFileDependency}[1]{%
  \typeout{(#1)}
  \@addtofilelist{#1}
  \IfFileExists{#1}{}{\typeout{No file #1.}}
}
\makeatother

\newcommand*{\myexternaldocument}[1]{%
    \externaldocument{#1}%
    \addFileDependency{#1.tex}%
    \addFileDependency{#1.aux}%
}

\begin{appendices}

\setcounter{figure}{0} 
\setcounter{table}{0} 
\setcounter{equation}{0} 

In the following, we provide additional discussion on ``3D Feature Fields'' (\cref{sec:more-related}), implementation details (\cref{sec:more-details}), more quantitative and qualitative results and analysis (\cref{sec:more-results}).
We also discuss the limitations of \featgs in \cref{sec:limitations}.
Please check the \vid for an overview of our framework and more results.

\section{3D Feature Fields}
\label{sec:more-related}

Beyond modeling the appearance, 3D neural fields~\cite{Neural_fields} (\eg, NeRF~\cite{NeRF}, 3DGS~\cite{3DGS}) can also model features, by aggregating 2D features extracted from multiple views into a 3D canonical frame.
The feature extractors can either be learned from data~\cite{Semantic_NeRF,Panoptic} in an end-to-end manner, or be off-the-shelf Visual Foundation Models (VFMs), such as DINO~\cite{DINO}, CLIP~\cite{CLIP}, Stable Diffusion~\cite{SD}, SAM~\cite{SAM,ravi2024sam2}, and LSeg~\cite{li2022languagedriven}.
Different VFMs equip the 3D feature field with various capabilities:  
CLIP, and LSeg, which connect language with images, are used by several works~\cite{lerf2023,Decomposing_NeRF,Feature_3dgs,Peng2023OpenScene,wang2021clip} to enable text-based querying and editing. 
SAM, which truly learns the concept of ``object'', has been used for grouping~\cite{Garfield,Gaussian_grouping}, segmentation~\cite{Feature_3dgs,Langsplat}, and 3D scene understanding~\cite{Gaussian_grouping,Feature_3dgs}. 
Meanwhile, the 3D feature fields distilled from DINO and SD show promising 
cross-instance and cross-frame 
consistency, as leveraged by FeatureNeRF~\cite{ye2023featurenerf} and N3F~\cite{tschernezki22neural} for various downstream tasks, such as keypoint transfer, co-segmentation, and video-based object retrieval. 
Additionally, DINO is also used by LERF~\cite{lerf2023} and DFFs~\cite{Decomposing_NeRF} to regularize CLIP features for finer decomposition.

What sets \featgs apart from these 3D feature field works, is their assumption of 3D-awareness and cross-view feature correspondence of VFMs, while \featgs questions this: Are they truly 3D-aware? If so, to what extent? Does the 3D-awareness come from color or shape? How can it be improved? 
\featgs provides a unified and neat analysis framework to address these questions, using VFM features for novel-view synthesis, instead of optimizing an additional 3D feature field to align with the 3D radiance field.

\section{Implementation Details}
\label{sec:more-details}

\featgs is implemented with PyTorch~\cite{paszke2019pytorch} and gsplat~\cite{gsplat}. 
For fair probing, images are resized to 512 for VFM feature extraction, reduced to 256 channels with PCA, then the feature map resolution is upsampled back to 512.
We use a 2-layer ReLU MLP for $g_{\Theta}$ with 256-dimensional hidden units. Adam optimizer is used to optimize the parameters of MLP, 3D Gaussians, and cameras.
At the warm start stage, we optimize the MLP parameters for 1K iterations with a learning rate that starts at $1\!\times\!10^{-2}$ and decays exponentially to $1\!\times\!10^{-4}$.
After this stage, optimization continues for another 7K iterations. 
We follow the learning rate strategy of vanilla 3DGS~\cite{3DGS}. For the MLP part, we maintain the original ratio but reduce the learning rate by an order of magnitude. 
To optimize the cameras, the learning rate starts at $1\!\times\!10^{-4}$ and decays exponentially to $1\!\times\!10^{-6}$ at 1K iteration.
The DUST3R~\cite{DUSt3R} checkpoint at 512 resolution initializes the point clouds and camera parameters. Photometric loss is computed at the original image resolution. Adaptive density control~\cite{3DGS} is omitted throughout the optimization process.
All experiments are conducted on a single NVIDIA GeForce RTX 4090 GPU.

\noindent \featgs evaluates a total of 11 models, as listed below:
\smallskip
\begin{itemize}
  \item \textbf{Raw Image Feature}. IUVRGB includes image index (I), pixel coordinates (UV), and colors (RGB), serving as a baseline for comparison.
  \item \textbf{Supervised 3D VFMs}. DUST3R~\cite{DUSt3R}, MASt3R~\cite{MASt3R} and MiDaS~\cite{MiDaS} are trained with pointmap regression, matching, and depth estimation objective using 3D datasets.
  \item \textbf{Self-supervised 2D VFMs}. DINO~\cite{DINO} and DINOv2~\cite{DINOv2} are trained with discriminative self-supervised objective using 2D datasets without annotations.
  \item \textbf{Supervised 2D VFMs}. SAM~\cite{SAM} and CLIP~\cite{CLIP} are trained with segmentation and contrastive objective using 2D datasets and corresponding annotations.
  \item \textbf{Distilled 2D VFMs}. RADIO~\cite{RADIO} merged DINOv2, SAM, and CLIP via model distillation on 2D data.
  \item \textbf{Image-reconstruction-based 2D VFMs}. MAE~\cite{MAE} and Stable Diffusion (SD)~\cite{SD} are trained with Mean Square Error (MSE) and denoising objective using 2D datasets to reconstruct images in pixel and feature space.
\end{itemize}

\section{Additional Results}
\label{sec:more-results}

\begin{figure*}[t!]
    \centering
    \captionsetup{type=figure}
    \begin{subfigure}[b]{0.33\linewidth}
        \centering
        \includegraphics[width=\linewidth,page=1]{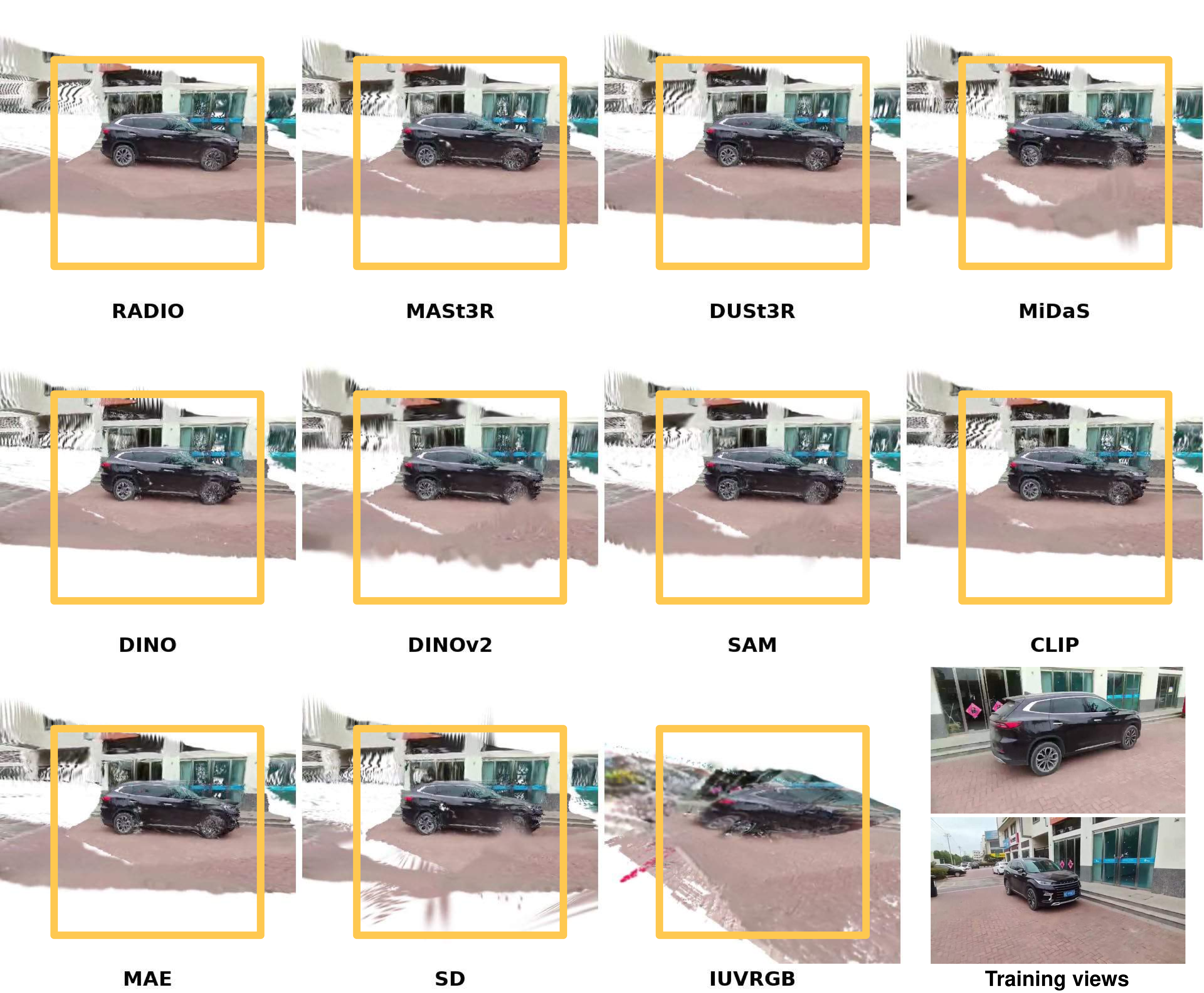}
    \end{subfigure}
    \begin{subfigure}[b]{0.33\linewidth}
        \centering
        \includegraphics[width=\linewidth,page=1]{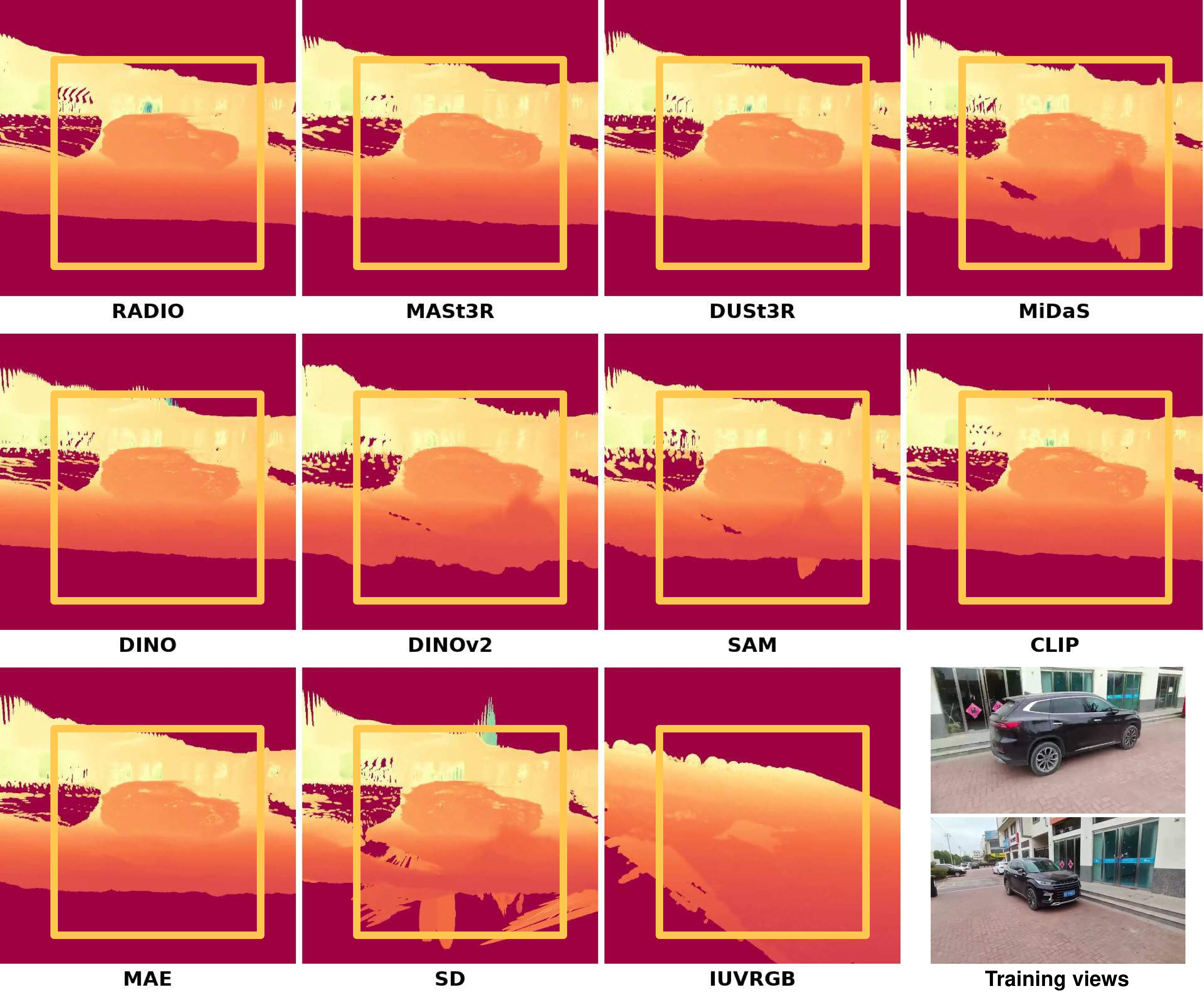}
    \end{subfigure}
    \begin{subfigure}[b]{0.33\linewidth}
        \centering
        \includegraphics[width=\linewidth,page=1]{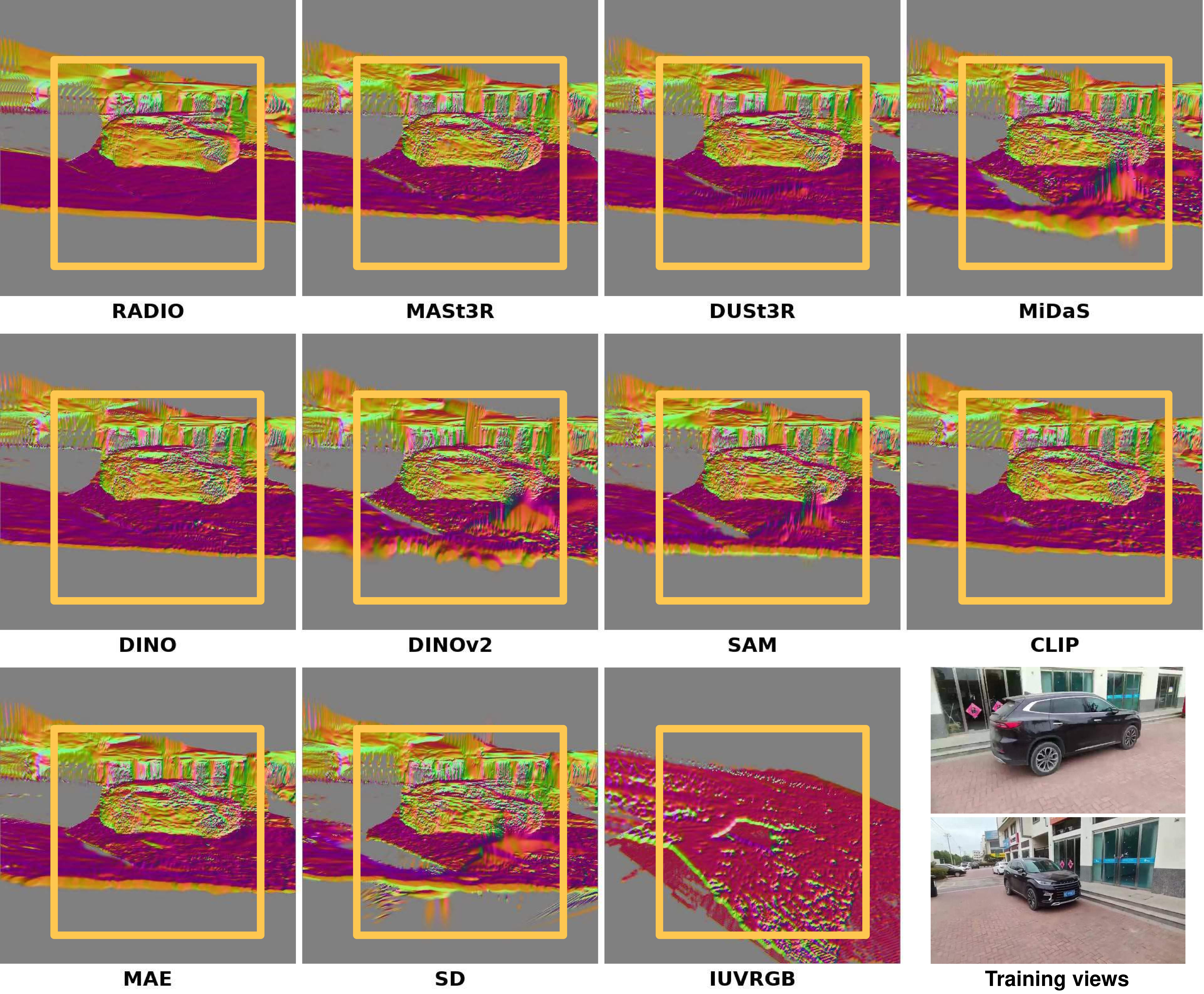}
    \end{subfigure}

    \begin{subfigure}[b]{0.33\linewidth}
        \centering
        \includegraphics[width=\linewidth,page=1]{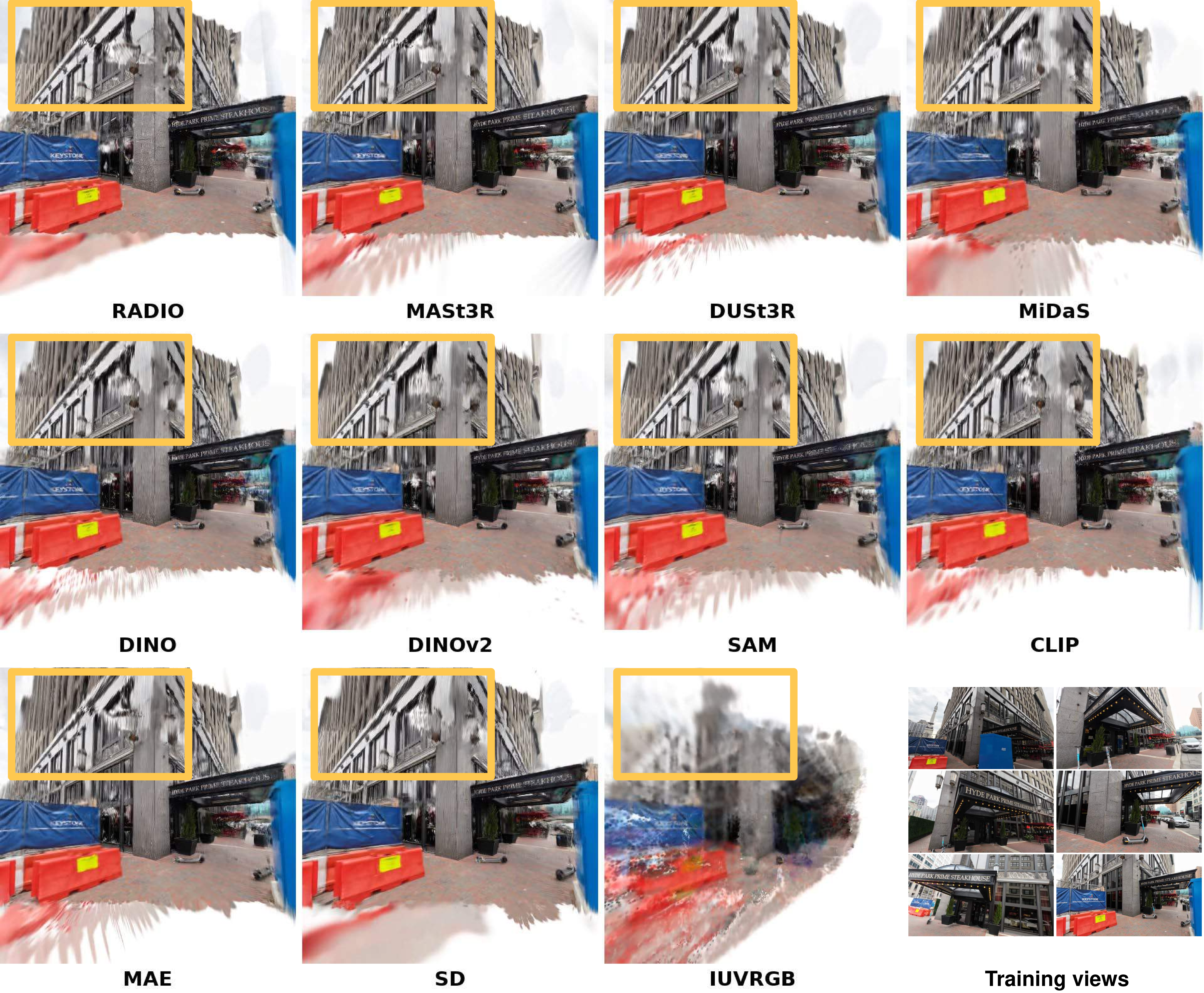}
        \caption{RGB Renderings}
        \label{fig:G_center_rgb}
    \end{subfigure}
    \begin{subfigure}[b]{0.33\linewidth}
        \centering
        \includegraphics[width=\linewidth,page=1]{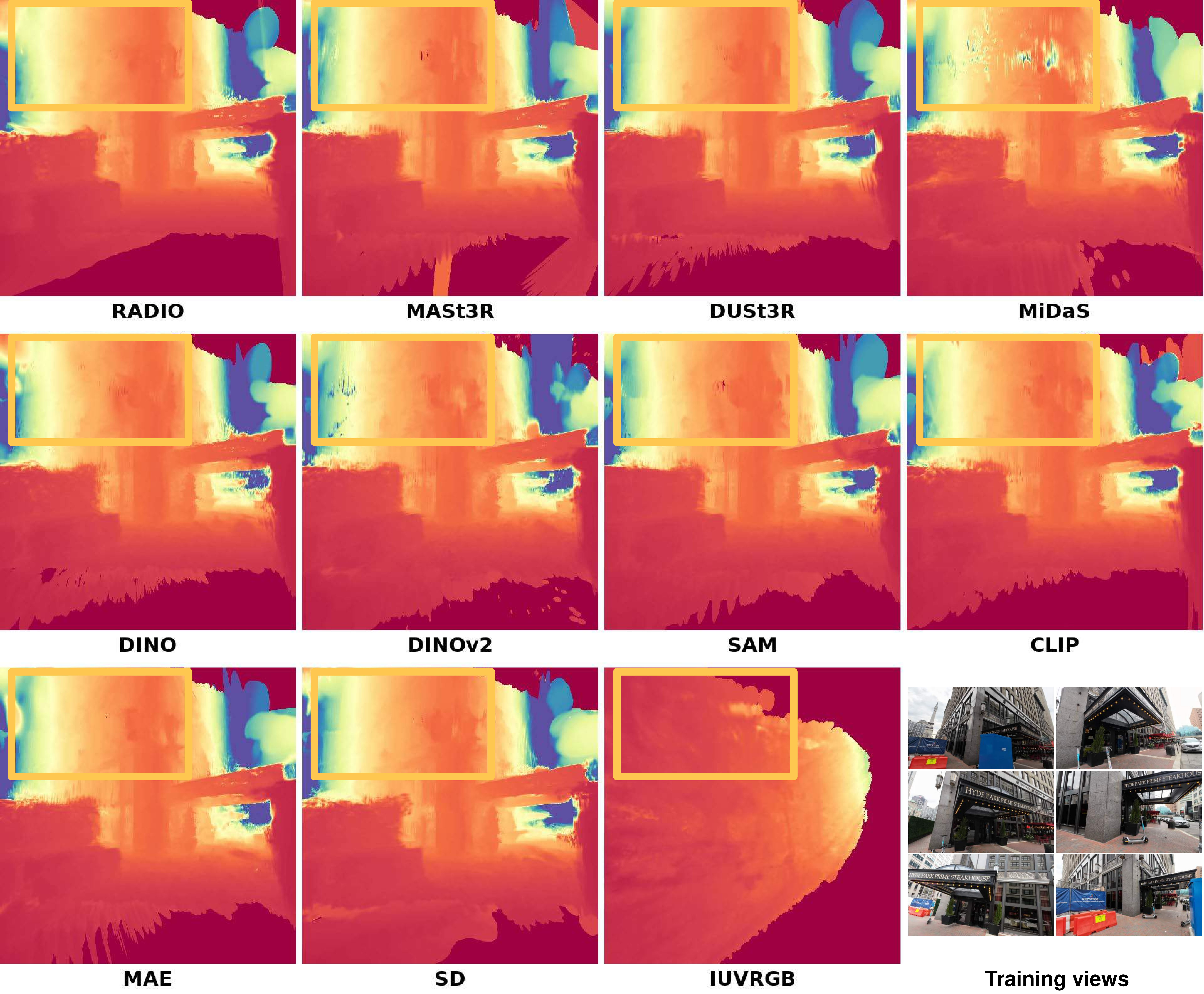}
        \caption{Expected Depth Renderings}
        \label{fig:G_center_depth}
    \end{subfigure}
    \begin{subfigure}[b]{0.33\linewidth}
        \centering
        \includegraphics[width=\linewidth,page=1]{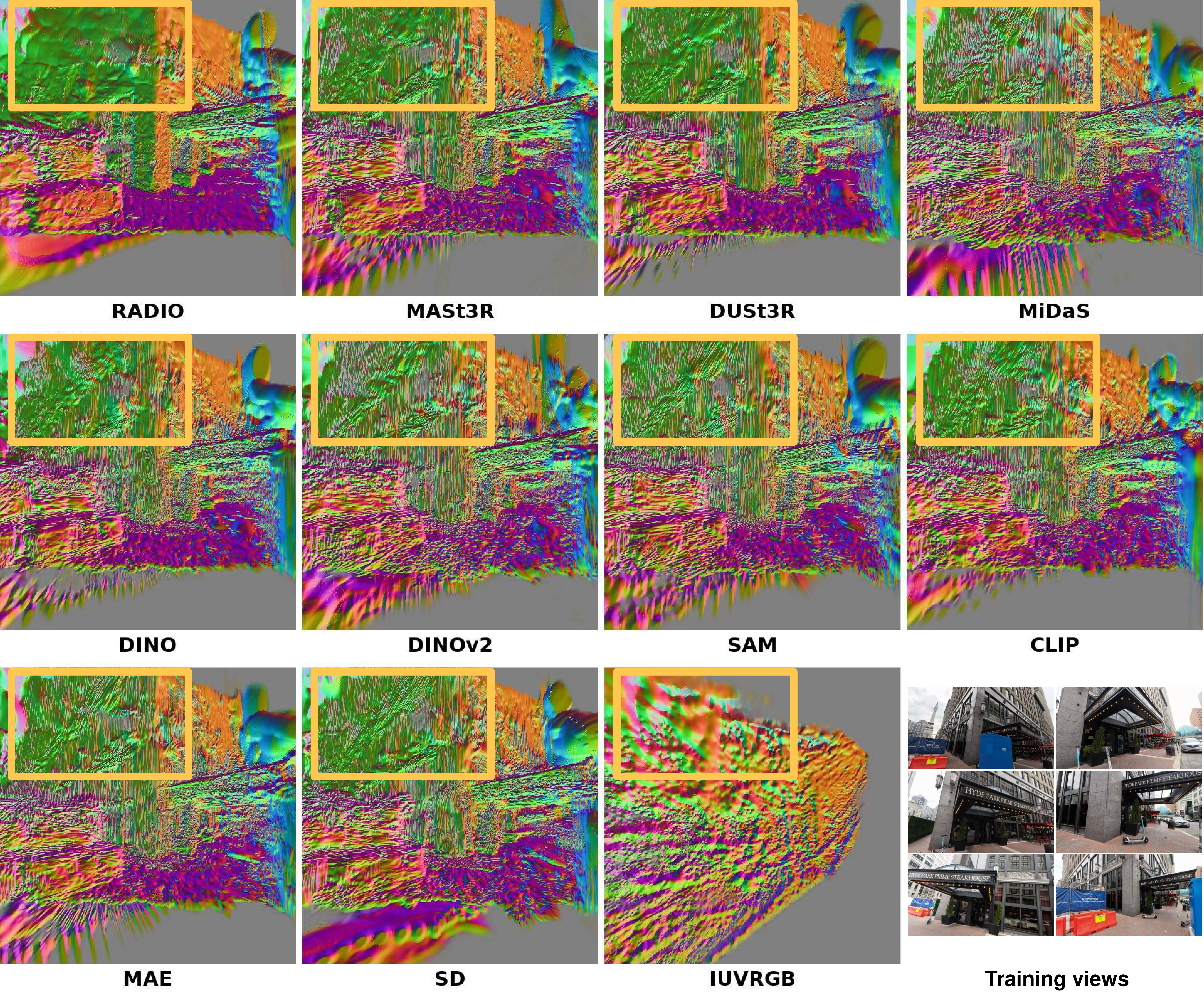}
        \caption{Expected Normal Renderings}
        \label{fig:G_center_normal}
    \end{subfigure}
    \caption{{\bf Novel View Synthesis of RGB Correlates with Depth and Normal.} We present qualitative examples, including RGB renderings, expected depth renderings, and expected normal renderings, of \gmode with different VFMs. This demonstrates that the NVS quality of \featgs probing results closely aligns with 2.5D metrics.}
    \label{fig:G_suv}

\end{figure*}

\begin{figure*}[t!]
    \centering
    \includegraphics[width=1\linewidth,page=1]{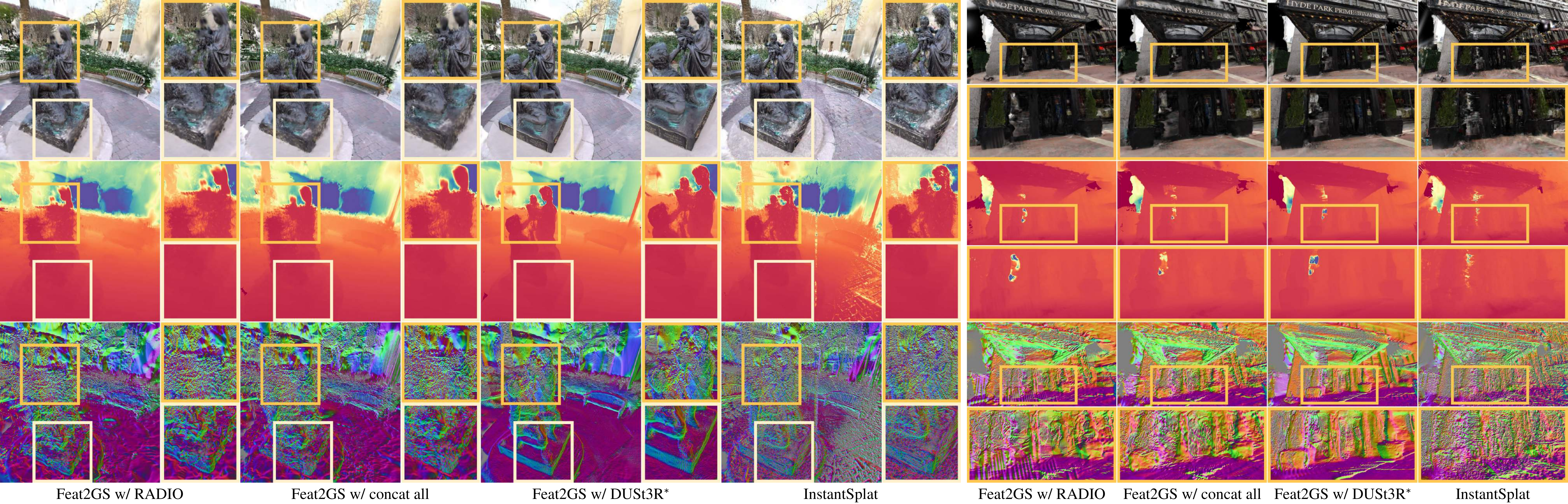}
    \caption{{\bf Novel View Synthesis of RGB Correlates with Depth and Normal.} We show qualitative examples, including RGB renderings, expected depth renderings, and expected normal renderings, of \gmode with different \featgs application baselines: feature pickup (\featgs w/ RADIO), feature ensembling (\featgs w/ concat all), and feature fine-tuning (\featgs w/ DUSt3R$^*$). This demonstrates that the NVS quality of \featgs application baselines closely aligns with 2.5D metrics.}
    \label{fig:baseline-sup}
  \end{figure*}

\pheading{Visualization of Depth and Normal.} 
In \cref{sec:results}, we identify the top four performers in \gmode as \radio $>$ \mastr $>$ \dustr $>$ \dino, while Stable Diffusion (SD) performs the worst, exhibiting broken geometry. We then present qualitative results of geometry with expected depth and normal rendering in \cref{fig:G_suv}. Additionally, we show the 2.5D renderings of \featgs application baselines in \cref{fig:baseline-sup}, both illustrating the strong correlation between NVS and depth/normal 2.5D metrics.

\captionsetup{type=figure}
\begin{figure}[t!]
  \centering
  \includegraphics[width=1\linewidth,page=1]{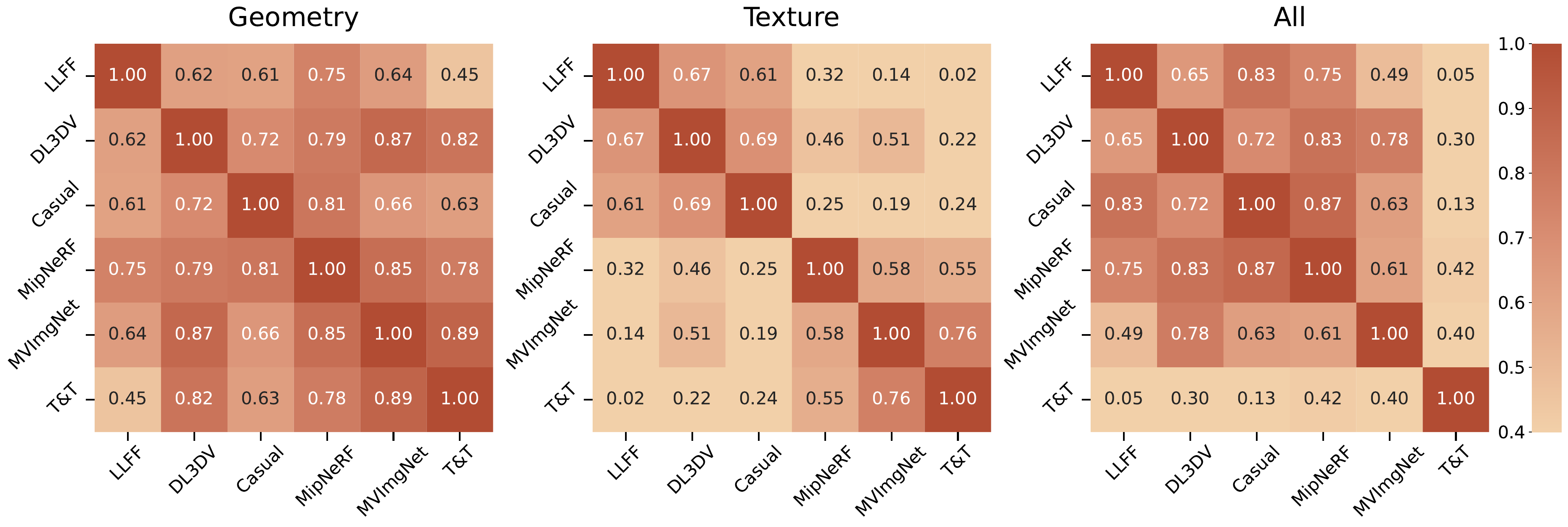}
  \vspace{-18pt}
  \caption{{\bf Performance correlations across different datasets.}
  In \tmode, the first three datasets (indoor, small-scale) and last three (outdoor, large-scale) show internal correlations but little between groups. In \amode, the T\&T dataset, with the highest complexity and widest view range, correlates minimally with others. These variations highlight the necessity of evaluating on diverse data. \featgs, removes 3D ground truth requirements, enabling diverse capture evaluation and reducing bias.
  }
  \label{fig:correlation_dataset}
  \vspace{-6pt}
\end{figure}

\begin{figure}[t!]
  \centering
  \includegraphics[width=1\linewidth,page=1]{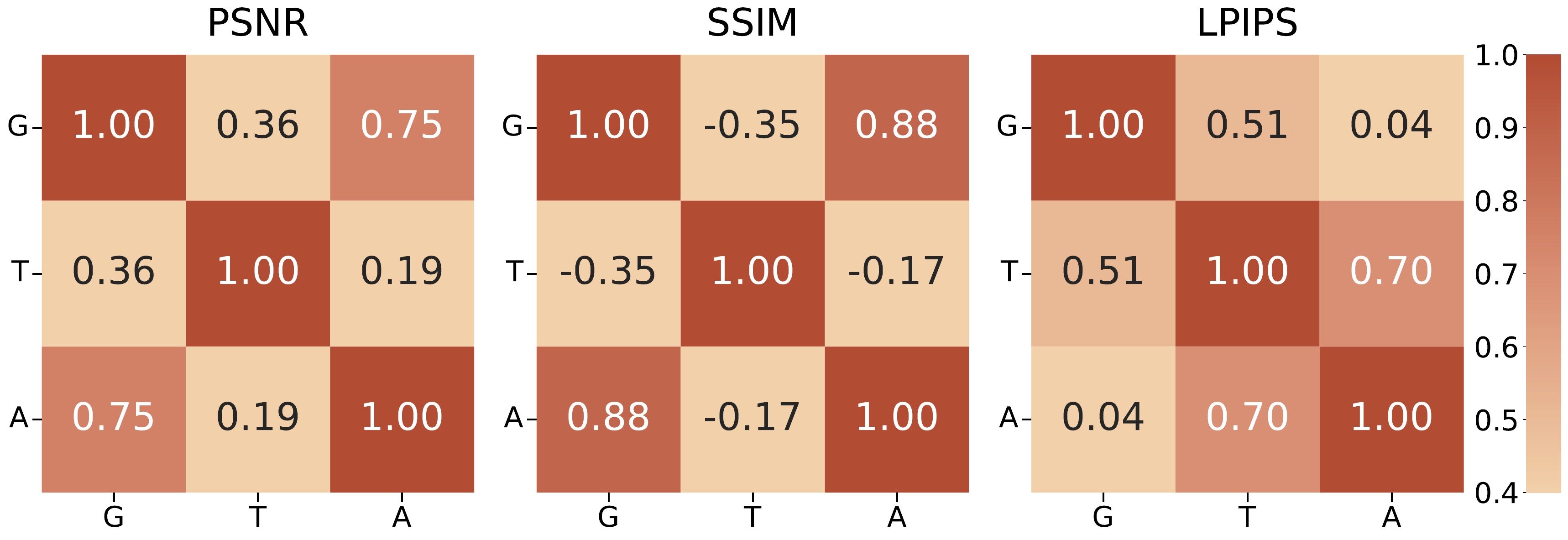}
  \vspace{-18pt}
  \caption{{\bf Performance Correlations of GTA across All Datasets.} 
  The \amode correlates strongly with \gmode in PSNR and SSIM (primarily reflect structural consistency), and is closely related to \tmode in LPIPS (commonly used to assess image sharpness), suggesting an optimal \amode depends on both high-performing \textbf{G}eometry and \tmode.}
  \label{fig:correlation}
  \vspace{-4pt}
\end{figure}

\begin{figure}[t!]
    \centering
    \includegraphics[width=1\linewidth,page=1]{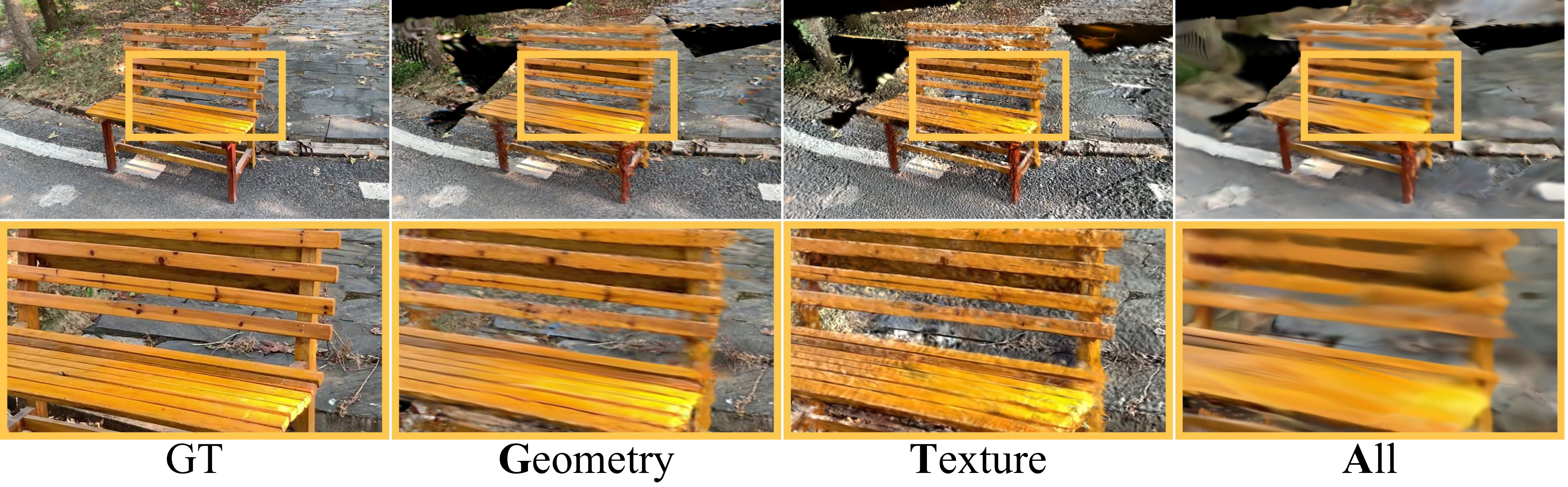}
    \vspace{-18pt}
    \caption{{\bf GTA Modes Comparison for the Same Region.} We present novel view synthesis of GTA modes using RADIO features. \tmode shows broken structures as it excludes VFM features for 3DGS geometry regression, while \amode is blurrier than \gmode due to reliance on VFM features for color regression. This highlights that the blurriness in the \amode arises from the lack of texture awareness in VFMs.
    }
    \vspace{-8pt}
    \label{fig:GTA_same}
  \end{figure}

\setlength{\tabcolsep}{4pt}
\begin{table}[t]
\centering
\resizebox{1\columnwidth}{!}{
\begin{tabular}{l|>{\raggedleft\arraybackslash}p{0.85cm}>{\raggedleft\arraybackslash}p{0.85cm}>{\raggedleft\arraybackslash}p{0.85cm}|>{\raggedleft\arraybackslash}p{0.85cm}>{\raggedleft\arraybackslash}p{0.85cm}>{\raggedleft\arraybackslash}p{0.85cm}|>{\raggedleft\arraybackslash}p{0.85cm}>{\raggedleft\arraybackslash}p{0.85cm}>{\raggedleft\arraybackslash}p{0.85cm}}
\toprule
\multicolumn{1}{c|}{} & \multicolumn{9}{c}{All Datasets} \\
\midrule
\multicolumn{1}{c|}{} & \multicolumn{3}{c|}{\textbf{G}eometry} & \multicolumn{3}{c|}{\textbf{T}exture} & \multicolumn{3}{c}{\textbf{A}ll} \\
\midrule
Feature    & \fontsize{8.5pt}{9pt}\selectfont{PSNR$\uparrow$} & \fontsize{8.5pt}{9pt}\selectfont{SSIM$\uparrow$} & \fontsize{8.5pt}{9pt}\selectfont{LPIPS$\downarrow$} & \fontsize{8.5pt}{9pt}\selectfont{PSNR$\uparrow$} & \fontsize{8.5pt}{9pt}\selectfont{SSIM$\uparrow$} & \fontsize{8.5pt}{9pt}\selectfont{LPIPS$\downarrow$} & \fontsize{8.5pt}{9pt}\selectfont{PSNR$\uparrow$} & \fontsize{8.5pt}{9pt}\selectfont{SSIM$\uparrow$} & \fontsize{8.5pt}{9pt}\selectfont{LPIPS$\downarrow$} \\
\midrule
DINOv2     &             \cellcolor[rgb]{1.00,0.82,0.82}19.59 &             \cellcolor[rgb]{1.00,0.82,0.82}.6406 &                \cellcolor[rgb]{1.00,0.82,0.82}.3364 &             \cellcolor[rgb]{1.00,0.82,0.82}18.03 &             \cellcolor[rgb]{1.00,0.94,0.82}.5951 &                \cellcolor[rgb]{1.00,0.94,0.82}.3291 &             \cellcolor[rgb]{1.00,0.82,0.82}19.50 &             \cellcolor[rgb]{1.00,0.82,0.82}.6388 &                \cellcolor[rgb]{1.00,0.94,0.82}.3760 \\
DINOv2$^+$ &             \cellcolor[rgb]{1.00,0.94,0.82}19.67 &             \cellcolor[rgb]{1.00,0.94,0.82}.6480 &                \cellcolor[rgb]{1.00,0.94,0.82}.3202 &             \cellcolor[rgb]{1.00,0.94,0.82}18.10 &             \cellcolor[rgb]{1.00,0.82,0.82}.5950 &                \cellcolor[rgb]{1.00,0.94,0.82}.3291 &             \cellcolor[rgb]{1.00,0.94,0.82}19.58 &             \cellcolor[rgb]{1.00,0.94,0.82}.6443 &                \cellcolor[rgb]{1.00,0.82,0.82}.3894 \\
DINOv2$^*$ &             \cellcolor[rgb]{0.56,0.75,0.38}19.78 &             \cellcolor[rgb]{0.56,0.75,0.38}.6552 &                \cellcolor[rgb]{0.56,0.75,0.38}.2962 &             \cellcolor[rgb]{0.56,0.75,0.38}18.18 &             \cellcolor[rgb]{0.56,0.75,0.38}.5968 &                \cellcolor[rgb]{0.56,0.75,0.38}.3232 &             \cellcolor[rgb]{0.56,0.75,0.38}19.80 &             \cellcolor[rgb]{0.56,0.75,0.38}.6614 &                \cellcolor[rgb]{0.56,0.75,0.38}.3247 \\
\midrule
DINO       &             \cellcolor[rgb]{1.00,0.82,0.82}19.63 &             \cellcolor[rgb]{1.00,0.82,0.82}.6452 &                \cellcolor[rgb]{1.00,0.82,0.82}.3256 &             \cellcolor[rgb]{1.00,0.92,0.82}18.03 &             \cellcolor[rgb]{0.56,0.75,0.38}.5961 &                \cellcolor[rgb]{1.00,0.94,0.82}.3282 &             \cellcolor[rgb]{1.00,0.82,0.82}19.55 &             \cellcolor[rgb]{1.00,0.82,0.82}.6427 &                \cellcolor[rgb]{1.00,0.94,0.82}.3793 \\
DINO$^+$   &             \cellcolor[rgb]{1.00,0.94,0.82}19.72 &             \cellcolor[rgb]{1.00,0.94,0.82}.6485 &                \cellcolor[rgb]{1.00,0.94,0.82}.3207 &             \cellcolor[rgb]{1.00,0.92,0.82}18.03 &             \cellcolor[rgb]{1.00,0.82,0.82}.5941 &                \cellcolor[rgb]{1.00,0.82,0.82}.3291 &             \cellcolor[rgb]{1.00,0.94,0.82}19.64 &             \cellcolor[rgb]{1.00,0.94,0.82}.6465 &                \cellcolor[rgb]{1.00,0.82,0.82}.3839 \\
DINO$^*$   &             \cellcolor[rgb]{0.56,0.75,0.38}19.74 &             \cellcolor[rgb]{0.56,0.75,0.38}.6557 &                \cellcolor[rgb]{0.56,0.75,0.38}.2918 &             \cellcolor[rgb]{0.56,0.75,0.38}18.09 &             \cellcolor[rgb]{1.00,0.94,0.82}.5949 &                \cellcolor[rgb]{0.56,0.75,0.38}.3235 &             \cellcolor[rgb]{0.56,0.75,0.38}19.69 &             \cellcolor[rgb]{0.56,0.75,0.38}.6630 &                \cellcolor[rgb]{0.56,0.75,0.38}.3154 \\
\midrule
CLIP       &             \cellcolor[rgb]{1.00,0.82,0.82}19.61 &             \cellcolor[rgb]{1.00,0.82,0.82}.6436 &                \cellcolor[rgb]{1.00,0.82,0.82}.3331 &             \cellcolor[rgb]{1.00,0.94,0.82}18.10 &             \cellcolor[rgb]{1.00,0.94,0.82}.5947 &                \cellcolor[rgb]{1.00,0.82,0.82}.3289 &             \cellcolor[rgb]{1.00,0.82,0.82}19.50 &             \cellcolor[rgb]{1.00,0.82,0.82}.6416 &                \cellcolor[rgb]{1.00,0.94,0.82}.3832 \\
CLIP$^+$   &             \cellcolor[rgb]{1.00,0.94,0.82}19.68 &             \cellcolor[rgb]{1.00,0.94,0.82}.6466 &                \cellcolor[rgb]{1.00,0.94,0.82}.3222 &             \cellcolor[rgb]{1.00,0.82,0.82}18.09 &             \cellcolor[rgb]{1.00,0.82,0.82}.5941 &                \cellcolor[rgb]{1.00,0.94,0.82}.3286 &             \cellcolor[rgb]{1.00,0.94,0.82}19.63 &             \cellcolor[rgb]{1.00,0.94,0.82}.6468 &                \cellcolor[rgb]{1.00,0.82,0.82}.3842 \\
CLIP$^*$   &             \cellcolor[rgb]{0.56,0.75,0.38}19.70 &             \cellcolor[rgb]{0.56,0.75,0.38}.6540 &                \cellcolor[rgb]{0.56,0.75,0.38}.2959 &             \cellcolor[rgb]{0.56,0.75,0.38}18.19 &             \cellcolor[rgb]{0.56,0.75,0.38}.5962 &                \cellcolor[rgb]{0.56,0.75,0.38}.3242 &             \cellcolor[rgb]{0.56,0.75,0.38}19.67 &             \cellcolor[rgb]{0.56,0.75,0.38}.6599 &                \cellcolor[rgb]{0.56,0.75,0.38}.3199 \\
\bottomrule
\end{tabular}
}
\caption{
{\bf Feature Upsampling$^+$ \vs Fine-tuning$^*$.} We report a quantitative comparison of \featgs application baselines between feature upsampling using the recent VFM feature upsampler~\cite{featup} and feature fine-tuning during the warm-start stage. While feature upsampling offers some benefits, fine-tuning achieves significantly higher improvement, particularly in the LPIPS metric.
}
\label{tab:feat_up_ft}
\end{table}

\begin{figure}[t]
    \centering
    \includegraphics[width=1\linewidth,page=1]{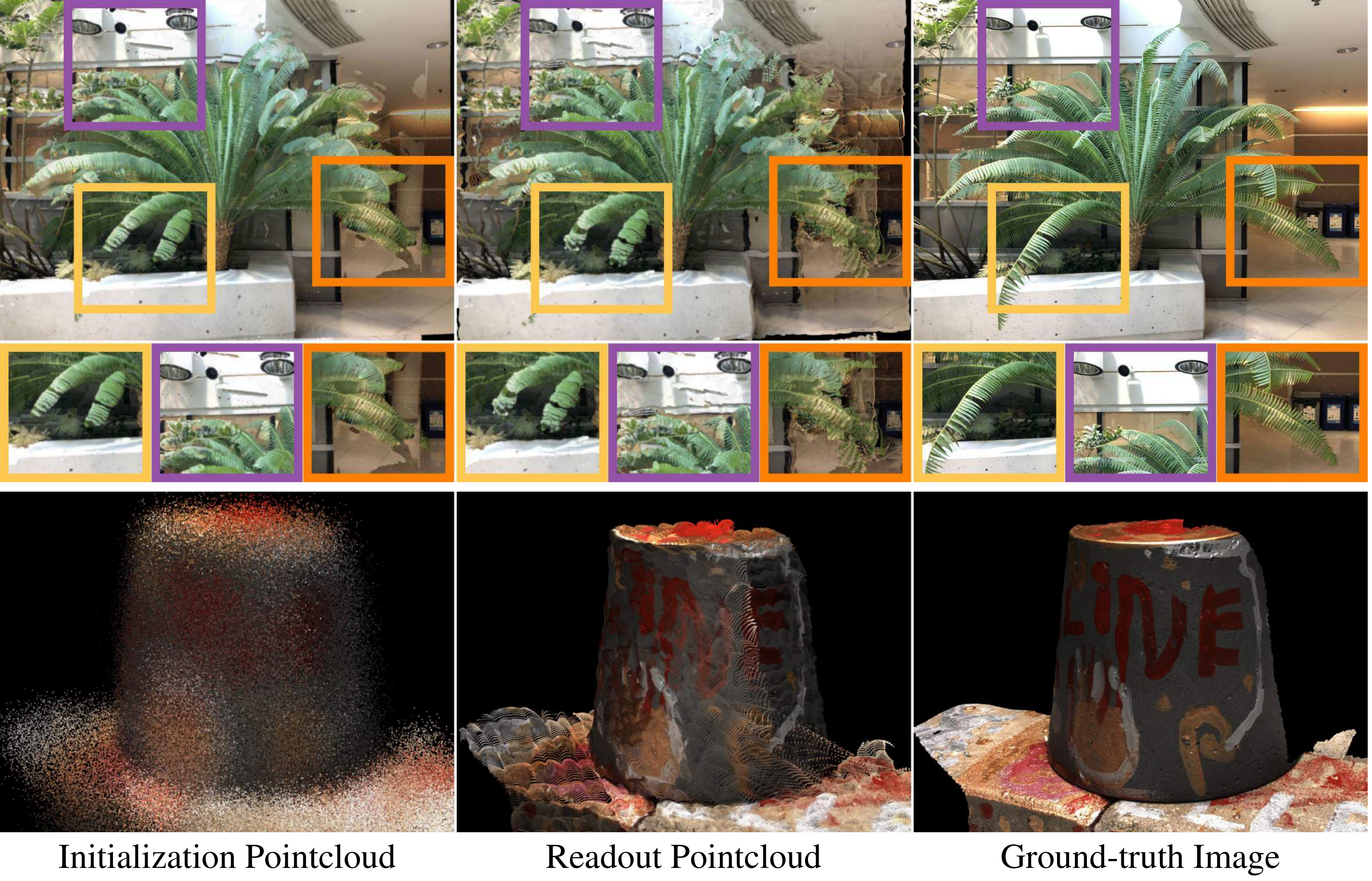}
    \vspace{-16pt}
    \caption{{\bf Failure Case.} \featgs can handle noisy initialization pointcloud (bottom row), but it struggles when the initialization pointcloud contains significant outliers (top row), \eg, severely displaced branches (yellow), misplaced lamps (purple), and missing parts of branchs (orange). These prevent plausible pointcloud readouts, even with the best geometry-aware VFM feature, RADIO.
    }
    \vspace{-6pt}
    \label{fig:limitation}
  \end{figure}

\begin{figure*}[t!]
    \centering
    \includegraphics[trim={0 2cm 0 0}, clip, width=1\linewidth,page=1]{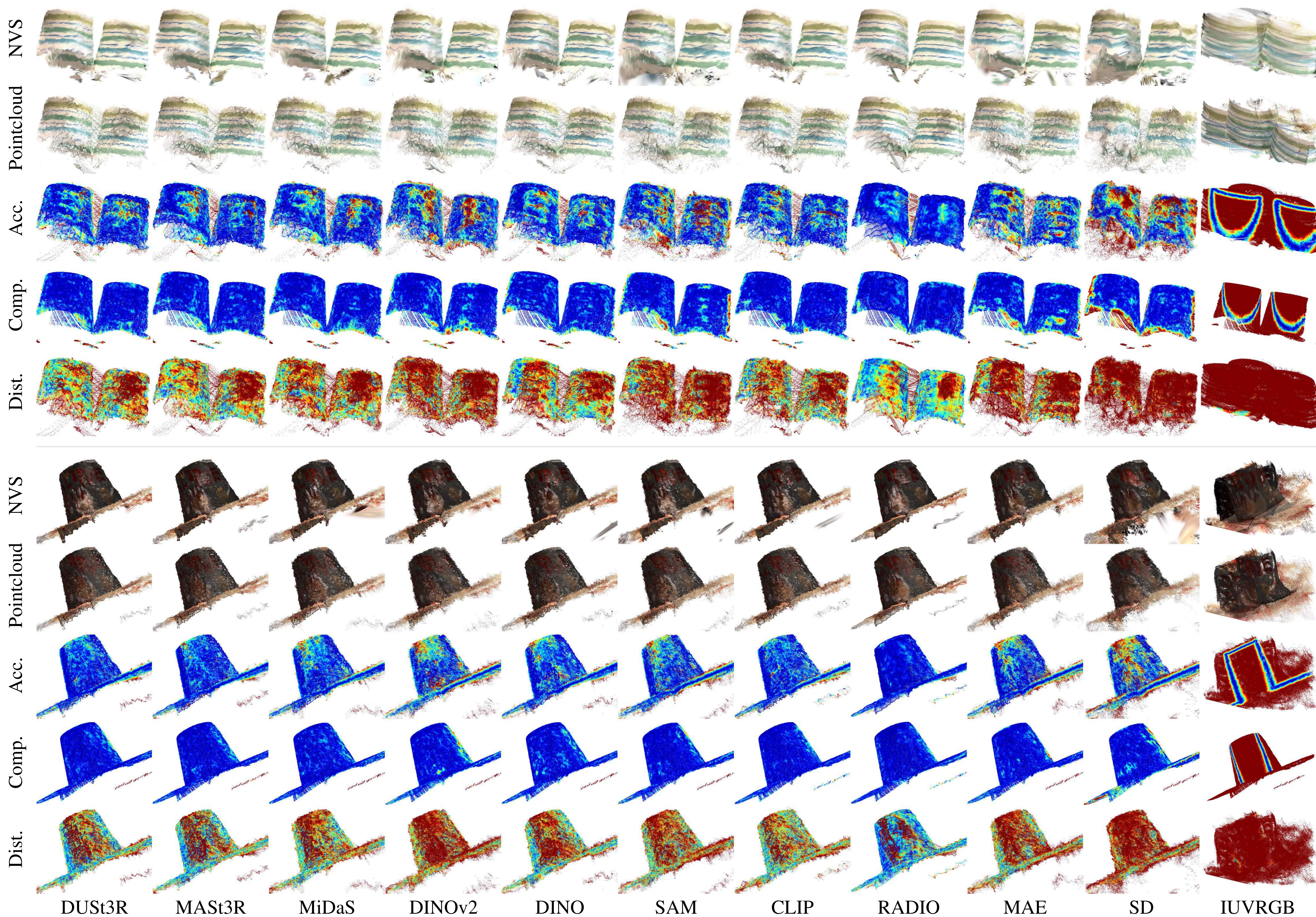}
    \scriptsize
    \setlength{\tabcolsep}{0.0pt}     %
    \begin{tabularx}{\linewidth}{
      >{\centering\arraybackslash}m{0.025\linewidth}
      >{\centering\arraybackslash}m{0.08864\linewidth}
      >{\centering\arraybackslash}m{0.08864\linewidth}
      >{\centering\arraybackslash}m{0.08864\linewidth}
      >{\centering\arraybackslash}m{0.08864\linewidth}
      >{\centering\arraybackslash}m{0.08864\linewidth}
      >{\centering\arraybackslash}m{0.08864\linewidth}
      >{\centering\arraybackslash}m{0.08864\linewidth}
      >{\centering\arraybackslash}m{0.08864\linewidth}
      >{\centering\arraybackslash}m{0.08864\linewidth}
      >{\centering\arraybackslash}m{0.08864\linewidth}
      >{\centering\arraybackslash}m{0.08864\linewidth}
      }
      \quad & \dustr~\cite{DUSt3R} & \mastr~\cite{MASt3R} & \midas~\cite{MiDaS} & \dinotwo~\cite{DINOv2} & \dino~\cite{DINO} & \sam~\cite{SAM} & CLIP~\cite{CLIP} & \radio~\cite{RADIO} & \mae~\cite{MAE} & SD~\cite{SD} & IUVRGB
   \end{tabularx}
    \vspace{-8pt}
    \caption{{\bf Novel View Synthesis as Proxy Task to Assess 3D.} We present qualitative examples from the DTU dataset, including NVS, Pointcloud (readout 3DGS positions), Accuracy (smallest distance from a readout point to ground-truth), Completeness (smallest distance from a ground-truth point to a readout point), and Distance (based on ground-truth point matching).
    Results show that NVS quality aligns with 3D metrics, proving its reliability as an indicator for 3D assessment.
    RADIO performs \textcolor{bestgreen}{best}, SD \textcolor{worstred}{worst}, with IUVRGB as a reference.
    \faSearch~\textbf{Zoom in} or check our \dtu to see more details.
    }
    \label{fig:supp_metric3d}
    \vspace{-18pt}
  \end{figure*}

\begin{figure*}[t!]
    \centering
    \includegraphics[width=1\linewidth,page=1]{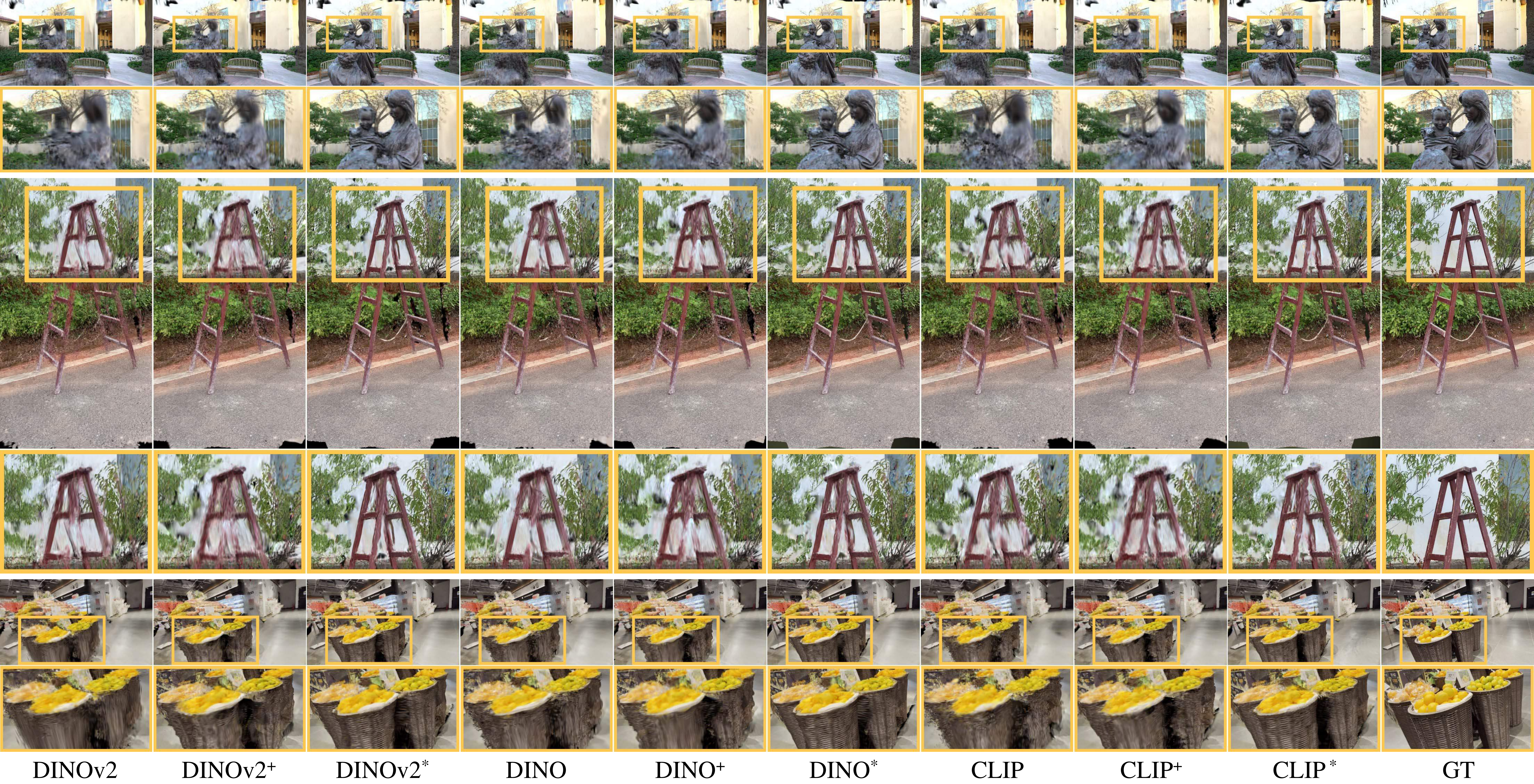}
    \vspace{-20pt}
    \caption{
    {\bf Feature Upsampling$^+$ \vs Fine-tuning$^*$.} We compare \featgs application baselines between feature upsampling using the recent VFM feature upsampler~\cite{featup} and feature fine-tuning during the warm-start stage. While feature upsampling improves the spatial resolution of features, feature fine-tuning provides greater details.
    Similar fine-tuning performance across different VFM features show that fine-tuning enriches embedded information, enabling high-quality reconstruction with any well-initialized features.
    }
    \vspace{-8pt}
    \label{fig:featup_ft}
  \end{figure*}

\pheading{Feature Upsampling \vs Fine-tuning.}
As discussed in \cref{sec:application}, the low-resolution features extracted from VFM encoders limit \featgs application baselines in rendering high-frequency details. We then compare two solutions to address this: feature upsampling (using VFM feature upsampler~\cite{featup} to improve the feature resolution) and feature fine-tuning (optimizing features during the warm-start stage). As shown in \cref{fig:featup_ft} and \cref{tab:feat_up_ft}, upsampling offers little improvement, while feature fine-tuning yields significantly better results.
Similar fine-tuning performance across various VFM features shows that fine-tuning increases resolution and enriches embedded information, allowing high-quality reconstruction with any well-initialized features.

\begin{figure*}[t!]
    \centering
    \captionsetup{type=figure}
    \begin{subfigure}[b]{0.49\linewidth}
        \centering
        \includegraphics[width=\linewidth,page=1]{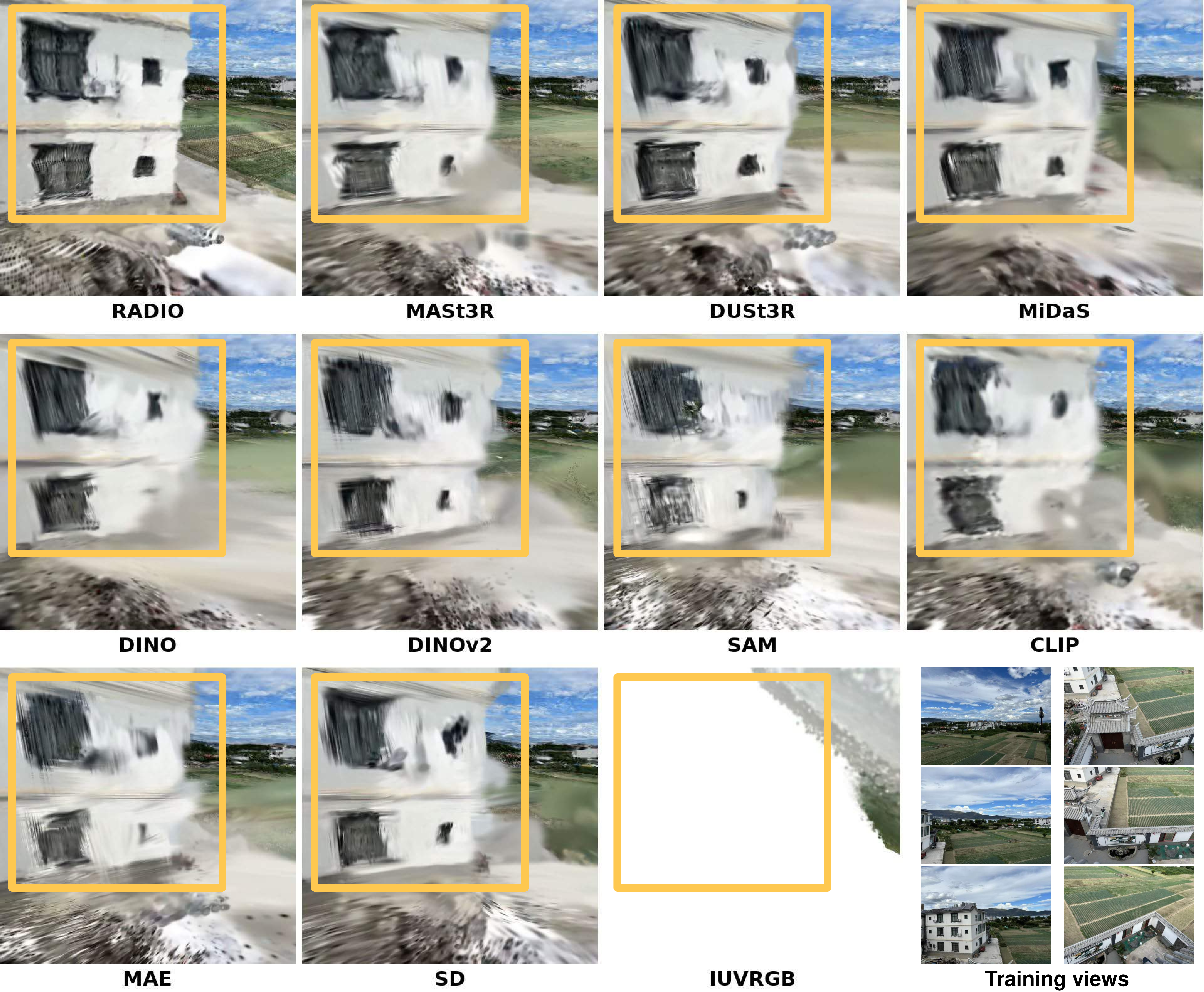}
        \caption{Position $\boldsymbol{x}$ Awareness}
    \end{subfigure}
    \begin{subfigure}[b]{0.49\linewidth}
        \centering
        \includegraphics[width=\linewidth,page=1]{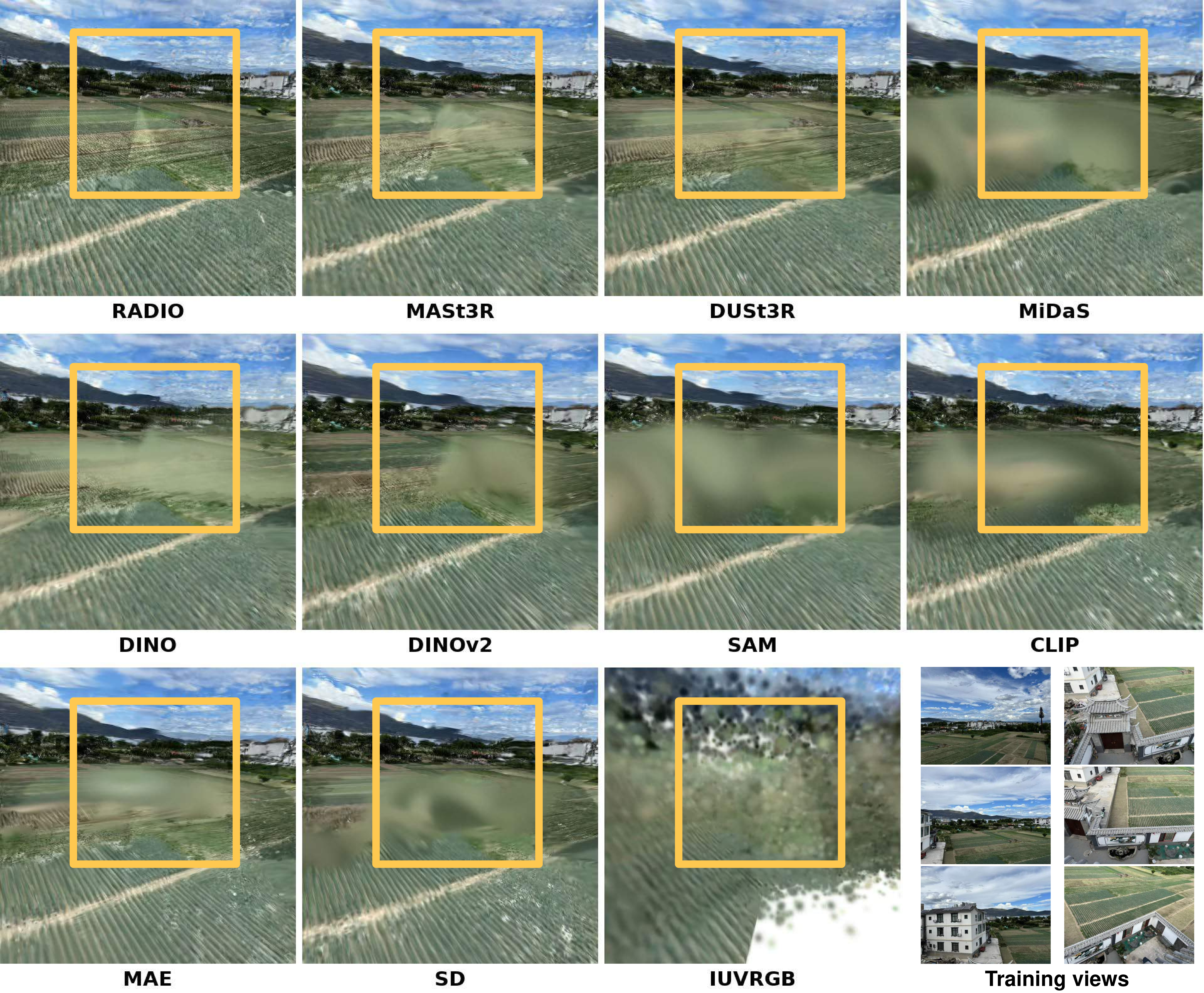}
        \caption{Covariance $\Sigma$ Awareness}
    \end{subfigure}
    \vspace{-10pt}
    \caption{{\bf Two Geometry Awareness Attributes.} We illustrate different attributes of geometry awareness. Position $\boldsymbol{x}$ awareness is typically reflected in the sharpness of readout details and edges, while covariance $\Sigma$ awareness is often observed in the flatness of planes.
    }
    \vspace{-8pt}
    \label{fig:erhai}
\end{figure*}

\pheading{Visualization of Two Geometry Awareness Attributes.} We provide qualitative examples in \cref{fig:erhai} to visualize two attributes of geometry awareness. Position $\boldsymbol{x}$ awareness is highlighted by the sharpness of readout details and edges, while covariance $\Sigma$ awareness reflects in plane flatness.

\pheading{Zero123 Outperforms SD in Objaverse-like Scenes.}
While Stable Diffusion (SD) performs poorly in most metrics due to its lack of multi-view consistency, does Zero123, which fine-tunes SD on Objaverse~\cite{Objaverse} multi-view dataset, achieve better cross-view consistency? As shown in \cref{tab:zero123vsSD}, Zero123 excels in Objaverse-like simple scenes (LLFF) but struggles with complex scenarios (Tanks and Temples), which might be attributed to catastrophic forgetting~\cite{forgetting}.

\pheading{\dino captures geometry well but PE artifacts hinder.} 
In \cref{sec:results}, we observe that \dino features capture geometry well, completely reconstructing the vehicle front (\cref{fig:GTA}) and wheel (\cref{fig:Geometry}).
In contrast, \dinotwo exhibits floating artifacts and distorted structures, likely caused by positional embedding (PE) artifacts noted in recent research~\cite{DVT,EmerNeRF}, as shown in \cref{fig:features}.
We observe that the artifacts in \dinotwo's features lead to degraded performance—an issue that becomes apparent when using a 2-layer MLP but is masked by the DPT head~\cite{DPT} utilized in prior work~\cite{Probe3D}. This explains why \dino outperforms \dinotwo in \featgs, but the opposite occurs in \probe~\cite{Probe3D} and suggests that while DPT can mitigate this issue, it persists and requires solutions such as registration~\cite{Registers}, denoising~\cite{DVT}, and 3D-aware training~\cite{fit3d,ConDense} to be fundamentally addressed.

\pheading{Texture Benefits from Image-Matching-Based Training.}
Both DUSt3R and MASt3R utilize CroCo~\cite{Croco}, pre-trained through cross-view completion similar to MAE~\cite{MAE}, enabling DUSt3R and MASt3R to exhibit texture awareness. 
But why does \mastr outperform DUSt3R on \tmode (see~\cref{fig:GTA})?
One possible explanation is that \mastr incorporates an additional image matching loss, promoting better awareness of fine-grained textures.

\setlength{\tabcolsep}{8pt}
\begin{table*}[t!]
\setlength{\tabcolsep}{4pt}
\centering
\resizebox{\textwidth}{!}{
\begin{tabular}{l|>{\raggedleft\arraybackslash}p{0.9cm}>{\raggedleft\arraybackslash}p{0.9cm}>{\raggedleft\arraybackslash}p{0.9cm}|>{\raggedleft\arraybackslash}p{0.9cm}>{\raggedleft\arraybackslash}p{0.9cm}>{\raggedleft\arraybackslash}p{0.9cm}|>{\raggedleft\arraybackslash}p{0.9cm}>{\raggedleft\arraybackslash}p{0.9cm}>{\raggedleft\arraybackslash}p{0.9cm}|>{\raggedleft\arraybackslash}p{0.9cm}>{\raggedleft\arraybackslash}p{0.9cm}>{\raggedleft\arraybackslash}p{0.9cm}|>{\raggedleft\arraybackslash}p{0.9cm}>{\raggedleft\arraybackslash}p{0.9cm}>{\raggedleft\arraybackslash}p{0.9cm}|>{\raggedleft\arraybackslash}p{0.9cm}>{\raggedleft\arraybackslash}p{0.9cm}>{\raggedleft\arraybackslash}p{0.9cm}|>{\raggedleft\arraybackslash}p{0.9cm}>{\raggedleft\arraybackslash}p{0.9cm}>{\raggedleft\arraybackslash}p{0.9cm}|>{\raggedleft\arraybackslash}p{0.9cm}>{\raggedleft\arraybackslash}p{0.9cm}>{\raggedleft\arraybackslash}p{0.9cm}|>{\raggedleft\arraybackslash}p{0.9cm}>{\raggedleft\arraybackslash}p{0.9cm}>{\raggedleft\arraybackslash}p{0.9cm}}
\toprule
\multicolumn{1}{c|}{} & \multicolumn{9}{c|}{LLFF} & \multicolumn{9}{c|}{DL3DV} & \multicolumn{9}{c}{Casual} \\
\midrule
\multicolumn{1}{c|}{} & \multicolumn{3}{c|}{\textbf{G}eometry} & \multicolumn{3}{c|}{\textbf{T}exture} & \multicolumn{3}{c|}{\textbf{A}ll} & \multicolumn{3}{c|}{\textbf{G}eometry} & \multicolumn{3}{c|}{\textbf{T}exture} & \multicolumn{3}{c|}{\textbf{A}ll} & \multicolumn{3}{c|}{\textbf{G}eometry} & \multicolumn{3}{c|}{\textbf{T}exture} & \multicolumn{3}{c}{\textbf{A}ll} \\
\midrule
Feature & \fontsize{8.5pt}{9pt}\selectfont{PSNR$\uparrow$} & \fontsize{8.5pt}{9pt}\selectfont{SSIM$\uparrow$} & \fontsize{8.5pt}{9pt}\selectfont{LPIPS$\downarrow$} & \fontsize{8.5pt}{9pt}\selectfont{PSNR$\uparrow$} & \fontsize{8.5pt}{9pt}\selectfont{SSIM$\uparrow$} & \fontsize{8.5pt}{9pt}\selectfont{LPIPS$\downarrow$} & \fontsize{8.5pt}{9pt}\selectfont{PSNR$\uparrow$} & \fontsize{8.5pt}{9pt}\selectfont{SSIM$\uparrow$} & \fontsize{8.5pt}{9pt}\selectfont{LPIPS$\downarrow$} & \fontsize{8.5pt}{9pt}\selectfont{PSNR$\uparrow$} & \fontsize{8.5pt}{9pt}\selectfont{SSIM$\uparrow$} & \fontsize{8.5pt}{9pt}\selectfont{LPIPS$\downarrow$} & \fontsize{8.5pt}{9pt}\selectfont{PSNR$\uparrow$} & \fontsize{8.5pt}{9pt}\selectfont{SSIM$\uparrow$} & \fontsize{8.5pt}{9pt}\selectfont{LPIPS$\downarrow$} & \fontsize{8.5pt}{9pt}\selectfont{PSNR$\uparrow$} & \fontsize{8.5pt}{9pt}\selectfont{SSIM$\uparrow$} & \fontsize{8.5pt}{9pt}\selectfont{LPIPS$\downarrow$} & \fontsize{8.5pt}{9pt}\selectfont{PSNR$\uparrow$} & \fontsize{8.5pt}{9pt}\selectfont{SSIM$\uparrow$} & \fontsize{8.5pt}{9pt}\selectfont{LPIPS$\downarrow$} & \fontsize{8.5pt}{9pt}\selectfont{PSNR$\uparrow$} & \fontsize{8.5pt}{9pt}\selectfont{SSIM$\uparrow$} & \fontsize{8.5pt}{9pt}\selectfont{LPIPS$\downarrow$} & \fontsize{8.5pt}{9pt}\selectfont{PSNR$\uparrow$} & \fontsize{8.5pt}{9pt}\selectfont{SSIM$\uparrow$} & \fontsize{8.5pt}{9pt}\selectfont{LPIPS$\downarrow$} \\
\midrule
SD      &             \cellcolor[rgb]{1.00,0.82,0.82}19.62 &             \cellcolor[rgb]{1.00,0.82,0.82}.7293 &                \cellcolor[rgb]{1.00,0.82,0.82}.2234 &             \cellcolor[rgb]{1.00,0.82,0.82}18.85 &             \cellcolor[rgb]{1.00,0.82,0.82}.7100 &                \cellcolor[rgb]{1.00,0.82,0.82}.2297 &             \cellcolor[rgb]{0.56,0.75,0.38}19.78 &             \cellcolor[rgb]{1.00,0.82,0.82}.7121 &                \cellcolor[rgb]{1.00,0.82,0.82}.2656 &             \cellcolor[rgb]{1.00,0.82,0.82}19.31 &             \cellcolor[rgb]{1.00,0.82,0.82}.7251 &                \cellcolor[rgb]{1.00,0.82,0.82}.3276 &             \cellcolor[rgb]{1.00,0.82,0.82}17.79 &             \cellcolor[rgb]{1.00,0.82,0.82}.6784 &                \cellcolor[rgb]{1.00,0.82,0.82}.3260 &             \cellcolor[rgb]{1.00,0.82,0.82}19.10 &             \cellcolor[rgb]{1.00,0.82,0.82}.7282 &                \cellcolor[rgb]{1.00,0.82,0.82}.3500 &             \cellcolor[rgb]{0.56,0.75,0.38}19.24 &             \cellcolor[rgb]{1.00,0.82,0.82}.6483 &                \cellcolor[rgb]{0.56,0.75,0.38}.3649 &             \cellcolor[rgb]{1.00,0.82,0.82}17.38 &             \cellcolor[rgb]{0.56,0.75,0.38}.5698 &                \cellcolor[rgb]{0.56,0.75,0.38}.3789 &             \cellcolor[rgb]{0.56,0.75,0.38}18.86 &             \cellcolor[rgb]{0.56,0.75,0.38}.6505 &                \cellcolor[rgb]{0.56,0.75,0.38}.4053 \\
Zero123 &             \cellcolor[rgb]{0.56,0.75,0.38}19.63 &             \cellcolor[rgb]{0.56,0.75,0.38}.7297 &                \cellcolor[rgb]{0.56,0.75,0.38}.2219 &             \cellcolor[rgb]{0.56,0.75,0.38}18.89 &             \cellcolor[rgb]{0.56,0.75,0.38}.7105 &                \cellcolor[rgb]{0.56,0.75,0.38}.2293 &             \cellcolor[rgb]{1.00,0.82,0.82}19.77 &             \cellcolor[rgb]{0.56,0.75,0.38}.7144 &                \cellcolor[rgb]{0.56,0.75,0.38}.2590 &             \cellcolor[rgb]{0.56,0.75,0.38}19.43 &             \cellcolor[rgb]{0.56,0.75,0.38}.7289 &                \cellcolor[rgb]{0.56,0.75,0.38}.3252 &             \cellcolor[rgb]{0.56,0.75,0.38}17.92 &             \cellcolor[rgb]{0.56,0.75,0.38}.6806 &                \cellcolor[rgb]{0.56,0.75,0.38}.3244 &             \cellcolor[rgb]{0.56,0.75,0.38}19.19 &             \cellcolor[rgb]{0.56,0.75,0.38}.7304 &                \cellcolor[rgb]{0.56,0.75,0.38}.3456 &             \cellcolor[rgb]{1.00,0.82,0.82}19.13 &             \cellcolor[rgb]{0.56,0.75,0.38}.6488 &                \cellcolor[rgb]{1.00,0.82,0.82}.3683 &             \cellcolor[rgb]{0.56,0.75,0.38}17.39 &             \cellcolor[rgb]{1.00,0.82,0.82}.5683 &                \cellcolor[rgb]{1.00,0.82,0.82}.3817 &             \cellcolor[rgb]{0.56,0.75,0.38}18.86 &             \cellcolor[rgb]{1.00,0.82,0.82}.6486 &                \cellcolor[rgb]{1.00,0.82,0.82}.4056 \\
\midrule
\multicolumn{1}{c|}{} & \multicolumn{9}{c|}{MipNeRF 360} & \multicolumn{9}{c|}{MVImgNet} & \multicolumn{9}{c}{Tanks and Temples} \\
\midrule
\multicolumn{1}{c|}{} & \multicolumn{3}{c|}{\textbf{G}eometry} & \multicolumn{3}{c|}{\textbf{T}exture} & \multicolumn{3}{c|}{\textbf{A}ll} & \multicolumn{3}{c|}{\textbf{G}eometry} & \multicolumn{3}{c|}{\textbf{T}exture} & \multicolumn{3}{c|}{\textbf{A}ll} & \multicolumn{3}{c|}{\textbf{G}eometry} & \multicolumn{3}{c|}{\textbf{T}exture} & \multicolumn{3}{c}{\textbf{A}ll} \\
\midrule
Feature & \fontsize{8.5pt}{9pt}\selectfont{PSNR$\uparrow$} & \fontsize{8.5pt}{9pt}\selectfont{SSIM$\uparrow$} & \fontsize{8.5pt}{9pt}\selectfont{LPIPS$\downarrow$} & \fontsize{8.5pt}{9pt}\selectfont{PSNR$\uparrow$} & \fontsize{8.5pt}{9pt}\selectfont{SSIM$\uparrow$} & \fontsize{8.5pt}{9pt}\selectfont{LPIPS$\downarrow$} & \fontsize{8.5pt}{9pt}\selectfont{PSNR$\uparrow$} & \fontsize{8.5pt}{9pt}\selectfont{SSIM$\uparrow$} & \fontsize{8.5pt}{9pt}\selectfont{LPIPS$\downarrow$} & \fontsize{8.5pt}{9pt}\selectfont{PSNR$\uparrow$} & \fontsize{8.5pt}{9pt}\selectfont{SSIM$\uparrow$} & \fontsize{8.5pt}{9pt}\selectfont{LPIPS$\downarrow$} & \fontsize{8.5pt}{9pt}\selectfont{PSNR$\uparrow$} & \fontsize{8.5pt}{9pt}\selectfont{SSIM$\uparrow$} & \fontsize{8.5pt}{9pt}\selectfont{LPIPS$\downarrow$} & \fontsize{8.5pt}{9pt}\selectfont{PSNR$\uparrow$} & \fontsize{8.5pt}{9pt}\selectfont{SSIM$\uparrow$} & \fontsize{8.5pt}{9pt}\selectfont{LPIPS$\downarrow$} & \fontsize{8.5pt}{9pt}\selectfont{PSNR$\uparrow$} & \fontsize{8.5pt}{9pt}\selectfont{SSIM$\uparrow$} & \fontsize{8.5pt}{9pt}\selectfont{LPIPS$\downarrow$} & \fontsize{8.5pt}{9pt}\selectfont{PSNR$\uparrow$} & \fontsize{8.5pt}{9pt}\selectfont{SSIM$\uparrow$} & \fontsize{8.5pt}{9pt}\selectfont{LPIPS$\downarrow$} & \fontsize{8.5pt}{9pt}\selectfont{PSNR$\uparrow$} & \fontsize{8.5pt}{9pt}\selectfont{SSIM$\uparrow$} & \fontsize{8.5pt}{9pt}\selectfont{LPIPS$\downarrow$} \\
\midrule
SD      &             \cellcolor[rgb]{1.00,0.82,0.82}20.71 &             \cellcolor[rgb]{0.56,0.75,0.38}.4962 &                \cellcolor[rgb]{1.00,0.82,0.82}.3985 &             \cellcolor[rgb]{1.00,0.82,0.82}18.89 &             \cellcolor[rgb]{1.00,0.82,0.82}.4472 &                \cellcolor[rgb]{1.00,0.82,0.82}.3839 &             \cellcolor[rgb]{1.00,0.82,0.82}20.59 &             \cellcolor[rgb]{1.00,0.82,0.82}.4929 &                \cellcolor[rgb]{1.00,0.82,0.82}.4672 &             \cellcolor[rgb]{0.56,0.75,0.38}19.08 &             \cellcolor[rgb]{0.56,0.75,0.38}.5881 &                \cellcolor[rgb]{0.56,0.75,0.38}.3185 &             \cellcolor[rgb]{1.00,0.82,0.82}16.63 &             \cellcolor[rgb]{1.00,0.82,0.82}.5313 &                \cellcolor[rgb]{1.00,0.82,0.82}.3389 &             \cellcolor[rgb]{1.00,0.82,0.82}19.06 &             \cellcolor[rgb]{1.00,0.82,0.82}.5838 &                \cellcolor[rgb]{1.00,0.82,0.82}.3660 &             \cellcolor[rgb]{0.56,0.75,0.38}18.69 &             \cellcolor[rgb]{0.56,0.75,0.38}.6422 &                \cellcolor[rgb]{0.56,0.75,0.38}.3772 &             \cellcolor[rgb]{1.00,0.82,0.82}17.32 &             \cellcolor[rgb]{1.00,0.82,0.82}.6217 &                \cellcolor[rgb]{1.00,0.82,0.82}.3374 &             \cellcolor[rgb]{0.56,0.75,0.38}18.55 &             \cellcolor[rgb]{0.56,0.75,0.38}.6467 &                \cellcolor[rgb]{1.00,0.82,0.82}.4020 \\
Zero123 &             \cellcolor[rgb]{0.56,0.75,0.38}20.74 &             \cellcolor[rgb]{1.00,0.82,0.82}.4942 &                \cellcolor[rgb]{0.56,0.75,0.38}.3966 &             \cellcolor[rgb]{0.56,0.75,0.38}19.07 &             \cellcolor[rgb]{0.56,0.75,0.38}.4520 &                \cellcolor[rgb]{0.56,0.75,0.38}.3817 &             \cellcolor[rgb]{0.56,0.75,0.38}20.72 &             \cellcolor[rgb]{0.56,0.75,0.38}.4953 &                \cellcolor[rgb]{0.56,0.75,0.38}.4599 &             \cellcolor[rgb]{1.00,0.82,0.82}19.05 &             \cellcolor[rgb]{1.00,0.82,0.82}.5842 &                \cellcolor[rgb]{1.00,0.82,0.82}.3253 &             \cellcolor[rgb]{0.56,0.75,0.38}16.75 &             \cellcolor[rgb]{0.56,0.75,0.38}.5332 &                \cellcolor[rgb]{0.56,0.75,0.38}.3376 &             \cellcolor[rgb]{0.56,0.75,0.38}19.09 &             \cellcolor[rgb]{0.56,0.75,0.38}.5873 &                \cellcolor[rgb]{0.56,0.75,0.38}.3588 &             \cellcolor[rgb]{1.00,0.82,0.82}18.50 &             \cellcolor[rgb]{1.00,0.82,0.82}.6376 &                \cellcolor[rgb]{1.00,0.82,0.82}.3806 &             \cellcolor[rgb]{0.56,0.75,0.38}17.59 &             \cellcolor[rgb]{0.56,0.75,0.38}.6241 &                \cellcolor[rgb]{0.56,0.75,0.38}.3363 &             \cellcolor[rgb]{1.00,0.82,0.82}18.34 &             \cellcolor[rgb]{1.00,0.82,0.82}.6409 &                \cellcolor[rgb]{0.56,0.75,0.38}.4011 \\
\bottomrule
\end{tabular}
}
\vspace{-8pt}
\caption{{ \bf Zero123 \vs Stable Diffusion (SD).} We report quantitative comparision between Zero123 and SD. It demonstrate that Zero123, which fine-tunes Stable Diffusion (SD) on Objaverse~\cite{Objaverse} multi-view dataset, captures geometry and texture better than SD in Objaverse-like simple scenes (LLFF) but performs worse in complex scenes (Tanks and Temples), which might be attributed to catastrophic forgetting~\cite{forgetting}.
}
\vspace{-14pt}
\label{tab:zero123vsSD}
\end{table*}

\setlength{\tabcolsep}{8pt}
\begin{table*}[t]
\setlength{\tabcolsep}{4pt}
\centering
\resizebox{\textwidth}{!}{
\begin{tabular}{l|rrr|rrr|rrr|rrr|rrr|rrr}
\toprule
\multicolumn{1}{c|}{} & \multicolumn{3}{c|}{Fortress (LLFF)} & \multicolumn{3}{c|}{Horns (LLFF)} & \multicolumn{3}{c|}{Orchids (LLFF)} & \multicolumn{3}{c|}{Room (LLFF)} & \multicolumn{3}{c|}{Trex (LLFF)} & \multicolumn{3}{c}{Center (DL3DV)} \\
\midrule
Feature & PSNR $\uparrow$                      & SSIM $\uparrow$                       & LPIPS $\downarrow$                    & PSNR $\uparrow$                      & SSIM $\uparrow$                       & LPIPS $\downarrow$                    & PSNR $\uparrow$                      & SSIM $\uparrow$                       & LPIPS $\downarrow$                    & PSNR $\uparrow$                      & SSIM $\uparrow$                       & LPIPS $\downarrow$                    & PSNR $\uparrow$                      & SSIM $\uparrow$                       & LPIPS $\downarrow$                    & PSNR $\uparrow$                      & SSIM $\uparrow$                       & LPIPS $\downarrow$                    \\
\midrule
DUSt3R  & \cellcolor[rgb]{0.83,0.87,0.64}22.46 & \cellcolor[rgb]{0.91,0.90,0.73}0.7766 & \cellcolor[rgb]{0.83,0.87,0.64}0.1868 & \cellcolor[rgb]{0.74,0.83,0.55}19.87 & \cellcolor[rgb]{0.65,0.79,0.46}0.7428 & \cellcolor[rgb]{0.65,0.79,0.46}0.1999 & \cellcolor[rgb]{0.91,0.90,0.73}16.91 & \cellcolor[rgb]{0.74,0.83,0.55}0.6095 & \cellcolor[rgb]{0.83,0.87,0.64}0.2669 & \cellcolor[rgb]{0.65,0.79,0.46}20.07 & \cellcolor[rgb]{0.65,0.79,0.46}0.8339 & \cellcolor[rgb]{0.65,0.79,0.46}0.2009 & \cellcolor[rgb]{0.65,0.79,0.46}20.06 & \cellcolor[rgb]{0.91,0.90,0.73}0.7580 & \cellcolor[rgb]{0.56,0.75,0.38}0.2073 & \cellcolor[rgb]{0.56,0.75,0.38}17.33 & \cellcolor[rgb]{0.56,0.75,0.38}0.6510 & \cellcolor[rgb]{0.74,0.83,0.55}0.3603 \\
\bottomrule
\end{tabular}
}
\caption{Quantitative Results for Individual Scenes in the \gmode}
\label{tab:perscene_G}
\end{table*}

\setlength{\tabcolsep}{8pt}
\begin{table*}[t]
\setlength{\tabcolsep}{4pt}
\centering
\resizebox{\textwidth}{!}{
\begin{tabular}{l|rrr|rrr|rrr|rrr|rrr|rrr}
\toprule
\multicolumn{1}{c|}{} & \multicolumn{3}{c|}{Fortress (LLFF)} & \multicolumn{3}{c|}{Horns (LLFF)} & \multicolumn{3}{c|}{Orchids (LLFF)} & \multicolumn{3}{c|}{Room (LLFF)} & \multicolumn{3}{c|}{Trex (LLFF)} & \multicolumn{3}{c}{Center (DL3DV)} \\
\midrule
Feature & PSNR $\uparrow$                      & SSIM $\uparrow$                       & LPIPS $\downarrow$                    & PSNR $\uparrow$                      & SSIM $\uparrow$                       & LPIPS $\downarrow$                    & PSNR $\uparrow$                      & SSIM $\uparrow$                       & LPIPS $\downarrow$                    & PSNR $\uparrow$                      & SSIM $\uparrow$                       & LPIPS $\downarrow$                    & PSNR $\uparrow$                      & SSIM $\uparrow$                       & LPIPS $\downarrow$                    & PSNR $\uparrow$                      & SSIM $\uparrow$                       & LPIPS $\downarrow$                    \\
\midrule
DUSt3R  & \cellcolor[rgb]{1.00,0.94,0.82}20.60 & \cellcolor[rgb]{1.00,0.84,0.82}0.7404 & \cellcolor[rgb]{1.00,0.89,0.82}0.2316 & \cellcolor[rgb]{0.83,0.87,0.64}18.98 & \cellcolor[rgb]{1.00,0.94,0.82}0.6957 & \cellcolor[rgb]{0.65,0.79,0.46}0.1959 & \cellcolor[rgb]{1.00,0.94,0.82}16.31 & \cellcolor[rgb]{0.74,0.83,0.55}0.5814 & \cellcolor[rgb]{1.00,0.94,0.82}0.2680 & \cellcolor[rgb]{0.91,0.90,0.73}20.10 & \cellcolor[rgb]{1.00,0.92,0.82}0.8135 & \cellcolor[rgb]{1.00,0.94,0.82}0.2152 & \cellcolor[rgb]{1.00,0.92,0.82}19.07 & \cellcolor[rgb]{1.00,0.94,0.82}0.7291 & \cellcolor[rgb]{0.83,0.87,0.64}0.2204 & \cellcolor[rgb]{0.83,0.87,0.64}15.78 & \cellcolor[rgb]{0.65,0.79,0.46}0.5908 & \cellcolor[rgb]{0.56,0.75,0.38}0.3593 \\
\bottomrule
\end{tabular}
}
\caption{Quantitative Results for Individual Scenes in the  \tmode}
\label{tab:perscene_T}
\end{table*}

\setlength{\tabcolsep}{8pt}
\begin{table*}[t]
\setlength{\tabcolsep}{4pt}
\centering
\resizebox{\textwidth}{!}{
\begin{tabular}{l|rrr|rrr|rrr|rrr|rrr|rrr}
\toprule
\multicolumn{1}{c|}{} & \multicolumn{3}{c|}{Fortress (LLFF)} & \multicolumn{3}{c|}{Horns (LLFF)} & \multicolumn{3}{c|}{Orchids (LLFF)} & \multicolumn{3}{c|}{Room (LLFF)} & \multicolumn{3}{c|}{Trex (LLFF)} & \multicolumn{3}{c}{Center (DL3DV)} \\
\midrule
Feature & PSNR $\uparrow$                      & SSIM $\uparrow$                       & LPIPS $\downarrow$                    & PSNR $\uparrow$                      & SSIM $\uparrow$                       & LPIPS $\downarrow$                    & PSNR $\uparrow$                      & SSIM $\uparrow$                       & LPIPS $\downarrow$                    & PSNR $\uparrow$                      & SSIM $\uparrow$                       & LPIPS $\downarrow$                    & PSNR $\uparrow$                      & SSIM $\uparrow$                       & LPIPS $\downarrow$                    & PSNR $\uparrow$                      & SSIM $\uparrow$                       & LPIPS $\downarrow$                    \\
\midrule
DUSt3R  & \cellcolor[rgb]{0.83,0.87,0.64}21.94 & \cellcolor[rgb]{0.74,0.83,0.55}0.6621 & \cellcolor[rgb]{0.91,0.90,0.73}0.3051 & \cellcolor[rgb]{0.74,0.83,0.55}20.14 & \cellcolor[rgb]{0.91,0.90,0.73}0.7186 & \cellcolor[rgb]{0.91,0.90,0.73}0.2790 & \cellcolor[rgb]{0.65,0.79,0.46}17.25 & \cellcolor[rgb]{0.91,0.90,0.73}0.6290 & \cellcolor[rgb]{1.00,0.92,0.82}0.2838 & \cellcolor[rgb]{1.00,0.89,0.82}19.99 & \cellcolor[rgb]{1.00,0.94,0.82}0.8304 & \cellcolor[rgb]{0.74,0.83,0.55}0.2361 & \cellcolor[rgb]{1.00,0.87,0.82}20.04 & \cellcolor[rgb]{1.00,0.89,0.82}0.7551 & \cellcolor[rgb]{0.91,0.90,0.73}0.2417 & \cellcolor[rgb]{0.83,0.87,0.64}16.85 & \cellcolor[rgb]{0.83,0.87,0.64}0.6532 & \cellcolor[rgb]{0.65,0.79,0.46}0.3773 \\
\bottomrule
\end{tabular}
}
\caption{Quantitative Results for Individual Scenes in the \amode}
\label{tab:perscene_A}
\end{table*}

\setlength{\tabcolsep}{8pt}
\begin{table*}[t]
\setlength{\tabcolsep}{4pt}
\centering
\resizebox{\textwidth}{!}{
\begin{tabular}{l|rrrrrr|rrrrrr|rrrrrr|rrrrrr}
\toprule
\multicolumn{1}{c|}{} & \multicolumn{6}{c|}{Scan1 (DTU)} & \multicolumn{6}{c|}{Scan23 (DTU)} & \multicolumn{6}{c|}{Scan4 (DTU)} & \multicolumn{6}{c}{Scan75 (DTU)} \\
\midrule
Feature & PSNR $\uparrow$                      & SSIM $\uparrow$                       & LPIPS $\downarrow$                    & Acc. $\downarrow$                     & Comp. $\downarrow$                    & Dist. $\downarrow$                    & PSNR $\uparrow$                      & SSIM $\uparrow$                       & LPIPS $\downarrow$                    & Acc. $\downarrow$                     & Comp. $\downarrow$                    & Dist. $\downarrow$                    & PSNR $\uparrow$                      & SSIM $\uparrow$                       & LPIPS $\downarrow$                    & Acc. $\downarrow$                     & Comp. $\downarrow$                    & Dist. $\downarrow$                    & PSNR $\uparrow$                      & SSIM $\uparrow$                       & LPIPS $\downarrow$                    & Acc. $\downarrow$                     & Comp. $\downarrow$                    & Dist. $\downarrow$                    \\
\bottomrule
\end{tabular}
}
\caption{Quantitative Results for Individual Scenes in the \gmode on the DTU dataset}
\label{tab:perscene_G_DTU}
\end{table*}

\section{Limitations and Future Works}
\label{sec:limitations}

\featgs has several limitations. 
First, \featgs requires initialization of camera pose and pointclouds estimated by unconstrained stereo reconstructor~\cite{MASt3R,spann3r,MonST3R}. 
While existing methods, DUSt3R~\cite{DUSt3R} in our case, are robust for initialization, failures sometimes occur. Although \featgs can handle noisy initialization pointcloud, it struggles with those containing significant outliers, as shown in \cref{fig:limitation}.
An exciting direction is to remove this dependence by leveraging VFM features to initialize poses~\cite{foundpose} and pointcloud~\cite{acezero}.
Second, \featgs is designed for controlled settings where scenes are captured in a short time frame under constant lighting. This limits its ability to handle long-term, in-the-wild datasets, where images might be captured hours or years apart, such as internet photo collections of landmarks~\cite{Phototourism,MegaScenes}.
Many works~\cite{nerfwild,Hallucinated,Splatfacto-W,Gaussianwild,Wildgaussians} show that gradients from differentiable rendering are helpful in this case. 
Extending \featgs with these unconstrained formulations could lead to lifelong in-the-wild probing.
Lastly, due to its reliance on 3D Gaussian Splatting, \featgs is currently limited to static scenarios. This is a reasonable assumption for evaluation in multi-view image collections, but restricts assessment in dynamic videos. 4D Gaussian Splatting~\cite{Dynamic_3D_Gaussians,som} may be used to overcome this limitation.

\clearpage\clearpage
\end{appendices}

\end{document}